\documentclass{article}

     \PassOptionsToPackage{numbers, compress}{natbib}

\usepackage{color,soul}
\sethlcolor{gray!30}

\usepackage[final]{neurips_2024}

\usepackage[utf8]{inputenc} %
\usepackage[T1]{fontenc}    %
\usepackage{hyperref}       %
\usepackage{url}            %
\usepackage{booktabs}       %
\usepackage{amsfonts}       %
\usepackage{nicefrac}       %
\usepackage{microtype}      %
\usepackage[table]{xcolor}
\usepackage[pdftex]{graphicx}
\usepackage{amsmath}
\usepackage{stmaryrd}
\usepackage{makecell}
\usepackage{dsfont}
\usepackage{mathtools}
\usepackage{caption}
\usepackage{listings}
\usepackage{algorithm}
\usepackage{algpseudocode}
\usepackage{subcaption}
\usepackage{comment}
\usepackage{float}

\title{Differentiable Task Graph Learning: \\Procedural Activity Representation and Online Mistake Detection from Egocentric Videos}

\author{%
  Luigi Seminara \quad Giovanni Maria Farinella \quad Antonino Furnari \\ \\ Department of Mathematics and Computer Science, University of Catania, Italy \\
  \texttt{luigi.seminara@phd.unict.it,\{giovanni.farinella,antonino.furnari\}@unict.it}
}

\begin{document}

\maketitle

\begin{abstract}
Procedural activities are sequences of key-steps aimed at achieving specific goals. They are crucial to build intelligent agents able to assist users effectively. In this context, task graphs have emerged as a human-understandable representation of procedural activities, encoding a partial ordering over the key-steps. While previous works generally relied on hand-crafted procedures to extract task graphs from videos, in this paper, we propose an approach based on direct maximum likelihood optimization of edges' weights, which allows gradient-based learning of task graphs and can be naturally plugged into neural network architectures. Experiments on the CaptainCook4D dataset demonstrate the ability of our approach to predict accurate task graphs from the observation of action sequences, with an improvement of +16.7\% over previous approaches. Owing to the differentiability of the proposed framework, we also introduce a feature-based approach, aiming to predict task graphs from key-step textual or video embeddings, for which we observe emerging video understanding abilities. Task graphs learned with our approach are also shown to significantly enhance online mistake detection in procedural egocentric videos, achieving notable gains of +19.8\% and +7.5\% on the Assembly101-O and EPIC-Tent-O datasets. Code for replicating the experiments is available at \url{https://github.com/fpv-iplab/Differentiable-Task-Graph-Learning}.
\end{abstract}

\section{Introduction}
Procedural activities are fundamental for humans to organize tasks, improve efficiency, and ensuring consistency in the desired outcomes, but require time and effort to be learned and achieved effectively. This makes the design of artificial intelligent agents able to assist users to correctly perform a task appealing~\cite{kanade2012first,plizzari2023outlook}.
Achieving these abilities requires building a flexible representation of a procedure, encapsulating knowledge on the partial ordering of key-steps arising from the specific context at hand. For example, a virtual assistant needs to understand that it is necessary to break eggs before mixing them or that the bike's brakes need to be released before removing the wheel. Importantly, for a system to be scalable, this representation should be automatically learned from observations (e.g., humans making a recipe many times) rather than explicitly programmed by an expert.

Previous approaches focused on directly tackling tasks requiring procedural knowledge such as action anticipation~\cite{girdhar2021anticipative,furnari2020rolling,roy2024interaction} and mistake detection~\cite{sener2022assembly101,flaborea2024prego,ding2023every,wang2023holoassist,ghoddoosian2023weakly} without developing explicit representations of the procedure. Other works proposed neural models able to develop implicit representations of the procedure by learning how to recover missing actions~\cite{zhong2023learning,narasimhan2023learning}, discover key-steps~\cite{dvornik2023stepformer,bansal2022my,bansal2024united}, or grounding them to video~\cite{dvornik2022graph2vid,lu2022set}.
A different approach~\cite{ashutosh2024video,dvornik2022graph2vid,grauman2023ego} consists in representing the structure of a procedure in the form of a \textit{task graph}, i.e., a Directed Acyclic Graph (DAG) in which nodes represent key-steps, and directed edges impose a partial ordering over key-steps, encoding dependencies between them (see Figure~\ref{fig:teaser}(a)).\footnote{See the supplementary material for more details.} 
Graphs provide an explicit representation which is readily interpretable by humans and easy to incorporate in downstream tasks such as detecting mistakes or validating the execution of a procedure.
While graphs have been historically used to represent constraints in complex tasks and design optimal sub-tasks scheduling~\cite{skiena1998algorithm}, graph-based representations mined from videos~\cite{ashutosh2024video}, key-step sequences~\cite{sohn2020meta,jang2023multimodal} or external knowledge bases~\cite{zhou2023procedure} have only recently emerged as a powerful representation of procedural activities able to support downstream tasks such as key-step recognition or forecasting~\cite{ashutosh2024video,zhou2023procedure}.
Despite these efforts, current methods rely on meticulously crafted graph mining procedures rather than setting graph generation in a learning framework, limiting the inclusion of task graph representations in end-to-end systems.

\begin{figure}[t]
    \centering
    \includegraphics[width=\textwidth]{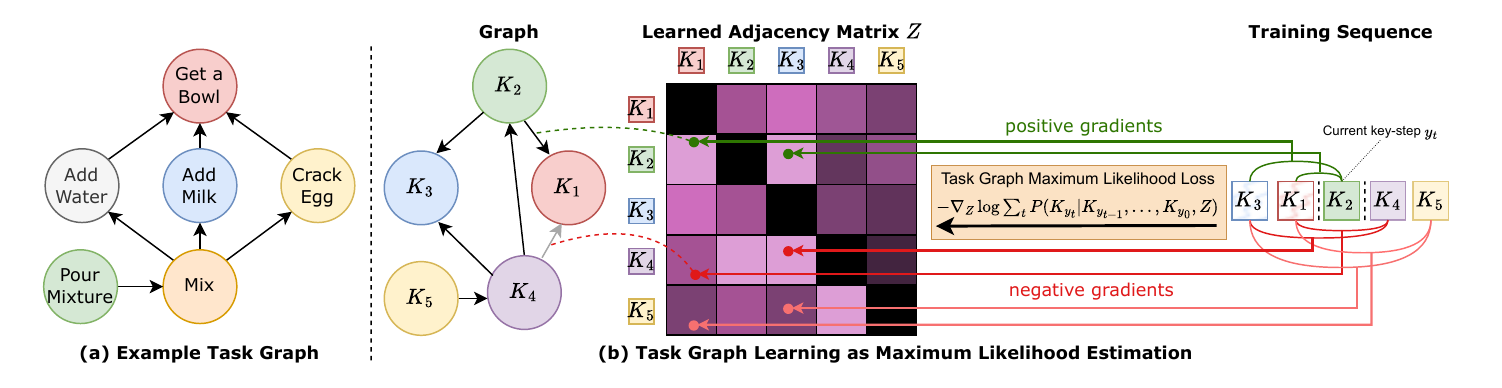}
    \caption{(a) An example task graph encoding dependencies in a ``mix eggs'' procedure. (b) We learn a task graph which encodes a partial ordering between actions (left), represented as an adjacency matrix $Z$ (center), from input action sequences (right). The proposed Task Graph Maximum Likelihood (TGML) loss directly supervises the entries of the adjacency matrix $Z$ generating gradients to maximize the probability of edges from past nodes ($K_3, K_1$) to the current node ($K_2$), while minimizing the probability of edges from past nodes to future nodes ($K_4, K_5$) in a contrastive manner.}
    \label{fig:teaser}
\end{figure}

In this work, we propose a novel approach to learn task graphs from demonstrations in the form of sequences of key-steps performed by real users in a video while executing a procedure. 
Given a directed graph represented as an adjacency matrix and a set of key-step sequences, we provide an estimate of the likelihood of observing the set of sequences given the constraints encoded in the graph. 
We hence formulate task graph learning under the well-understood framework of Maximum Likelihood (ML) estimation, and propose a novel differentiable Task Graph Maximum Likelihood (TGML) loss function which can be naturally plugged into any neural-based architecture for direct optimization of task graph from data.
Intuitively, our TGML loss generates positive gradients to strengthen the weights of directed edges $B \to A$ when observing the $<\ldots,A,\ldots,B,\ldots>$ structure, while pushing down the weights of all other edges in a contrastive manner (see Figure~\ref{fig:teaser}(b)).
To evaluate the effectiveness of the proposed framework, we propose two approaches to task graph learning. The first approach, called ``Direct Optimization (DO)'', uses the proposed TGML loss to directly optimize the weights of the adjacency matrix, which constitute the only parameters of the model. The output of the optimization procedure is the learned graph.
The second approach, termed Task Graph Transformer (TGT) is a feature-based model which uses a transformer encoder and a relation head to predict the adjacency matrix from either text or video key-step embeddings.

We validate the ability of our framework to learn meaningful task graphs on the CaptainCook4D dataset~\cite{peddi2023captaincook4d}. Comparisons with state-of-the-art approaches show superior performance of both proposed approaches on task graph generation, with boosts of up to $+16.7\%$ over prior methods.
On the same dataset, we show that our feature-based approach implicitly gains video understanding abilities on two fundamental tasks~\cite{zhou2015temporal}: pairwise ordering and future prediction.
We finally assess the usefulness of the learned graph-based representation on the downstream task of online mistake detection in procedural egocentric videos.
To tackle this task, we observe that procedural errors mainly arise from the execution of a given key-step without the correct execution of its pre-conditions.
We hence design an approach which uses the learned graph to check whether pre-conditions for the current action are satisfied, signaling a mistake when they are not, obtaining significant gains of +19.8\% and +7.5\% in the online mistake detection benchmark recently introduced in~\cite{flaborea2024prego} on Assembly101~\cite{sener2022assembly101} and EPIC-Tent~\cite{jang2019epic}, showcasing the relevance and quality of the learned graph-based representations.

The contributions of this work are the following: 1) We introduce a novel framework for learning \textit{task graphs} from action sequences, which relies on maximum likelihood estimation to provide a differentiable loss function which can be included in end-to-end models and optimized with gradient descent; 2) We propose two approaches to task graph learning based on direct optimization of the adjacency matrix and processing key-step text or video embeddings, which offer significant improvements over previous methods in task graph generation and shows emerging video understanding abilities; 
3) We showcase the usefulness of task graphs in general, and the learned graph-based representations in particular, on the downstream task of online mistake detection from video, where we improve over competitors.
The code to replicate the experiments is available at~\url{https://github.com/fpv-iplab/Differentiable-Task-Graph-Learning}.

\section{Related Work}

\textbf{Procedure Understanding\hspace{1mm}}
Previous investigations considered different tasks related to procedure understanding, such as inferring key-steps from video in an unsupervised way~\cite{zhou2018towards,zhukov2019cross,elhamifar2020self,bansal2022my,bansal2024united,dvornik2023stepformer}, grounding key-steps in procedural video~\cite{lu2022set,dvornik2021drop,dvornik2022graph2vid,miech2020end}, recognizing the performed procedure~\cite{lin2022learning}, inferring key-step orderings~\cite{bansal2022my,bansal2024united,lu2022set,dvornik2022graph2vid,zhong2023learning}, and procedure structure verification~\cite{narasimhan2023learning}. Recently, task graphs, mined from video or external knowledge such as WikiHow articles, have been investigated as a powerful representation of procedures and proved advantageous for learning representations useful for downstream tasks such as key-step recognition and forecasting~\cite{zhou2023procedure,ashutosh2024video}.

Differently from previous works~\cite{narasimhan2023learning,zhong2023learning}, we aim to develop an explicit and human readable representation of the procedure which can be directly plugged in to enable downstream tasks~\cite{ashutosh2024video}, rather than an implicit representation obtained with pre-training objective~\cite{zhou2023procedure,narasimhan2023learning}. 
As a departure from previous paradigms which carefully designed task graph construction procedures~\cite{ashutosh2024video,zhou2023procedure,sohn2020meta,jang2023multimodal}, we frame task prediction in a general learning framework, enabling models to learn task graphs directly from input sequences, and propose a differentiable loss function based on maximum likelihood.

\textbf{Task Graph Construction\hspace{1mm}}
A line of works investigated the construction of task graphs from natural language descriptions of procedures (e.g., recipes) using rule-based graph parsing~\cite{schumacher2012extraction,dvornik2022graph2vid}, defining probabilistic models~\cite{kiddon2015mise}, fine-tuning language models~\cite{sakaguchi2021proscript}, or proposing learning-based approaches~\cite{dvornik2022graph2vid} involving parsers and taggers trained on text corpora of recipes~\cite{donatelli2021aligning,yamakata2020english}. While these approaches do not require any action sequence as input, they depend on the availability of text corpora including procedural knowledge, such as recipes, which often fail to encapsulate the variety of ways in which the procedure may be executed~\cite{ashutosh2024video}. 
Other works proposed hand-crafted approaches to infer task graphs observing sequences of actions depicting task executions~\cite{jang2023multimodal, sohn2020meta}. Recent work designed procedures to mine task graphs from videos and textual descriptions of key-steps~\cite{ashutosh2024video} or cross-referencing visual and textual representations from corpora of procedural text and videos~\cite{zhou2023procedure}.

Differently from previous efforts, we rely on action sequences, grounded in video, rather than natural language descriptions of procedures or recipes~\cite{sakaguchi2021proscript,dvornik2022graph2vid} and frame task graph generation as a learning problem, providing a differentiable objective rather than resorting to hand-designed algorithms and task extraction procedures~\cite{jang2023multimodal, sohn2020meta,ashutosh2024video,zhou2023procedure}.

\textbf{Online Mistake Detection in Procedural Videos\hspace{1mm}} 
Despite the interest in procedural learning, mistake detection has been systematically investigated only recently.
Some methods considered fully supervised scenarios in which mistakes are explicitly labeled in video and mistake detection is performed offline~\cite{sener2022assembly101,wang2023holoassist,peddi2023captaincook4d}. Other approaches considered weak supervision, with mistakes being labeled only at the video level~\cite{ghoddoosian2023weakly}. Finer-grade spatial and temporal annotations are exploited in~\cite{ding2023every} to build knowledge graphs, which are then leveraged to perform mistake detection. 
Recently, the authors of~\cite{flaborea2024prego} proposed an online mistake detection benchmark incorporating videos from the Assembly101~\cite{sener2022assembly101} and EPIC-Tent~\cite{jang2019epic} datasets, as well as PREGO, an approach to online mistake detection in procedural egocentric videos. 

Rather than addressing online mistake detection with implicit representations~\cite{flaborea2024prego} or carefully designed knowledge bases~\cite{sener2022assembly101}, we design a simple approach which relies on learned explicit task graph representations. As we show in the experiments, this leads to obtain significant performance gains over previous methods, even when the predicted graphs are suboptimal, while best results are obtained with task graphs learned within the proposed framework.

\section{Technical Approach}
\label{sec:tech}

\subsection{Task Graph Maximum Likelihood Learning Framework}
\textbf{Preliminaries\hspace{1mm}} 
Let $\mathcal{K}=\{K_0=S,K_1,\ldots,K_n,K_{n+1}=E\}$ be the set of key-steps involved in the procedure, where $S$ and $E$ are placeholder ``start'' and ``end'' key-steps denoting the \textit{start} and \textit{end} of the procedure. We define the task graph as a directed acyclic graph, i.e., a tuple $G=(\mathcal{K},\mathcal{A},\omega)$, where $\mathcal{K}$ is the set of nodes (the key-steps), $\mathcal{A} = \mathcal{K} \times\mathcal{K}$ is the set of possible directed edges indicating ordering constraints between pairs of key-steps, and $\omega : \mathcal{A} \to [0,1]$ is a function assigning a score to each of the edges in $\mathcal{A}$. 
An edge $(K_i,K_j) \in \mathcal{A}$ (also denoted as $K_i \to K_j$) indicates that $K_j$ is a \textit{pre-condition} of $K_i$ (for instance $\text{mix} \to \text{crack egg}$) with score $\omega(K_i,K_j)$. We assume normalized weights for outgoing edges, i.e., $\sum_j w(K_i,K_j)=1 \forall i$.
We also represent the graph $G$ as the adjacency matrix $Z \in [0,1]^{(n+2) \times (n+2)}$, where $Z_{ij} = \omega(K_i,K_j)$. For ease of notation, we will denote the graph $G=(\mathcal{K},\mathcal{A},\omega)$ simply with its adjacency matrix $Z$ in the rest of the paper. 
We assume that a set of $N$ sequences $\mathcal{Y}=\{y^{(k)}\}_{k=1}^N$ showing possible orderings of the key-steps $\mathcal{K}$ is available, where the generic sequence $y \in \mathcal{Y}$ is defined as a set of indexes to key-steps $\mathcal{K}$, i.e., $y = <y_0,\ldots,y_t,\ldots,y_{m+1}>$, with $y_t \in \{0,\ldots,n+1\}$. We further assume that each sequence starts with key-step $S$ and ends with key-step $E$, i.e., $y_0=0$ and $y_{m+1}=n+1$\footnote{In practice, we prepend/append $S$ and $E$ to each sequence.} and note that different sequences $y^{(i)}$ and $y^{(j)}$ have in general different lengths. Since we are interested in modeling key-step orderings, we assume that sequences do not contain repetitions.\footnote{Since sequences may in practice contain repetitions, we map each sequence containing repetitions to multiple sequences with no repetitions (e.g., $ABCAD \to (ABCD, BCAD)$).} We frame task graph learning as determining an adjacency matrix $\hat Z$ such that sequences in $\mathcal{Y}$ can be seen as topological sorts of $\hat Z$. A principled way to approach this problem is to provide an estimate of the likelihood $P(\mathcal{Y}|Z)$ and choose the maximum likelihood estimate $\hat Z = \underset{Z}{\arg\max} \text{ } P(\mathcal{Y}|Z)$.

\textbf{Modeling Sequence Likelihood for an Unweighted Graph\hspace{1mm}} 
Let us consider the special case of an unweighted graph, i.e., $\bar Z \in \{0,1\}^{(n+2) \times (n+2)}$. We wish to estimate $P(y|Z)$, the likelihood of the generic sequence $y \in \mathcal{Y}$ given graph $Z$.
Formally, let $Y_t$ be the random variable related to the event ``key-step $K_{y_t}$ appears at position $t$ in sequence $y$''. We can factorize the conditional probability $P(y|Z)$ as:
\begin{equation}
\begin{gathered}
P(y|Z)= P(Y_0,\ldots,Y_{|y|}|Z) = P(Y_0|Z) \cdot P(Y_1|Y_0,Z) \cdot \ldots \cdot P(Y_{|y|}|Y_0,\ldots,Y_{|y|-1},Z).
\end{gathered}
\label{eq:factorization}
\end{equation}
We assume that the probability of observing a given key-step $K_{y_t}$ at position $t$ in $y$ depends on the previously observed key-steps ($K_{y_{t-1}},\ldots,K_{y_0}$), but not on their ordering, i.e., the probability of observing a given key-step depends on whether its pre-conditions are satisfied, regardless of the order in which they have been satisfied. 
Under this assumption, we write $P(Y_t | Y_{t-1}, \dots, Y_0,Z)$ simply as $P(K_{y_t} | K_{y_{t-1}}, \dots, K_{y_0},Z)$. 
Without loss of generality, in the following, we denote the current key-step as $K_i=K_{y_t}$, the indexes of key-steps \textit{observed} at time $t$ as $\mathcal{J} = \mathcal{O}(y,t) = \{y_{t-1}, \dots, y_0\}$, and the corresponding set of observed key-steps as $K_{\mathcal{J}} = \{K_i|i\in \mathcal{J}\}$.
Similarly, we define $\bar{\mathcal{J}} = \overline{\mathcal{O}(y,t)} = \{0,\ldots,n+1\} \setminus \mathcal{O}(y,t)$ and $K_{\bar{\mathcal{J}}}$ as the sets of indexes and corresponding key-steps \textit{unobserved} at position $t$, i.e., those which do not appear before $y_t$ in the sequence.
Given the factorization above, we are hence interested in estimating the general term $P(K_{y_t} | K_{y_{t-1}}, \dots, K_{y_0}) = P(K_i|K_{\mathcal{J}})$.
We can estimate the probability of observing key-step $K_i$ given the set of observed key-steps $K_\mathcal{J}$ and the constraints imposed by $\bar Z$, following Laplace's classic definition of probability~\cite{marquis1820theorie} as ``the ratio of the number of favorable cases to the number of possible cases''. 
Specifically, if we were to randomly sample a key-step from $\mathcal{K}$ following the constraints of $\bar Z$, and having observed key-steps $K_\mathcal{J}$, sampling $K_i$ would be a favorable case if all pre-conditions of $K_i$ were satisfied, i.e., if $\sum_{j \in \bar{ \mathcal{J}}} Z_{ij}=0$ (there are no pre-conditions in unobserved key-steps $K_{\bar{\mathcal{J}}}$).
Similarly, sampling a key-steps $K_h$ is a ``possible case'' if $\sum_{j \in \bar{\mathcal{J}}} Z_{hj} = 0$. We can hence define the probability of observing key-step $K_i$ after observing all key-steps $K_\mathcal{J}$ in a sequence as follows:
\begin{equation}
    P(K_i|K_\mathcal{J},\bar Z) = \frac{\text{number of favorable cases}}{\text{number of possible cases}} = \frac{\mathds{1}(\sum_{j \in \bar{\mathcal{J}}} \bar Z_{ij} = 0)}{\sum_{h \in \bar{\mathcal{J}}} \mathds{1}(\sum_{j \in \bar{\mathcal{J}}} \bar Z_{hj} = 0)}
    \label{eq:binary_factor}
\end{equation}
where $\mathds{1}(\cdot)$ denotes the indicator function, and in the denominator, we are counting the number of key-steps that have not appeared yet are ``possible cases'' under the given graph $Z$. Likelihood $P(y|Z)$ can be obtained by plugging Eq.~\eqref{eq:binary_factor} into Eq.~\eqref{eq:factorization}.

\textbf{Modeling Sequence Likelihood for a Weighted Graph\hspace{1mm}} 
To enable gradient-based learning, we consider the general case of a continuous adjacency matrix $Z \in [0,1]^{(n+2) \times (n+2)}$. 
We generalize the concept of ``possible cases'' discussed in the previous section with the concept of ``feasibility of sampling a given key-step $K_i$, having observed a set of key-steps $K_\mathcal{J}$, given graph $Z$'', which we define as the sum of all weights of edges between observed key-steps $K_\mathcal{J}$ and $K_i$: $f(K_i|K_\mathcal{J},Z) = \sum_{j \in \mathcal{J}} Z_{ij}$.
Intuitively, if key-step $k_i$ has many satisfied pre-conditions, we are more likely to sample it as the next key-step. We hence define $P(K_i|K_\mathcal{J},Z)$ as ``the ratio of the feasibility of sampling $K_i$ to the sum of the feasibilities of sampling any unobserved key-step'':
\begin{equation}
    P(K_i|K_\mathcal{J},Z) = \frac{f(K_i | K_\mathcal{J},Z)}{\sum_{h \in \bar{\mathcal{J}}} f(K_{h}|K_\mathcal{J},Z)}  = \frac{\sum_{j \in \mathcal{J}}Z_{ij}}{\sum_{h \in \bar{\mathcal{J}}} \sum_{j \in \mathcal{J}} Z_{hj}}
    \label{eq:factor}
\end{equation}

\begin{figure}
    \centering
    \includegraphics[width=\linewidth]{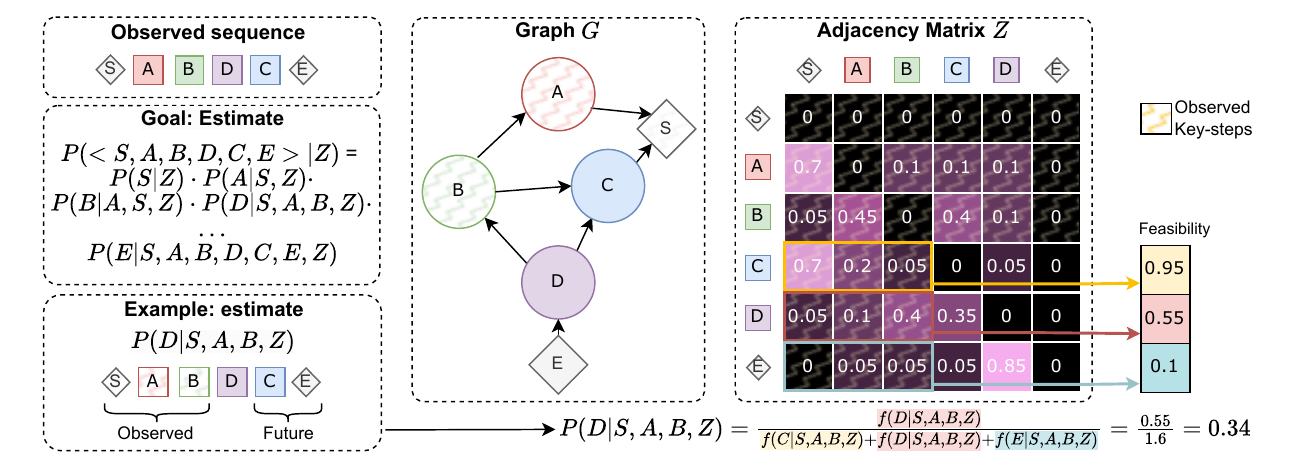}
    \caption{Given a sequence $<S,A,B,D,C,E>$, and a graph $G$ with adjacency matrix $Z$, our goal is to estimate the likelihood $P(<S,A,B,D,C,E>|Z)$, which can be done by factorizing the expression into simpler terms. The figure shows an example of computation of probability $P(D|S,A,B,Z)$ as the ratio of the ``feasibility of sampling key-step D, having observed key-steps S, A, and B'' to the sum of all feasibility scores for unobserved symbols. Feasibility values are computed by summing weights of edges $D \to X$ for all observed key-steps $X$.}
    \label{fig:prob}
\end{figure}

Figure~\ref{fig:prob} illustrates the computation of the likelihood in Eq.~\eqref{eq:factor}.
Plugging Eq.~\eqref{eq:factor} into Eq.~\eqref{eq:factorization}, we can estimate the likelihood of a sequence $y$ given graph $Z$ as:
\begin{align}
    \label{eq:factorization_2}
    P(y|Z)
    = P(S|Z)\prod_{t = 1}^{|y|}{P(K_{y_t}|K_{\mathcal{O}(y,t)},Z)} = \prod_{t = 1}^{|y|} \frac{\sum_{j \in \mathcal{O}(y,t)}Z_{y_tj}}{\sum_{h \in \overline{\mathcal{O}(y,t)}} \sum_{j \in \mathcal{O}(y,t)} Z_{hj}}.
\end{align}
Where we set $P(K_{y_0}|Z)=P(S|Z)=1$ as sequences always start with the start node $S$.

\textbf{Task Graph Maximum Likelihood Loss Function\hspace{1mm}} 
Assuming that sequences $y^{(i)} \in \mathcal{Y}$ are independent and identically distributed, we define the likelihood of $\mathcal{Y}$ given graph $Z$ as follows:
\begin{equation}
    P(\mathcal{Y}| Z) = \prod_{k=1}^{|\mathcal{Y}|}{P(y^{(k)}|Z)} = 
    \prod_{k=1}^{|\mathcal{Y}|}\prod_{t = 1}^{|y^{(k)}|}\frac{\sum_{j \in \mathcal{O}(y^{(k)},t)}Z_{y_tj}}{\sum_{h \in \overline{\mathcal{O}(y^{(k)},t)}} \sum_{j \in \mathcal{O}(y^{(k)},t)} Z_{hj}}.
    \label{eq:likelihood}
\end{equation}
We can find the optimal graph $Z$ by maximizing the likelihood in Eq.~\eqref{eq:likelihood}, which is equivalent to minimizing the negative log-likelihood $-\log P(\mathcal{Y},Z)$, leading to formulating the following loss:
\begin{align}
    \mathcal{L}(\mathcal{Y},Z) =  - \sum_{k = 1}^{|Y|} \sum_{t = 1}^{|y^{(k)}|} \big(\textcolor{cyan}{\log{\sum_{\mathclap{j \in \mathcal{O}(y^{(k)},t)}}Z_{y_tj}}} 
    - \beta \cdot \textcolor{teal}{\log{\sum_{\mathclap{\substack{h \in \overline{\mathcal{O}(y^{(k)},t)}\\{j \in \mathcal{O}(y^{(k)},t)}}}}  Z_{hj}}} \big)
    \label{eq:loss}
\end{align}
where $\beta$ is a hyper-parameter. We refer to Eq.~\eqref{eq:loss} as the \textit{Task Graph Maximum Likelihood (TGML)} loss function. Since Eq.~\eqref{eq:loss} is differentiable with respect to all $Z_{ij}$ values, we can learn the adjacency matrix $Z$ by minimizing the loss with gradient descent to find the estimated graph $\hat Z = \arg_Z\max \mathcal{L}(\mathcal{Y},Z)$. 
Eq.~\eqref{eq:loss} works as a contrastive loss in which the \textcolor{cyan}{first logarithmic term} aims to \textit{maximize}, at every step $t$ of each input sequence, the weights $Z_{y_tj}$ of edges $K_{y_t} \to K_j$ going from the current key-step $K_{y_t}$ to all previously observed key-steps $K_j$, while the \textcolor{teal}{second logarithmic term (contrastive term)} aims to \textit{minimize} the weights of edges $K_h \to K_j$ between key-steps yet to appear $K_h$ and already observed key-steps $K_j$.
The hyper-parameter $\beta$ regulates the influence of the summation in the \textcolor{teal}{contrastive term} which, including many more addends, can dominate gradient updates. As in other contrastive learning frameworks~\cite{oord2018representation, radford2021learning}, our approach only includes positives and negatives and it does not explicitly consider anchor examples.

\subsection{Models}
\label{sec:models}
\textbf{Direct Optimization (DO)\hspace{1mm}} 
The first model aims to directly optimize the parameters of the adjacency matrix by performing gradient descent on the TGML loss (Eq.~\eqref{eq:loss}).
We define the parameters of this model as an edge scoring matrix $A \in \mathbb{R}^{(n+2) \times (n+2)}$, where $n$ is the number of key-steps, plus the placeholder start ($S$) and end ($E$) nodes, and $A_{ij}$ is a score assigned to edge $K_i \rightarrow K_j$.
To prevent the model from learning edge weights eluding the assumptions of directed acyclic graphs, we mask black cells in Figure~\ref{fig:prob} with $-\infty$.
To constrain the elements of $Z$ in the $[0,1]$ range and obtain normalized weights, we softmax-normalize the rows of the scoring matrix to obtain the adjacency matrix $Z = softmax(A)$. Note that elements masked with $-\infty$ will be automatically mapped to $0$ by the softmax function similarly to~\cite{vaswani2017attention}. We train this model by performing batch gradient descent directly on the score matrix $A$ with the proposed TGML loss. We train a separate model per procedure, as each procedure is associated to a different task graph. 
As many applications require an unweighted graph, we binarize the adjacency matrix with the threshold $\frac{1}{n}$, where $n$ is the number of nodes. We also employ a post-processing stage in which we remove redundant edges, loops, and add obvious missing connections to $S$ and $E$ nodes.\footnote{\label{note:supp}See the supplementary material for more details.}

\begin{figure}
    \centering
    \includegraphics[width=\linewidth]{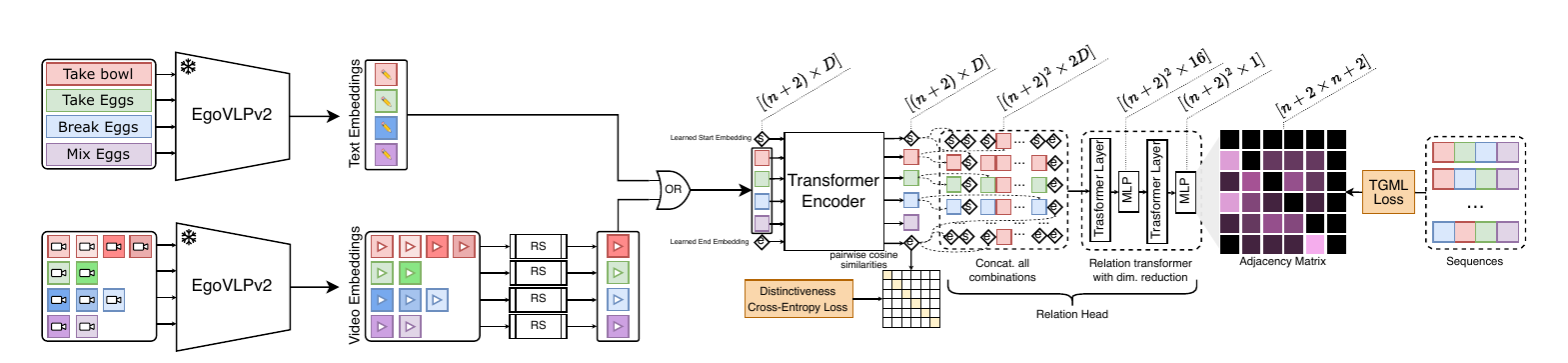}
    \caption{Our Task Graph Transformer (TGT) takes as input either $D$-dimensional text embeddings extracted from key-step names or video embeddings extracted from key-step segments. In both cases, we extract features with a pre-trained EgoVLPv2 model. For video embeddings, multiple embeddings can refer to the same action, so we randomly select one for each key-step (RS blocks). Learnable start (S) and end (E) embeddings are also included. Key-step embeddings are processed using a transformer encoder and regularized with a distinctiveness cross-entropy to prevent representation collapse. The output embeddings are processed by our relation head, which concatenates vectors across all $(n + 2)^2$ possible node pairs, producing $(n + 2) \times (n + 2) \times 2D$ relation vectors. These vectors are then processed by a relation transformer, which progressively maps them to an $(n + 2) \times (n + 2)$ adjacency matrix. The model is supervised with input sequences using our proposed Task Graph Maximum Likelihood (TGML) loss.}
    \label{fig:architecture}
\end{figure}

\textbf{Task Graph Transformer (TGT)\hspace{1mm}} 
Figure~\ref{fig:architecture} illustrates the proposed model, which is termed Task Graph Transformer (TGT). The proposed model can take as input either $D$-dimensional embeddings of textual descriptions of key-steps or $D$-dimensional video embeddings of key-step segments extracted from video. In the first case, the model takes as input the same set of embeddings at each forward pass, while in the second case, at each forward pass, we randomly sample a video embedding per key-step from the training videos (hence each key-step embedding can be sampled from a different video). We also include two $D$-dimensional learnable embeddings for the $S$ and $E$ nodes. 
All key-step embeddings are processed by a transformer encoder, which outputs $D$-dimensional vectors enriched with information from other embeddings. To prevent representation collapse, we apply a regularization loss encouraging distinctiveness between pairs of different nodes. Let $X$ be the matrix of embeddings produced by the transformer model. 
We L2-normalize features, then compute pairwise cosine similarities $Y = X \cdot X^T \cdot \exp(T)$ as in~\cite{radford2021learning}.
To prevent the transformer encoder from mapping distinct key-step embeddings to similar representations, we enforce the values outside the diagonal of $Y$ to be smaller than the values in the diagonal. This is done by encouraging each row of the matrix $Y$ to be close to a one-hot vector with a cross-entropy loss. Regularized embeddings are finally passed through a relation transformer head which considers all possible pairs of embeddings and concatenates them in a $(n+2) \times (n+2) \times 2D$ matrix $R$ of relation vectors. For instance, $R[i,j]$ is the concatenation of vectors $X[i]$ and $X[j]$. Relation vectors are passed to a transformer layer which aims to mine relationships among relation vectors, followed by a multilayer perceptron to reduce dimensionality to $16$ units and another pair of transformer layer and multilayer perceptron to map relation vectors to scalar values, which are reshaped to size $(n+2) \times (n+2)$ to form the score matrix $A$. We hence apply the same optimization procedure as in the DO method to supervise the whole architecture.

\section{Experiments and Results}
\label{sec:experiments}

\subsection{Graph Generation}
\textbf{Problem Setup\hspace{1mm}} 
We evaluate the ability of our approach to learn task graph representations on CaptainCook4D~\cite{peddi2023captaincook4d}, a dataset of egocentric videos of 24 cooking procedures performed by 8 volunteers. Each procedure is accompanied by a task graph describing key-steps constraints. 
We tackle task graph generation as a weakly supervised learning problem in which models have to generate valid graphs by only observing labeled action sequences (weak supervision) rather than relying on task graph annotations (strong supervision), which are not available at training time. 
All models are trained on videos that are free from ordering errors or missing steps to provide a likely representation of procedures. 
We use the two proposed methods in the previous section to learn $24$ task graph models, one per procedure, and report average performance across procedures.
\begin{figure}
\begin{minipage}[t]{0.57\textwidth}%
  \captionof{table}{Task graph generation results on CaptainCook4D. Best results are in \textbf{bold}, second best results are \underline{underlined}, best results among competitors are \hl{highlighted}. Confidence interval bounds computed at $90\%$ conf. for $5$ runs.}
  \label{tab:captaincook}
  \centering
  \begin{tabular}{llll}
    \toprule
    Method         & Precision     & Recall     & F$_1$          \\
    \midrule
    MSGI \cite{sohn2020meta}                    & 11.9 & 14.0 & 12.8 \\
    LLM                 & 52.9 & 57.4 & 55.0\\
    Count-Based \cite{ashutosh2024video}             & 66.7 & 55.6 & 60.6 \\
    MSG\(^2\) \cite{jang2023multimodal}               & \hl{70.9} & \hl{71.6} & \hl{71.1} \\
    \rowcolor{teal!30}
    {TGT-text (Ours)}               & {\underline{79.9} \scriptsize \(\pm 8.8\)}& {\underline{81.9} \scriptsize \(\pm 6.9\)} & {\underline{80.8} \scriptsize \(\pm 8.0\)}\\
    \rowcolor{teal!30}
    {DO (Ours)}        & {\textbf{86.4} \scriptsize $\pm 1.5$ }& {\textbf{89.7} \scriptsize \(\pm 1.5\) }& {\textbf{87.8} \scriptsize \(\pm 1.5\) }\\
    \rowcolor{green!30}
    Improvement & +15.5 & +18.1 & +16.7 \\
    \bottomrule
  \end{tabular}
    \end{minipage}%
    \hspace{1.5mm}
    \begin{minipage}[t]{0.4\textwidth}%
  \captionof{table}{We compare the abilities of our TGT model trained on visual features to generalize to two fundamental video understanding tasks, i.e., pairwise ordering and future prediction. Despite not being explicitly trained for these tasks, our model exhibits video understanding abilities, surpassing the baseline.}
  \label{tab:understanding}
  \centering
  \begin{tabular}{lll}
    \toprule
    Method         & Ordering     & Fut. Pred.  \\
    \midrule
    Random         & 50.0         & 50.0 \\
    \rowcolor{teal!30}
    TGT-video& \textbf{77.3} & \textbf{74.3} \\
    \rowcolor{green!30}
    Improvement & +27.3 & +24.3 \\
    \bottomrule
  \end{tabular}
    \end{minipage}%
\end{figure}

\textbf{Compared Approaches\hspace{1mm}} 
We compare our methods with previous approaches to task graph generation, and in particular with MSGI \cite{sohn2020meta} and MSG$^2$ \cite{jang2023multimodal}, which are approaches for task graph generation based on Inductive Logic Programming (ILP). 
We also consider the recent approach proposed in~\cite{ashutosh2024video} which generates a graph by counting co-occurrences of matched video segments. Since we assume labeled actions to be available at training time, we do not perform video matching and use ground truth segment matching provided by the annotations. This approach is referred to as ``Count-Based''. Given the popularity of large language models as reasoning modules, we also consider a baseline which uses a large language model\footnote{We base our experiments on ChatGPT~\cite{achiam2023gpt}.} to generate a task graph from key-step descriptions, without any access to key-step sequences.$^{\ref{note:supp2}}$
We refer to this model as ``LLM''.

\textbf{Graph Generation Results\hspace{1mm}} 
Results in Table~\ref{tab:captaincook} highlight the complexity of the task, with classic approaches based on inductive logic, such as MSGI, achieving poor performance ($12.8$ $F_1$), language models and count-based statistics reconstructing only basic elements of the graph ($55.0$ and $60.6$ $F_1$ for LLM and Count-Based respectively), and even more recent methods based on inductive logic and heuristics only partially predicting the graph ($71.1$ $F_1$ of $MSG^2$). The proposed Direct Optimization (DO) approach outperforms all other methods, achieving the highest scores across all measures, with improvements in the $[+15.5, +18.1]$ range with respect to the best competitor $MSG^2$. This result highlights the effectiveness of the proposed framework to learn task graph representations from key-step sequences, especially considering the simplicity of the DO method, which performs gradient descent directly on the adjacency matrix. We obtain a slightly higher recall as compared to the precision ($89.7$ vs $86.4$), showing that our approach tends to retrieve most ground truth edges, while hallucinating some pre-conditions, probably due to the dataset being unbalanced towards the most common ways of completing a procedure. Second best results are consistently obtained by our feature-based TGT approach, showing the generality of our learning framework and the potential of integrating it into complex neural architectures. Tight confidence intervals for DO highlight the stability of the proposed loss.
The lower performance of TGT, as compared to DO, may be due to the relatively small size of the dataset, which makes it hard for complex architecture to generalize. 

\textbf{Video Understanding Results\hspace{1mm}} 
Table~\ref{tab:understanding} reports the performance of TGT trained on videos on two fundamental video understanding tasks~\cite{zhou2015temporal} of pairwise clip ordering and future prediction.\footnote{\label{note:supp2}See the supplementary material for more details.} For pairwise ordering, we feed our TGT model with video embeddings of two clips and sort them according to the predicted adjacency matrix, placing first the clip identified as a pre-condition. For future predictions, given an anchor clip, we have to choose which among two other clips is the correct future. Despite TGT not being explicitly trained for pairwise ordering and future predictions, it exhibits emerging video understanding abilities, surpassing the random baseline.

\subsection{Online Mistake Detection}
\textbf{Problem Setup\hspace{1mm}} 
We follow the PREGO benchmark {and used the datasets (Assembly101-O and EPIC-Tent-O)} recently proposed in~\cite{flaborea2024prego}, in which models are tasked to perform online action detection from procedural egocentric videos. To evaluate the usefulness of task graphs on this downstream task, we design a system which flags the current action as a mistake if its pre-conditions in the predicted graph do not appear in previously observed actions.$^{\ref{note:supp2}}$

\textbf{Competitors\hspace{1mm}} 
We compare our approach with respect to the PREGO model proposed in~\cite{flaborea2024prego}, which detects mistakes based on the comparison between the currently observed action and an action predicted by a forecasting module. We note that PREGO is based on an implicit representation of the procedure (the forecasting module), while our approach is based on the explicit task graph representation, learned with the proposed framework. We also compare our approach with respect to baselines based on all graph prediction approaches compared in Table~\ref{tab:captaincook} to assess how the ability to predict accurate graphs affects downstream performance. For all methods, we report results based on ground truth action segments and on action sequences predicted by a MiniRoad~\cite{an2023miniroad} instance, a state-of-the-art online action detection module trained on each target dataset.

\textbf{Results\hspace{1mm}} 
Results in Table~\ref{tab:online_mistake_detection_results} highlight the usefulness of the learned task graphs for downstream applications. The proposed DO method achieves significant gains over prior art with improvements of $+19.8$ and $+7.5$ in average $F_1$ score on {Assembly101-O} and {EPIC-Tent-O} respectively when ground truth action sequences are considered to make predictions. While TGT is the second-best performer on Assembly101-O, it obtains best results on EPIC-Tent-O ($64.1$ vs $58.3$ in average $F_1$ score). This is due to the nature of action annotations in the two datasets. Indeed, while key-step names are informative in EPIC-Tent (e.g., ``Place Vent Cover'', ``Open Stake Bag'', or ``Spread Tent''), they are less distinctive in Assembly101 (e.g., ``attach cabin'', ``attach interior'', or ``screw chassis''). This highlights the flexibility of the proposed learning framework which can work in purely abstract, symbolic settings, with the DO approach, but can also leverage semantics with TGT when beneficial. Interestingly, the second best performers are graph-based approaches, with $MSG^2$ achieving an average $F_1$ of $56.1$ on Assembly101-O and the simple Count-Based approach obtaining an average $F_1$ score of $56.6$ on EPIC-Tent-O. In contrast, PREGO obtains average $F_1$ scores of $39.4$ and $32.1$ on Assembly101-O and EPIC-Tent-O respectively, suggesting the potential of explicit graph-based representations for mistake detection, versus the implicit one of PREGO. Breaking down performance into correct and mistake $F_1$ scores reveal some degree of unbalance of our approaches and the main competitor $MSG^2$ towards identifying correct actions rather than mistakes. This suggests that the related graph-based representations tend to detect some spurious pre-conditions, probably due to the limited demonstrations included in the videos, while the implicit PREGO model exhibits a skew with respect to mistakes. Further breaking down $F_1$ scores into related precision and recall values highlights that the main failure modes are due to large imbalances between precision and recall. For instance, the Count-Based method achieves a precision of only $4.8$ with a recall of $85.7$ in predicting correct segments on Assembly101-O. In contrast, the proposed approach obtains balanced precision and recall values in detecting correct segments in Assembly101-O ($98.2$/$83.4$) and EPIC-Tent-O ($94.1$/$93.5$), and detecting mistakes in EPIC-Tent-O ($33.3$/$35.7$), while the prediction of mistakes on Assembly101-O is more skewed ($46.7$/$90.4$). 
Results based on action sequences predicted from video (bottom part of Table~\ref{tab:online_mistake_detection_results}) highlight the challenging nature of the task when considering noisy action sequences (see Figure~\ref{fig:test}). While the explicit task graph representation may not accurately reflect the predicted noisy action sequences, we still observe improvements over previous approaches of $+7.3$ and $+1.3$ in average $F_1$ score in Assembly101-O and EPIC-Tent-O. Remarkably, best competitors are still graph-based methods, such as $MSG^2$ and the Count-Based approach, with significant improvements over the implicit representation of the PREGO model ($32.5$ average $F_1$ versus $53.5$ of the proposed DO model). Also, in this case, we observe that graph-based methods tend to be skewed towards detecting correct action sequences. In this regard, our TGT model only achieves $38.2$ in mistake $F_1$ score, a drop in $5.7$ points over the best performer, the Count-Based method, which, on the other hand, only achieves an $F_1$ score of $2.6$ when predicting correct segments. 
\begin{table}
  \caption{Online mistake detection results. Results obtained with ground truth action sequences are denoted with $^*$, while results obtained on predicted action sequences are denoted with $^+$.}
  \label{tab:online_mistake_detection_results}
  \centering
  \resizebox{\linewidth}{!}{
  \setlength\tabcolsep{4pt}
      \begin{tabular}{lccccccccclccccccccc}
      
        \toprule
        & \multicolumn{7}{c}{Assembly101-O} & \multicolumn{8}{c}{EPIC-Tent-O} \\
        \cmidrule(lr){2-8} \cmidrule(lr){10-16}
        & \multicolumn{1}{c}{Avg} & \multicolumn{3}{c}{Correct} & \multicolumn{3}{c}{Mistake} & & \multicolumn{1}{c}{Avg} & \multicolumn{3}{c}{Correct} & \multicolumn{3}{c}{Mistake} \\
        \cmidrule(lr){2-2} \cmidrule(lr){3-5} \cmidrule(lr){6-8} \cmidrule(lr){10-10} \cmidrule(lr){11-13} \cmidrule(lr){14-16} 
        Method & F$_1$ & F$_1$ & \scriptsize{Prec} & \scriptsize{Rec} & F$_1$ & \scriptsize{Prec} & \scriptsize{Rec} & & F$_1$ & F$_1$ & \scriptsize{Prec} & \scriptsize{Rec} & F$_1$ & \scriptsize{Prec} & \scriptsize{Rec} \\
        \midrule
        Count-Based$^*$ \cite{ashutosh2024video}  & 26.0 & 9.2 & \scriptsize{4.8} & \scriptsize{85.7} & 42.8 & \scriptsize{97.8} & \scriptsize{27.4} & & \hl{56.6} & 92.5 & \scriptsize{92.8} & \scriptsize{92.2} & 20.7 & \scriptsize{20.0} & \scriptsize{21.4}  \\
        LLM$^*$  & 29.3 & 15.1 & \scriptsize{8.3} & \scriptsize{87.2} & 43.4 & \scriptsize{96.7} & \scriptsize{27.9} & & 47.7 & 86.3 & \scriptsize{82.4} & \scriptsize{90.6} & 9.1 & \scriptsize{13.3} & \scriptsize{6.9}  \\
        MSGI$^*$ \cite{sohn2020meta}   & 33.1 & 22.7 & \scriptsize{13.1} & \scriptsize{84.4} & 43.5 & \scriptsize{93.4} & \scriptsize{28.3} & & 44.5 & 66.9 & \scriptsize{51.6} & \scriptsize{95.2} & \hl{22.0} & \scriptsize{73.3} & \scriptsize{12.9}  \\
        PREGO$^*$ \cite{flaborea2024prego}   & 39.4 & 32.6 & \scriptsize{89.7} & \scriptsize{19.9} & 46.3 & \scriptsize{30.7} & \scriptsize{94.0} & & 32.1 & 45.0 & \scriptsize{95.7} & \scriptsize{29.4} & 19.1 & \scriptsize{10.7} & \scriptsize{86.7}  \\
        MSG$^{2*}$ \cite{jang2023multimodal}  & \hl{56.1} & \hl{63.9} & \scriptsize{51.5} & \scriptsize{84.2} & \hl{48.2} & \scriptsize{73.6} & \scriptsize{35.8} & & 54.1 & \hl{92.9} & \scriptsize{94.1} & \scriptsize{91.7} & 15.4 & \scriptsize{13.3} & \scriptsize{18.2}  \\
        \rowcolor{teal!30} TGT-text (Ours)$^*$ & \underline{62.8} & \underline{69.8} & \scriptsize{56.8} & \scriptsize{90.6} & \underline{55.7} & \scriptsize{84.1} & \scriptsize{41.7} & & \textbf{64.1} & \textbf{93.8} & \scriptsize{94.1} & \scriptsize{93.5} & \textbf{34.5} & \scriptsize{33.3} & \scriptsize{35.7} \\
        \rowcolor{teal!30} DO (Ours)$^*$ & \textbf{75.9} & \textbf{90.2} & \scriptsize{98.2} & \scriptsize{83.4} & \textbf{61.6} & \scriptsize{46.7} & \scriptsize{90.4} & & \underline{58.3} & \underline{93.5} & \scriptsize{94.8} & \scriptsize{92.4} & \underline{23.1} & \scriptsize{20.0} & \scriptsize{27.3} \\
        \rowcolor{green!30} Improvement$^*$ & +19.8 & +26.3 & &  & +13.4 & & & & +7.5 & +0.9 & & & +12.5 & &\\
        \midrule
        Count-Based$^+$ \cite{ashutosh2024video}  & 23.2 & 2.6 & \scriptsize{1.3} & \scriptsize{66.7} & \textbf{\hl{43.9}} & \scriptsize{98.4} & \scriptsize{28.2} & & 40.4 & 59.2 & \scriptsize{42.9} & \scriptsize{95.5} & 21.6 & \scriptsize{80.0} & \scriptsize{12.5}  \\
        LLM$^+$ & 28.1 & 15.1 & \scriptsize{7.8} & \scriptsize{65.5} & 42.3 & \scriptsize{89.5} & \scriptsize{27.7} & & 35.9 & 61.6 & \scriptsize{46.7} & \scriptsize{90.4} & 10.2 & \scriptsize{40.0} & \scriptsize{5.8}  \\
        MSGI$^+$ \cite{sohn2020meta}   & 28.4 & 14.0 & \scriptsize{7.8} & \scriptsize{67.9} & \underline{42.7} & \scriptsize{90.7} & \scriptsize{28.0} & & 40.4 & 59.2 & \scriptsize{42.9} & \scriptsize{95.5} & 21.6 & \scriptsize{80.0} & \scriptsize{12.5}  \\
        PREGO$^+$ \cite{flaborea2024prego}   & 32.5 & 23.1 & \scriptsize{68.8} & \scriptsize{13.9} & 41.8 & \scriptsize{27.8} & \scriptsize{84.1} & & 29.4 & 41.6 & \scriptsize{97.9} & \scriptsize{26.4} & 17.2 & \scriptsize{9.5} & \scriptsize{93.3}  \\
        MSG$^{2+}$ \cite{jang2023multimodal}  & \hl{46.2} & \hl{59.1} & \scriptsize{51.2} & \scriptsize{70.0} & 33.2 & \scriptsize{44.5} & \scriptsize{26.5} & & \underline{{\hl{45.2}}} & {\hl{67.5}} & \scriptsize{52.4} & \scriptsize{95.1} & \underline{{\hl{22.9}}} & \scriptsize{73.3} & \scriptsize{13.6}  \\
        \rowcolor{teal!30} TGT-text (Ours)$^+$ & \underline{53.0} & \underline{67.8} & \scriptsize{62.3} & \scriptsize{74.5} & 38.2 & \scriptsize{46.2} & \scriptsize{32.6} & & 43.8 & \textbf{69.5} & \scriptsize{55.8} & \scriptsize{92.1} & 18.2 & \scriptsize{53.3} & \scriptsize{11.0} \\
        \rowcolor{teal!30} DO (Ours)$^+$ & \textbf{53.5} & \textbf{78.9} & \scriptsize{85.0} & \scriptsize{73.5} & 28.1 & \scriptsize{22.5} & \scriptsize{37.3} & & \textbf{46.5} & \underline{69.3} & \scriptsize{54.4} & \scriptsize{95.2} & \textbf{23.7} & \scriptsize{73.3} & \scriptsize{14.1} \\
        \rowcolor{green!30} Improvement$^+$ & +7.3 & +19.8 & &  & -5.7 & & & & +1.3 & +1.2 & & & +1.2 & & \\
        
        \bottomrule
      \end{tabular}
  }
\end{table}

\section{Limitations}
\label{sec:limitations}
\begin{figure}
\centering
\begin{subfigure}{.49\textwidth}
  \centering
  \includegraphics[width=\linewidth]{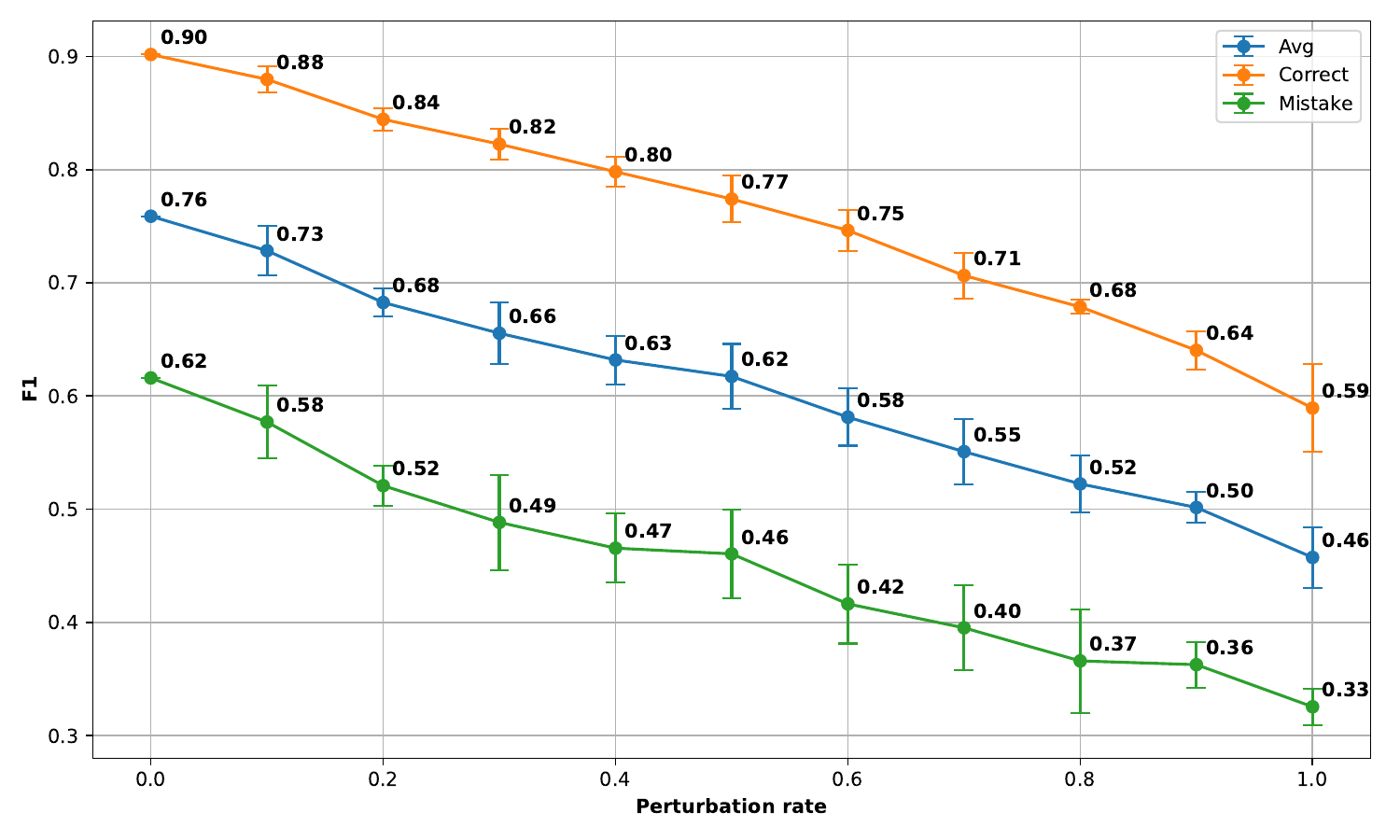}
  \label{fig:sub1}
\end{subfigure}
\begin{subfigure}{.49\textwidth}
  \centering
  \includegraphics[width=\linewidth]{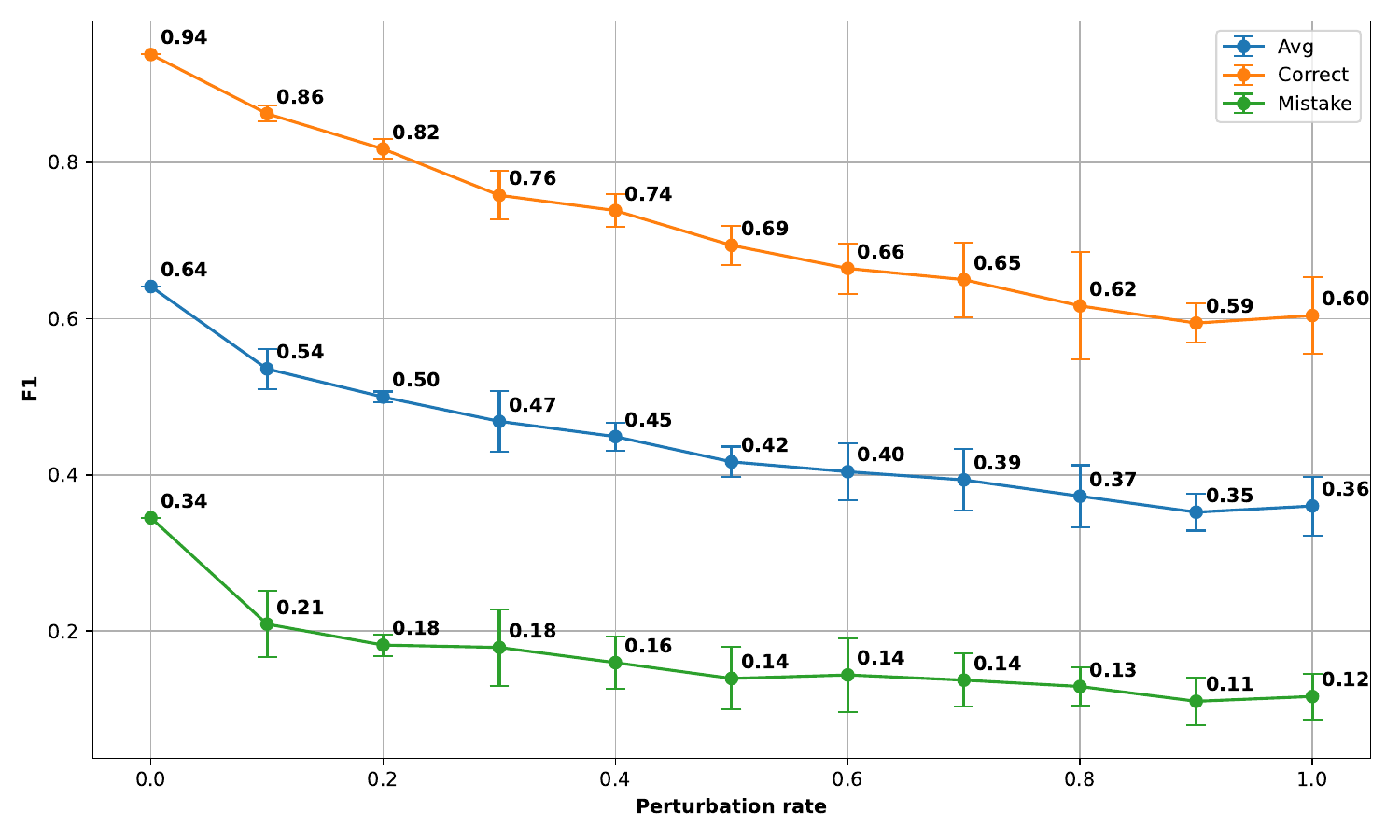}
  \label{fig:sub2}
\end{subfigure}
\caption{To further investigate the effect of noise, we conducted an analysis based on the controlled perturbation of ground truth action sequences, with the aim to simulate noise in the action detection process. At inference, we perturbed each key-step with a probability \(\alpha\) (the ``perturbation rate''), with three kinds of perturbations: insert (inserting a new key-step with a random action class), delete (deleting a key-step), or replace (randomly changing the class of a key-step). The plots show the trend of the F1 score (Average, Correct, and Mistake) as the perturbation rate increases in the case of Assembly101-O (left) and EPIC-Tent-O (right). Results suggest that the proposed approach can still bring benefits even in the presence of imperfect action detections, with the average F1 score dropping down $10-15$ points with a moderate noise level of $20\%$.}
\label{fig:test}
\end{figure}

The proposed approach requires the availability of key-step sequences, a common assumption of works addressing other video understanding tasks~\cite{caba2015activitynet, kay2017kinetics, jang2019epic, grauman2022ego4d, grauman2023ego}. While our method is applicable to any fully supervised video understanding dataset, future works should focus on overcoming such limitation and taking advantage of the vast amount of unlabeled video and textual data sets. While the proposed TGT method has shown promising results when trained directly on video features, the investigation of task graph learning in the absence of labeled key-step sequences is beyond the scope of this paper. We noted a reduced ability of our approach to work with noisy action sequences and a tendency to hallucinate pre-conditions, likely due to the limited expressivity of key-step sequences arising from videos showing the most common ways to perform a procedure. The performance of our designed system to detect mistakes is influenced by the quality of action recognition (see Figure~\ref{fig:test}). If the action recognition module fails to detect an action, the method may incorrectly signal a missing pre-condition.
Conversely, if an action is falsely detected as performed, the method may fail to signal an actual mistake. Future improvements in online action recognition will enhance the robustness of our method. Furthermore, our approach does not explicitly model ``optional'' key-steps, which can lead to incorrect error signaling if optional steps are treated as mandatory. This issue could potentially be addressed through the integration of specialized modules capable of detecting optional nodes.
Another limitation of task graph representations, both in this work and in prior approaches~\cite{ashutosh2024video, zhou2023procedure, sohn2020meta, jang2023multimodal}, is their inability to explicitly model repeatable key-steps. Recent advancements such as~\cite{grauman2023ego} have introduced a ``repeatable'' node attribute to task graphs, but this extension is based on manual annotations, and the automatic learning of such attributes from data remains an open problem. Despite this limitation, the proposed error detection model demonstrates an ability to handle cases where key-steps may recur (e.g., spreading peanut butter). At test time, pre-conditions of key-steps are verified via the predicted task graph, even if a key-step has appeared earlier in the sequence. Nevertheless, more effective modeling of repeatable key-steps, especially in contexts where specific repetitions are required (e.g., ``cut three slices of bread''), remains an important area for future research. Future work should explore methods for incorporating these requirements into task graph learning frameworks. Our method follows the setup of PREGO~\cite{flaborea2024prego}, which defines the Assembly101-O and EPIC-Tent-O datasets as curated versions of their originals to account for open-set procedural errors such as ``order'', ``omission'', ``correction'', and ``repetition'' mistakes. These are procedural mistakes, as distinguished from ``proficiency errors'' described in prior works~\cite{grauman2023ego}. The proposed method focuses on procedural mistakes at the abstract level of executed actions, and thus, would not be directly applicable to proficiency error detection. In real-world systems, this limitation could be mitigated by integrating subsystems that specialize in detecting different types of errors. Developing an integrated approach that addresses both procedural and proficiency errors is a promising direction for future research.

\section{Conclusion}
We considered the problem of learning task graph representations of procedures from video demonstrations. Framing task graph learning as a maximum likelihood estimation problem, we proposed a differentiable loss which allows direct optimization of the adjacency matrix through gradient descent and can be plugged into more complex neural network architectures. Experiments on three datasets show that the proposed approach can learn accurate task graphs, develop video understanding abilities, and improve the downstream task of online mistake detection surpassing state of the art methods. 
We release our code at the following URL: \url{https://github.com/fpv-iplab/Differentiable-Task-Graph-Learning}.

\section{Acknowledgments}
This research is supported in part by the PNRR PhD scholarship ``Digital Innovation: Models, Systems and Applications'' DM 118/2023, by the project Future Artificial Intelligence Research (FAIR) – PNRR MUR Cod. PE0000013 - CUP: E63C22001940006, and by the Research Program PIAno di inCEntivi per la Ricerca di Ateneo 2020/2022 — Linea di Intervento 3 ``Starting Grant'' EVIPORES Project - University of Catania.

We thank the authors of~\cite{flaborea2024prego} and in particular Alessandro Flaborea and Guido D'Amely for sharing the code to replicate experiments in the PREGO benchmark.
{
\small
\bibliographystyle{plain}
\bibliography{bibtex}

}

\newpage
\appendix

\begin{figure}
    \centering
    \includegraphics[width=0.5\linewidth]{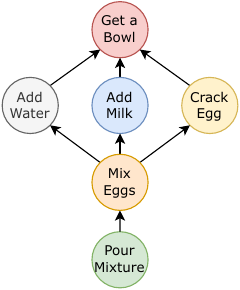}
    \caption{Example of a task graph where each node represents a key-step in the procedure, with directed edges indicating the necessary preconditions for each step.}
    \label{fig:example_of_task_graph}
\end{figure}
\section{Task Graph}
An example of a task graph is illustrated in Figure~\ref{fig:example_of_task_graph}. A task graph is a Directed Acyclic Graph (DAG) where nodes represent key-steps and directed edges impose a partial order on these steps, indicating the necessary preconditions for each node. For example, the key-step ``Mix'' has preconditions such as ``Add Water'', ``Add Milk'', and ``Crack Egg''. This formulation of task graphs is not a novel contribution of this paper but was originally introduced in~\cite{grauman2023ego}.

\section{Evaluation Measures}
This appendix details the evaluation measures used to assess performance experimentally for the two considered tasks of task graph generation and online mistake detection.

\paragraph{Task Graph Generation} Task graph generation is evaluated by comparing a generated graph $\hat G = (\hat{\mathcal{K}},\hat{\mathcal{A}})$ with a ground truth graph $G= (\mathcal{K},\mathcal{A})$. Since task graphs aim to encode ordering constraints between pairs of nodes, we evaluate task graph generation as the problem of identifying valid pre-conditions (hence valid graph edges) among all possible ones. We hence adopt classic detection evaluation measures such as precision, recall, and $F_1$ score. In this context, we define True Positives (TP) as all edges included in both the predicted and ground truth graph (Eq.~\eqref{eq:tp}), False Positives (FP) as all edges included in the predicted graph, but not in the ground truth graph (Eq.~\eqref{eq:fp}), and False Negatives (FN) as all edges included in the ground truth graph, but not in the predicted one (Eq.~\eqref{eq:tn}). Note that true negatives are not required to compute precision, recall and $F_1$ score.
\\
\begin{minipage}{0.33\linewidth}
\begin{equation}
TP=\hat{\mathcal{A}} \cap \mathcal{A}
\label{eq:tp}
\end{equation}    
\end{minipage}
\begin{minipage}{0.33\linewidth}
\begin{equation}
    FP=\hat{\mathcal{A}} \setminus \mathcal{A}
    \label{eq:fp}
\end{equation}
\end{minipage}
\begin{minipage}{0.33\linewidth}
\begin{equation}
    FN=\mathcal{A} \setminus \hat{\mathcal{A}}
    \label{eq:tn}
\end{equation}
\end{minipage}

\paragraph{Online Mistake Detection} We follow previous works on mistake detection from procedural egocentric videos~\cite{sener2022assembly101,wang2023holoassist,flaborea2024prego} and evaluate online mistake detection with standard precision, recall, and $F_1$ scores. We break down metrics by the ``correct'' and ``mistake'' classes, as well as report average values.

\section{Implementation Details}
\label{app:implementation_details}
This appendix provides implementation details to replicate the experiments discussed in Section~\ref{sec:experiments}.
\subsection{Data Augmentation}
In procedural tasks, it is common for certain actions to be repeated multiple times throughout the execution of a task. For example, in the EPIC-Tent dataset~\cite{jang2019epic}, an operation such as "reading the instructions" may be performed repeatedly at any point during the task. To model key-step orderings within the framework of topological sorts, our approach assumes that sequences should not contain such repetitions. Since repetitions denote that a specific action can appear at different stages of a procedure, we expand each sequence with repetitions to all distinct sequences obtained by dropping repeated actions.
This data augmentation strategy enhances the robustness of our model on Assembly101~\cite{sener2022assembly101} and EPIC-Tent~\cite{jang2019epic}, while it was not necessary for the CaptainCook4D dataset~\cite{peddi2023captaincook4d}, as sequences do not contain any repetitions.

\subsection{Early Stopping}
The learning process was conducted without the use of a validation set. To avoid overfitting and saving computation we defined a ``Sequence Accuracy (SA)'' score used to determine when the model reaches a learning plateau. We early stop models when an SA value of at least $0.95$ is reached, and if the model shows no SA improvement for $25$ consecutive epochs. The SA score is as follows:
\begin{equation}
     \text{SA} = \frac{1}{|\mathcal{Y}|} \sum_{y \in \mathcal{Y}} \frac{1}{|y|} \sum_{i=0}^{|y|-1} c(y_i, y[:i], pred(y_i))
\end{equation} 
where \( \mathcal{Y} \) defined sequences in the training set, \( y \) is a sequence from \( \mathcal{Y} \), \( y_i \) is the \( i \)-th element of sequence \( y \), \(y[:i]\) are the predecessors of the \(i\)-th element in the sequence $y$, and \( pred(y_i, Z) \) are the predicted predecessors for \( y_i \) from the current binarized adjacency matrix $Z$. The function \( c \) is defined as:
\begin{align}
     &c(y_i, y[:i], pred(y_i, Z)) = \begin{cases} 
       1 & \text{if } |y[:i]| = 0 \text{ and } |pred(y_i, Z)| = 0 \\
       \frac{|y[:i] \cap pred(y_i, Z)|}{|pred(y_i, Z)|} & \text{if } |y[:i]| > 0 \text{ and } |pred(y_i, Z)| > 0 \\
       0 & \text{otherwise}
    \end{cases}
 \end{align}
The SA score measures the compatibility of each sequence with the current task graph based on the ratio of correctly predicted predecessors of the current symbol $y_i$ of the sequence to the total number of predicted predecessors for $y_i$ in the current task graph.

\subsection{Hyperparameters}
Table~\ref{tab:hyperparameters1} details the hyperparameters employed in the experiments for task graph generation on the CaptainCook4D dataset~\cite{peddi2023captaincook4d}. During the training of TGT, we utilized a pre-trained EgoVLPv2~\cite{pramanick2023egovlpv2} on Ego-Exo4D~\cite{grauman2023ego} to extract text and video embeddings. The temperature value $T$ used in the cross-entropy distinctiveness loss was set to $0.9$ as in~\cite{radford2021learning}. The $\beta$ parameter was linearly annealed from an initial value of 1.0 to a final value of 0.05, with updates occurring every 100 epochs. This gradual decrease in $\beta$ mimics the warm-up strategy of~\cite{vaswani2017attention}, enabling smoother optimization early in training and leading to improved convergence as training progresses.

Table~\ref{tab:hyperparameters2} details the hyperparameters employed in the experiments for task graph generation on the Assembly101-O and EPIC-Tent-O datasets. For the downstream task of online mistake detection within the DO model framework, we extended the maximum training epochs to 1200, particularly for Assembly101-O. This change was necessary because, even after 1000 epochs, the model continued to exhibit many cycles among its 86 nodes. Extending the number of epochs allows the model additional time to learn and minimize these cycles, which is crucial given the complexity of the graph. 
In the TGT configuration, we reduced the dropout rate, while the $\beta$ parameter was gradually annealed from an initial value of 1.0 to 0.55 to prevent overfitting.

The reader is referred to the code for additional implementation details.

\begin{figure}
    \begin{minipage}[t]{0.49\textwidth}
    \captionof{table}{List of hyper-parameters used in the models training process for task graph generation using CaptainCook4D~\cite{peddi2023captaincook4d}.}
    \label{tab:hyperparameters1}
    \centering
    \begin{tabular}{lcc}
    \toprule
    & \multicolumn{2}{c}{Value} \\
    \cmidrule(lr){2-3}
    Hyper-parameter & DO & TGT \\
    \midrule
    Learning Rate & 0.1 & 0.000001 \\
    Max Epochs & 1000 & 3000\\
    Optimizer & Adam & Adam\\
    $\beta$ & 0.005 & 1.0 \(\sim\) 0.05\\
    Dropout Rate & - & 0.25\\
    \bottomrule
    \end{tabular}
    \end{minipage}
    \hspace{1.5mm}
    \begin{minipage}[t]{0.49\textwidth}
    \captionof{table}{List of hyper-parameters used in the models training process for task graph generation using Assembly101-O and EPIC-Tent-O.}
    \label{tab:hyperparameters2}
    \centering
    \begin{tabular}{lcc}
    \toprule
    & \multicolumn{2}{c}{Value} \\
    \cmidrule(lr){2-3}
    Hyper-parameter & DO & TGT \\
    \midrule
    Learning Rate & 0.1 & 0.000001 \\
    Max Epochs & 1200 & 1200\\
    Optimizer & Adam & Adam\\
    $\beta$ & 0.005 & 1.0 \(\sim\) 0.55\\
    Dropout Rate & - & 0.1\\
    \bottomrule
    \end{tabular}
    \end{minipage}
\end{figure}

\subsection{LLM Prompt}
Below is the prompt that was employed to instruct the model on its task, which involves identifying pre-conditions for given procedural steps.

\begin{verbatim}
I would like you to learn to answer questions by telling me the steps 
that need to be performed before a given one.

The questions refer to procedural activities and these are of the following type:

Q - Which of the following key steps is a pre-condition for the current key step 
"add brownie mix"?

- add oil
- add water
- break eggs
- mix all the contents
- mix eggs
- pour the mixture in the tray
- spray oil on the tray
- None of the above

Your task is to use your immense knowledge and your immense ability to tell me 
which preconditions are among those listed that must necessarily be carried out 
before the key step indicated in quotes in the question.

You have to give me the answers and a very brief explanation of why you chose them.

Provide the correct preconditions answer inside a JSON format like this:

{
   "add brownie mix": ["add oil", "add water", "break eggs"]
}
\end{verbatim}

\subsection{Data Split}
\label{app:split}
The CaptainCook4D dataset~\cite{peddi2023captaincook4d} comprises various error types, including order errors, timing errors, temperature errors, preparation errors, missing steps errors, measurement errors, and technique errors. Of these, missing steps and order errors directly impact the sequence integrity. Consequently, for our task graph generation, we utilized only those sequences of actions free from these specific types of errors. Table~\ref{tab:data_split} shows statistics on the CaptainCook4D subsets used for task graph generation.

For Online Mistake Detection, we considered the datasets defined by the authors of PREGO~\cite{flaborea2024prego}.

In the context of pairwise ordering and forecasting, we employed the subset of the CaptainCook4D dataset designated for task graph generation (refer to Table~\ref{tab:data_split}) and divided it into training and testing sets. This division was carefully managed to ensure that 50\% of the scenarios were equally represented in both the training and testing sets.

\begin{table}
\centering
\caption{A detailed breakdown of the data used from the CaptainCook4D dataset~\cite{peddi2023captaincook4d} for the task graph generation. This table categorizes each scenario by the number of videos, segments, and total duration in hours. The ``Total'' row aggregates the dataset characteristics.}
\label{tab:data_split}
\begin{tabular}{cccc}
\toprule
Scenario & Videos & Segments & Duration \\
\midrule
Microwave Egg Sandwich & 5 & 60 & 0.9h\\
Dressed Up Meatballs & 8 & 128 & 2.7h \\
Microwave Mug Pizza & 6 & 84 & 1.2h \\
Ramen & 11 & 165 & 2.7h \\
Coffee & 9 & 144 & 2.2h \\
Breakfast Burritos & 8 & 88 & 1.5h \\
Spiced Hot Chocolate & 7 & 49 & 0.9h \\
Microwave French Toast & 11 & 121 & 2.2h \\
Pinwheels & 5 & 95 & 0.8h \\
Tomato Mozzarella Salad & 13 & 117 & 1.3h \\
Butter Corn Cup & 5 & 60 & 1.4h \\
Tomato Chutney & 5 & 95 & 2.6h \\
Scrambled Eggs & 6 & 138 & 2.6h \\
Cucumber Raita & 12 & 132 & 2.7h \\
Zoodles & 6 & 78 & 1.1h \\
Sauted Mushrooms & 7 & 126 & 2.9h \\
Blender Banana Pancakes & 10 & 140 & 2.4h \\
Herb Omelet with Fried Tomatoes & 8 & 120 & 2.4h \\
Broccoli Stir Fry & 10 & 250 & 5.2h \\
Pan Fried Tofu & 9 & 171 & 3.6h \\
Mug Cake & 9 & 180 & 3.0h \\
Cheese Pimiento & 7 & 77 & 1.6h \\
Spicy Tuna Avocado Wraps & 9 & 153 & 2.6h \\
Caprese Bruschetta & 8 & 88 & 2.4h \\
\rowcolor{teal!30} Total & 194 & 2859 & 53.0h\\
\bottomrule
\end{tabular}
\end{table}

\subsection{Pairwise ordering and future prediction}
We setup the pairwise ordering and future prediction video understanding tasks following~\cite{zhou2015temporal}. 

\paragraph{Pairwise Ordering} Models take as input two randomly shuffled video clips and are tasked with recognizing the correct ordering between key-steps. We sample all consecutive triplets of labeled segments from test videos, discard the middle one, and consider the first and third ones as input pair. We evaluate models using accuracy.

\paragraph{Future Prediction} Models take as input an anchor video clip and two randomly shuffled video clips and are tasked to select which of the two clips is the correct future of the anchor clip. We sample all consecutive triplets of labeled segments from test videos and consider the middle clip as the anchor and the remaining two clips as the two options. We evaluate models using accuracy.

\paragraph{Model} We trained our TGT model using video embeddings extracted with a pre-trained EgoVLPv2~\cite{pramanick2023egovlpv2} on Ego-Exo4D~\cite{grauman2023ego}. During the training process, if multiple video embeddings are associated with the same key-step across the training sequences, one embedding per key-step is randomly selected. The model is trained for task graph generation on the training video and tested for pairwise ordering and future prediction on the test set. 

For pairwise ordering, we feed our model with two clips and obtain a $4 \times 4$ adjacency matrix, where the nodes represent \textit{START}, $A$, $B$, \textit{END}.
We establish the order between $A$ and $B$ based on the fulfillment of at least one of the following conditions: (a) if the weight of the edge $A \rightarrow B$ is greater than the weight of the edge $B \rightarrow A$, we conclude that $A$ precedes $B$; (b) by analyzing the sequences $<\text{START}, A, B, \text{END}>$ and $<\text{START}, B, A, \text{END}>$, we calculate their probabilities using Eq.~\eqref{eq:factorization_2}. If $P(<\text{START}, A, B, \text{END}> \mid Z)$ is greater than $P(<\text{START}, B, A, \text{END}> \mid Z)$, we infer that $A$ precedes $B$; (c) if the weight of the edge \textit{END} $\rightarrow B$ is greater than that of \textit{END} $\rightarrow A$, it implies that $B$ is a necessary precondition for concluding the procedure, indicating that $B$ follows $A$, and consequently, $A$ precedes $B$. If none of these conditions hold, we determine that $B$ precedes $A$.

For future prediction, we feed three clips and obtain a $5 \times 5$ adjacency matrix, where the nodes represent \textit{START}, $A$, $anchor$, $B$, and \textit{END}. We hence inspect the weights of edges $anchor \to A$ and $anchor \to B$ and choose as the future clip, the one related to the smallest weight (a small weight indicates that the selected clip is not a precondition). Another method to determine the future clip is by calculating the probabilities of the sequences $<\text{START}, A, anchor, B, \text{END}>$ and $<\text{START}, B, anchor, A, \text{END}>$ using Eq.~\eqref{eq:factorization_2}. If \( P(<\text{START}, A, anchor, B, \text{END}> \mid Z) \) is greater than \( P(<\text{START}, B, anchor, A, \text{END}> \mid Z) \), we infer that the sequence involving $A$ before $B$ is more probable, indicating that $B$ is the future clip for $anchor$. Conversely, if the probability of the second sequence is greater, then $A$ is deemed the future clip for $anchor$.

\subsection{Scalability of Task Graph Transformer (TGT)}

The Direct Optimization (DO) approach requires a separate training session for each procedure. Task Graph Transformer (TGT) offers more flexibility by allowing different sets of key-step embeddings at each forward pass, ideally enhancing scalability. Leveraging this capability, we trained a single TGT text model for all CaptainCook4D procedures. This was achievable due to TGT's ability to handle varying embeddings per forward pass, enabling simultaneous optimization across multiple procedures during training. As shown in Table~\ref{tab:scalability1}, the confidence intervals of both the single and unified models highlight some performance variance. The unified model exhibits lower average precision, recall, and F1 scores compared to the individually trained models, with a larger confidence interval. However, the results suggest that TGT-text models can still generalize across diverse procedures, reducing training complexity while maintaining reasonable performance.

\begin{table}[t]
    \centering
    \caption{Performance comparison between the single TGT-text model trained across all CaptainCook4D procedures and the unified model. The confidence intervals in the single models indicate that the unified method performs comparably to training individual models for each procedure.}
    \begin{tabular}{llll}
        \toprule
        Method         & Precision     & Recall & F\(_1\)  \\
        \midrule
        TGT-text (single) & 79.9 \scriptsize \(\pm 8.8\) & 81.9 \scriptsize \(\pm 6.9\) & 80.8 \scriptsize \(\pm 8.0\) \\
        TGT-text (unified) & 61.5 \scriptsize \(\pm 12.0\) & 68.2 \scriptsize \(\pm 10.3\) & 64.5 \scriptsize \(\pm 11.9\) \\
        \bottomrule
    \end{tabular}
    \label{tab:scalability1}
\end{table}

\subsection{Graph Post-processing}
\begin{figure}[t]
    \centering
    \includegraphics[width=0.5\linewidth]{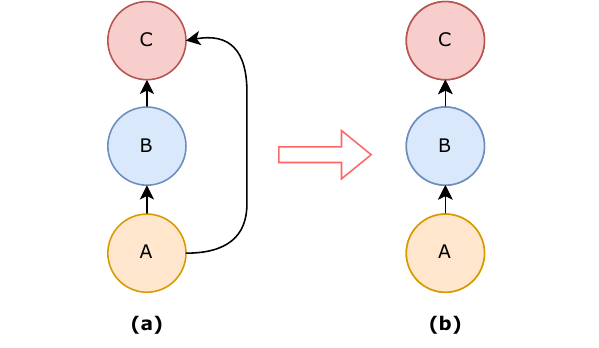}
    \caption{An example of transitive dependency between nodes. In (a) node A depends on B and C, but B depends on C, in this case, we can remove the edge between A and C for transitivity and we obtain the graph in (b).}
    \label{fig:transitive_dependency}
\end{figure}

We binarize the adjacency matrix with the threshold $\frac{1}{n}$, where $n$ is the number of nodes. After this thresholding phase, it is possible to encounter situations like the one illustrated in Figure~\ref{fig:transitive_dependency}, where node A depends on nodes B and C, and node B depends on node C. Due to the transitivity of the pre-conditions, we can remove the edge connecting node A to node C, as node B must precede node A. Sometimes, it may occur that a node does not serve as a pre-condition for any other node; in such cases, the END node should be directly connected to this node. Conversely, if a node has no pre-conditions, an edge is added from the current node to the START node. 

At the end of the training process, obtaining a graph containing cycles is also possible. In such cases, all cycles within the graph are considered, and the edge with the lowest score within each cycle is removed. This method ensures that the graph remains a Directed Acyclic Graph (DAG).

\subsection{Details on Online Mistake Detection}
Given the noisy sequences in Assembly101~\cite{sener2022assembly101} and EPIC-Tent~\cite{jang2019epic}, a distinct approach was adopted during the post-processing phase of task graph generation. Specifically, if a key-step in the task graph has only two pre-conditions and one is the START node, the other pre-condition will be removed regardless of its score, otherwise we apply the transitivity dependences reduction aforementioned. This approach allows for a graph with fewer pre-conditions in the initial steps.

In the case of Assembly101, which includes multiple procedural tasks, we opted to consider a single task graph that summarizes all the procedures, rather than generating individual graphs for each.

\subsection{Qualitative Examples}
\begin{figure}[H]
\centering
\begin{subfigure}{.49\textwidth}
  \centering
  \includegraphics[width=\linewidth]{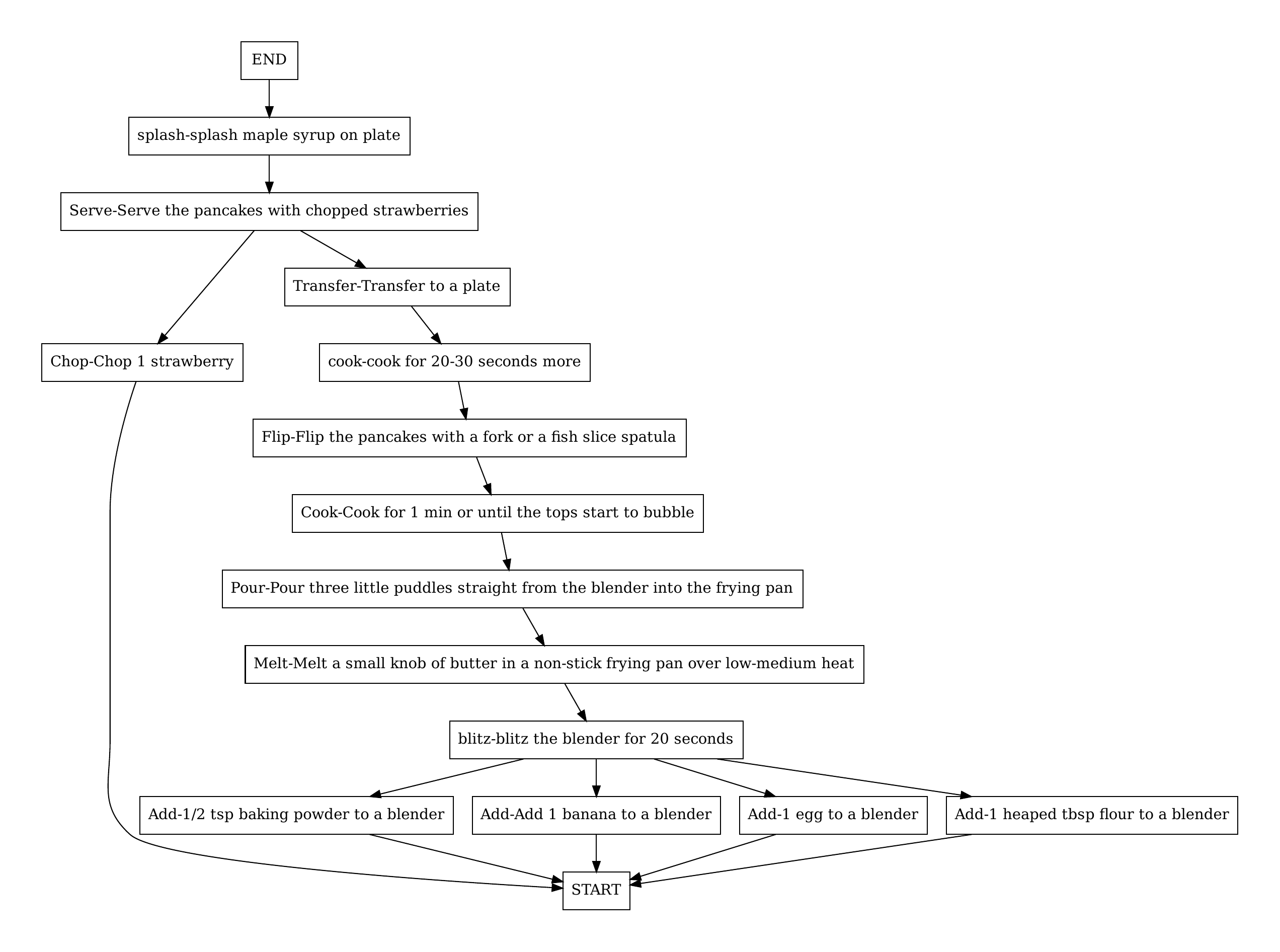}
  \caption{}
\end{subfigure}
\begin{subfigure}{.49\textwidth}
  \centering
  \includegraphics[width=\linewidth]{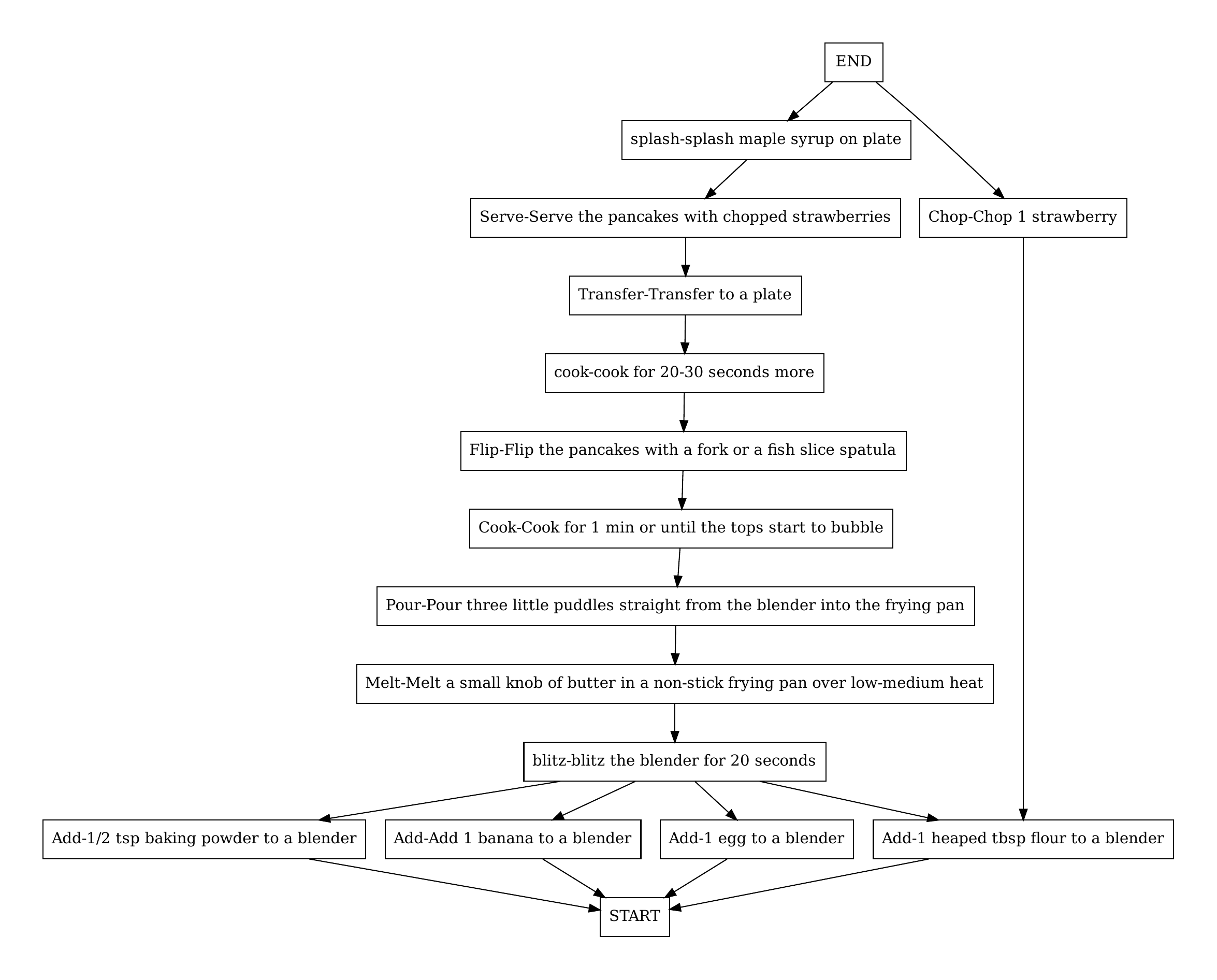}
  \caption{}
\end{subfigure}
\caption{(a) Ground truth task graph and (b) predicted task graph of the scenario Breakfast Burritos.}
\label{fig:qualitative1}
\end{figure}

\begin{figure}[H]
\centering
\begin{subfigure}{.49\textwidth}
  \centering
  \includegraphics[width=\linewidth]{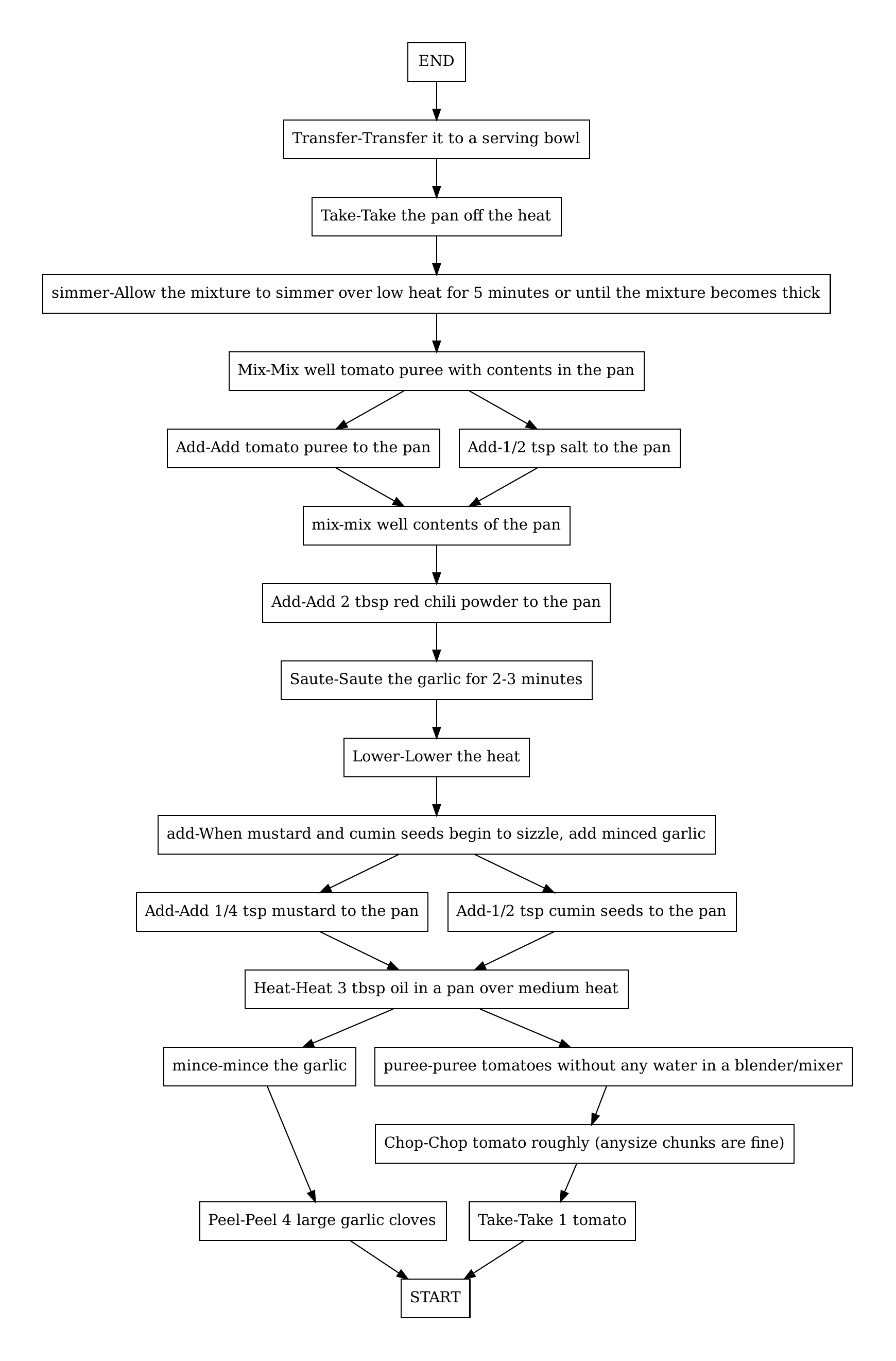}
  \caption{}
\end{subfigure}
\begin{subfigure}{.49\textwidth}
  \centering
  \includegraphics[width=\linewidth]{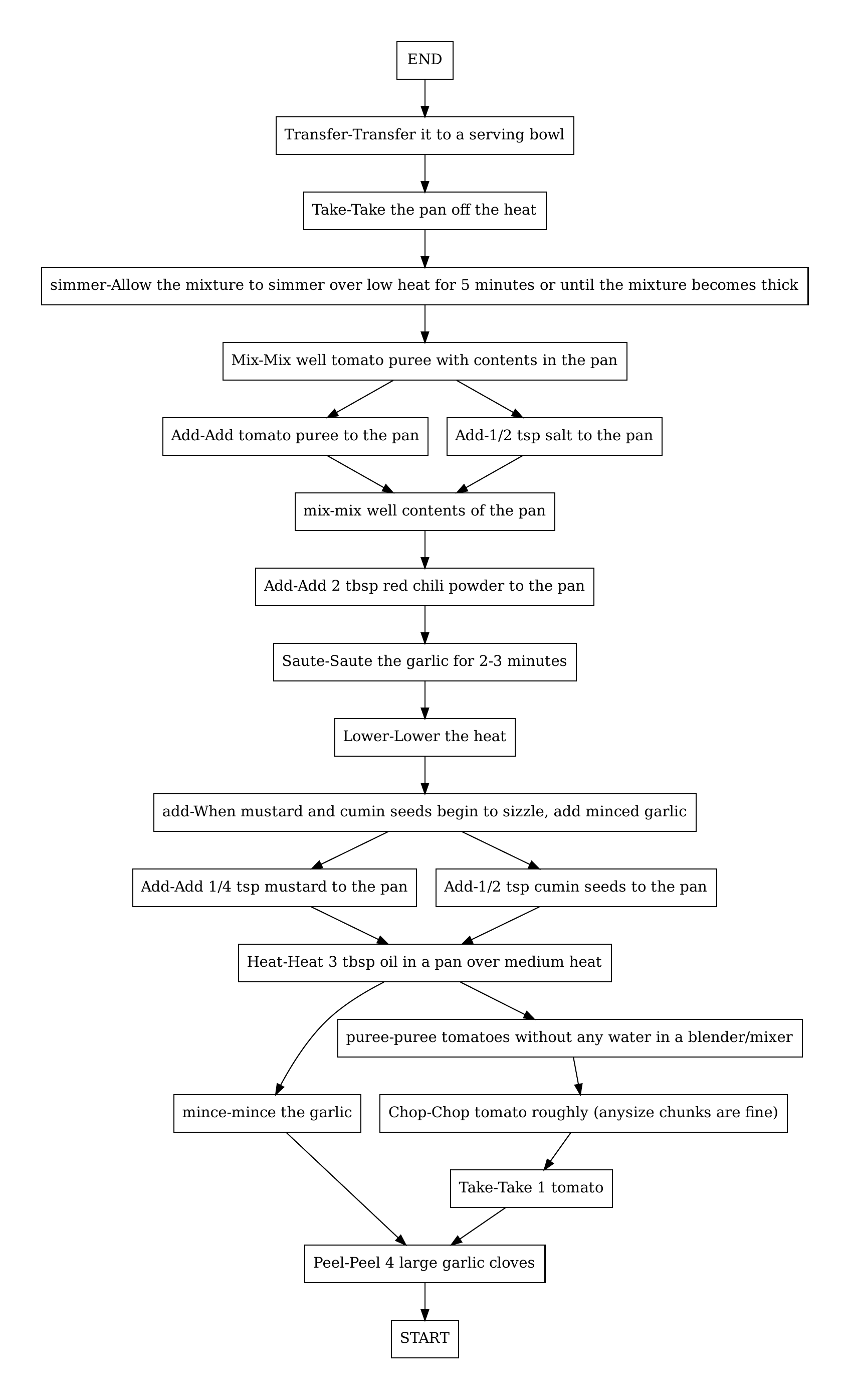}
  \caption{}
\end{subfigure}
\caption{(a) Ground truth task graph and (b) predicted task graph of the scenario Tomato Chutney.}
\label{fig:qualitative24}
\end{figure}

Figures~\ref{fig:qualitative1} -~\ref{fig:qualitative24} report qualitative examples of prediction using our Direct Optimization (DO) method on the procedures of CaptainCook4D. The task graphs must be read in a bottom-up manner, where the START node (bottom) is at the lowest position and represents the first node with no preconditions, while the END node (up) is the final step of the procedure.

\begin{figure}[t]
    \centering
    \includegraphics[width=1\linewidth]{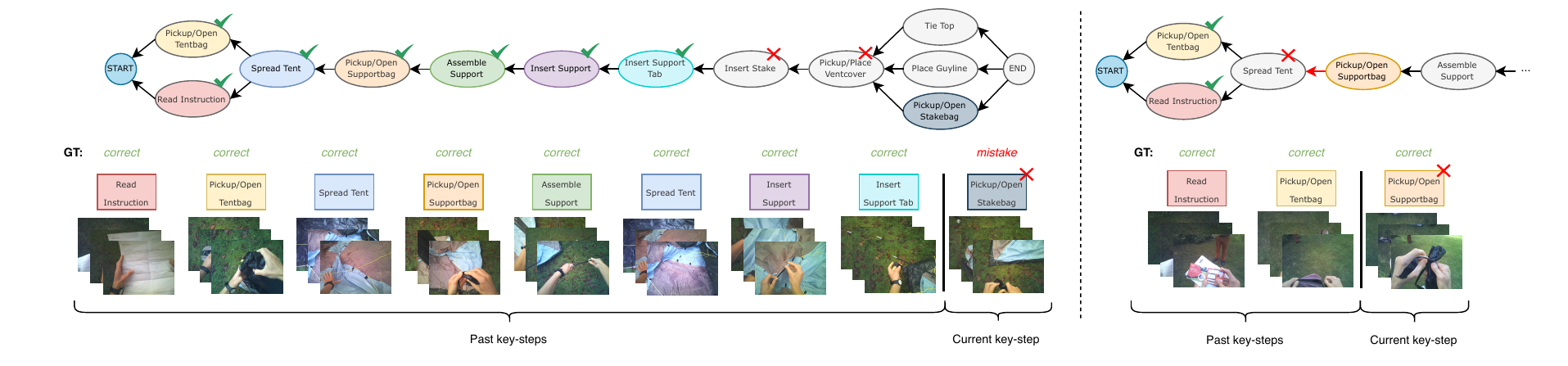}
    \caption{A success (left) and failure (right) case on EPIC-Tent-O. Past key-steps' colors match nodes' colors. On the left, the current key-step ``Pickup/Open Stakebag'' is correctly evaluated as a mistake because the step ``Pickup/Place Ventcover'' is a precondition of the current key-step, but it is not included among the previous key-steps. On the right, ``Pickup/Open Supportbag'' is incorrectly evaluated as mistake because the step ``Spread Tent'' is precondition of the current key-step, but it is not included among the previous key-steps. This is due to the fact that our method wrongly predicted ``Spread Tent'' as a pre-condition of ``Pickup/Open Supportbag'', probably due to the two actions often occurring in this order.}
    \label{fig:qualitative_mistake}
\end{figure}

Figure~\ref{fig:qualitative_mistake} reports a qualitative analysis of the generated task graph for detecting the mistakes on EPIC-Tent-O.

\subsection{Experiments Compute Resources}
\label{app:resources}
The experiments involving the training of the DO model on symbolic data from the CaptainCook4D dataset proved to be highly efficient. We were able to generate all the task graphs in approximately half an hour using a Tesla V100S-PCI GPU. This GPU allowed us to run up to 8 training processes simultaneously. In contrast, training the TGT models for all scenarios in the CaptainCook4D dataset required about 24 hours, with the same GPU supporting the concurrent training of up to 2 models. Additionally, once the task graphs were obtained, executing the PREGO benchmarks for mistake detection was significantly faster, requiring online action prediction, which could be performed in real-time on a Tesla V100S-PCI GPU.

\section{Societal Impact}
\label{sec:societal}
Reconstructing task graphs from procedural videos may enable the construction of agents able to assist users during the execution of the task. Learning task graphs from videos may be affected by geographical or cultural biases appearing in the data (e.g., specific ways of performing given tasks), which may limit the quality of the feedback returned to the user, potentially leading to harm. We expect that training data of sufficient quality should limit such risks.

\begin{figure}[H]
\centering
\begin{subfigure}{.49\textwidth}
  \centering
  \includegraphics[width=\linewidth]{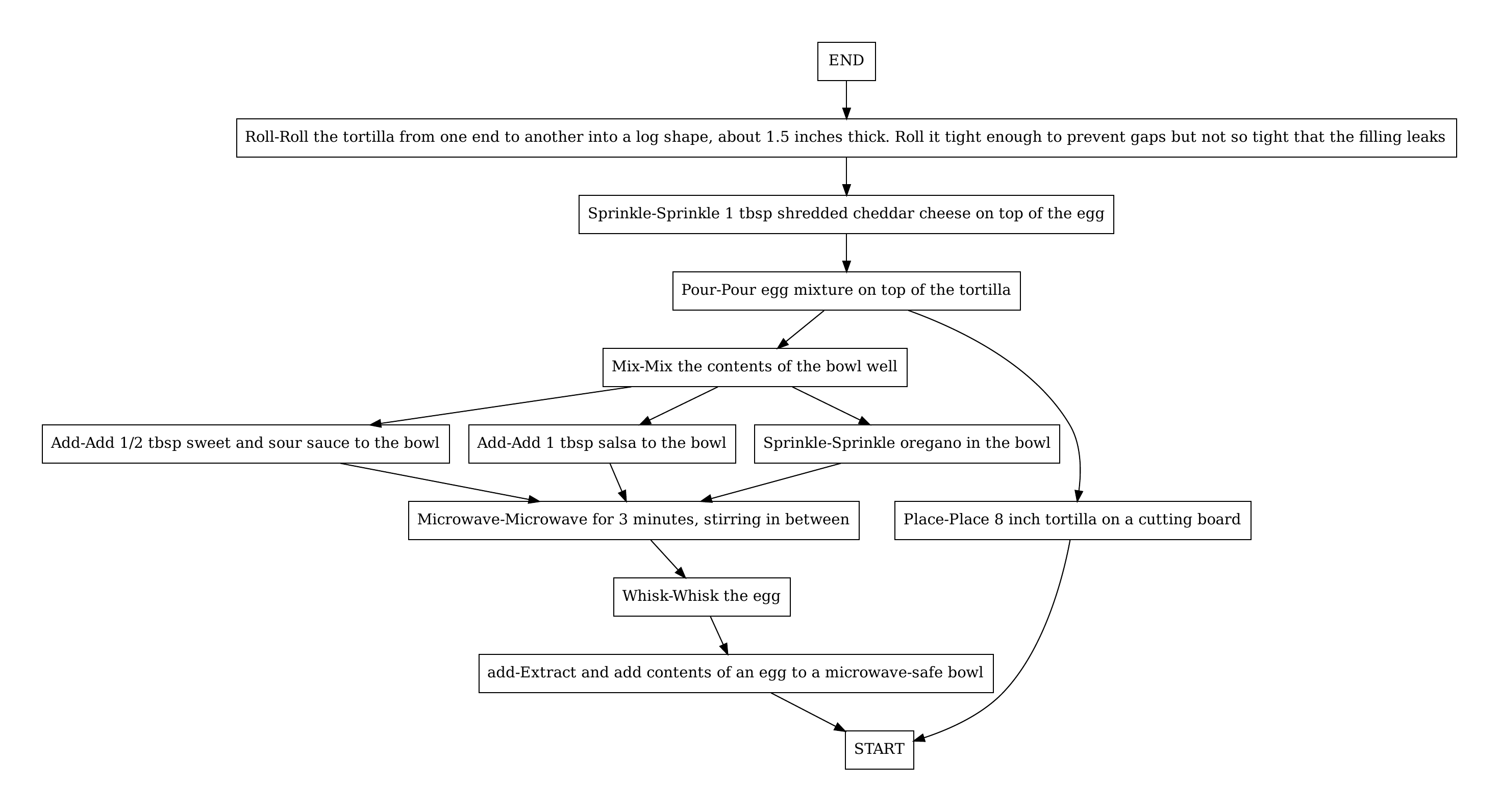}
  \caption{}
\end{subfigure}
\begin{subfigure}{.49\textwidth}
  \centering
  \includegraphics[width=\linewidth]{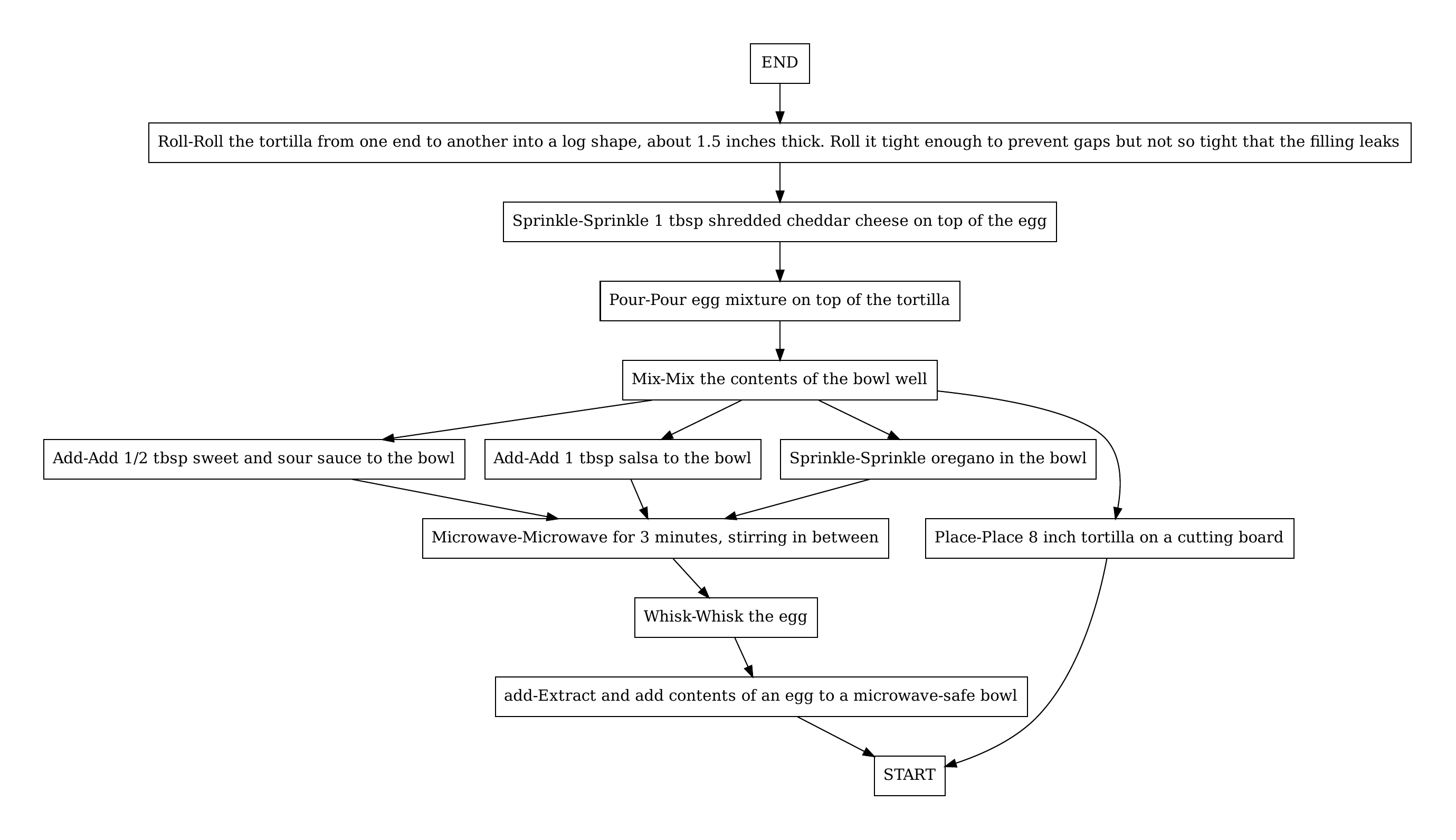}
  \caption{}
\end{subfigure}
\caption{(a) Ground truth task graph and (b) predicted task graph of the scenario Breakfast Burritos.}
\end{figure}

\begin{figure}[H]
\centering
\begin{subfigure}{.49\textwidth}
  \centering
  \includegraphics[width=\linewidth]{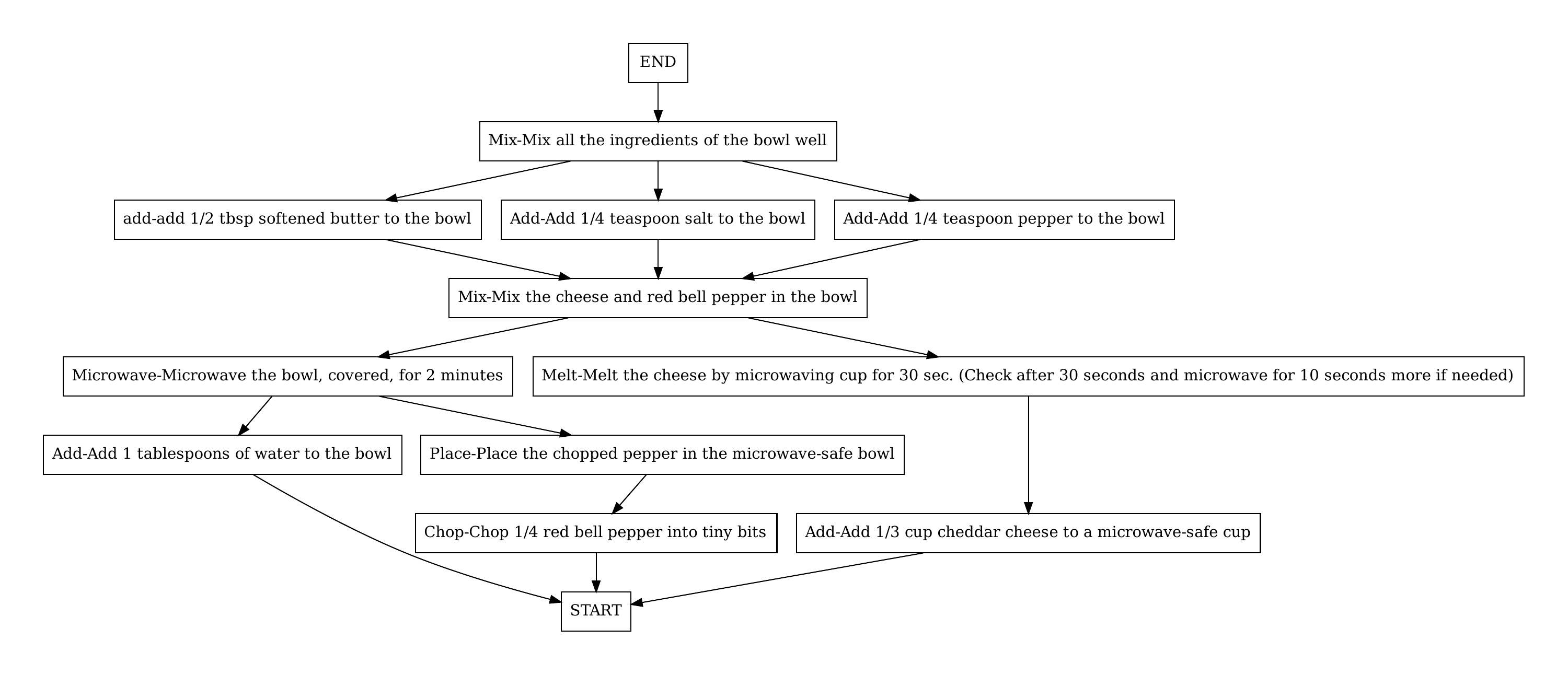}
  \caption{}
\end{subfigure}
\begin{subfigure}{.49\textwidth}
  \centering
  \includegraphics[width=\linewidth]{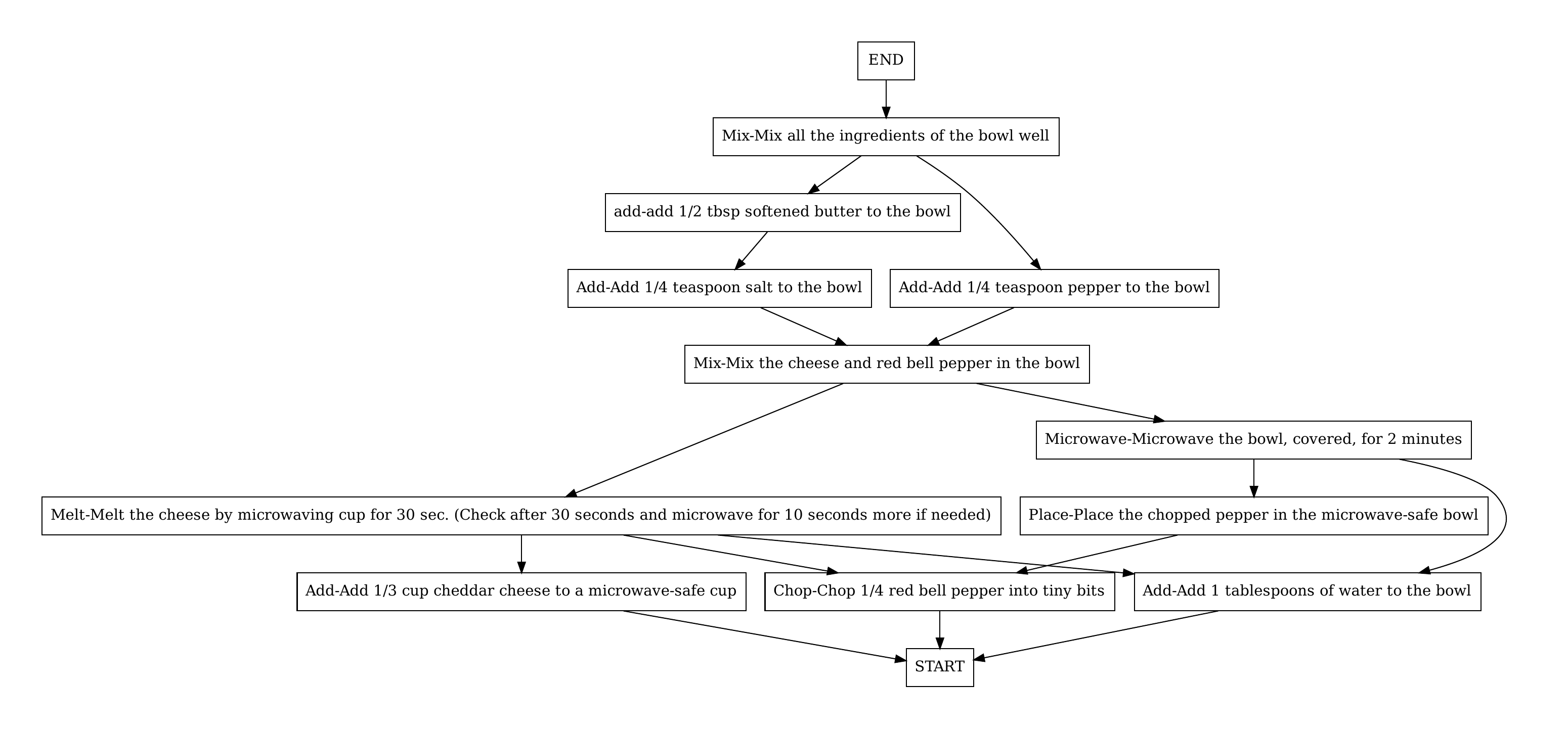}
  \caption{}
\end{subfigure}
\caption{(a) Ground truth task graph and (b) predicted task graph of the scenario Cheese Pimiento.}
\end{figure}

\begin{figure}[H]
\centering
\begin{subfigure}{.49\textwidth}
  \centering
  \includegraphics[width=\linewidth]{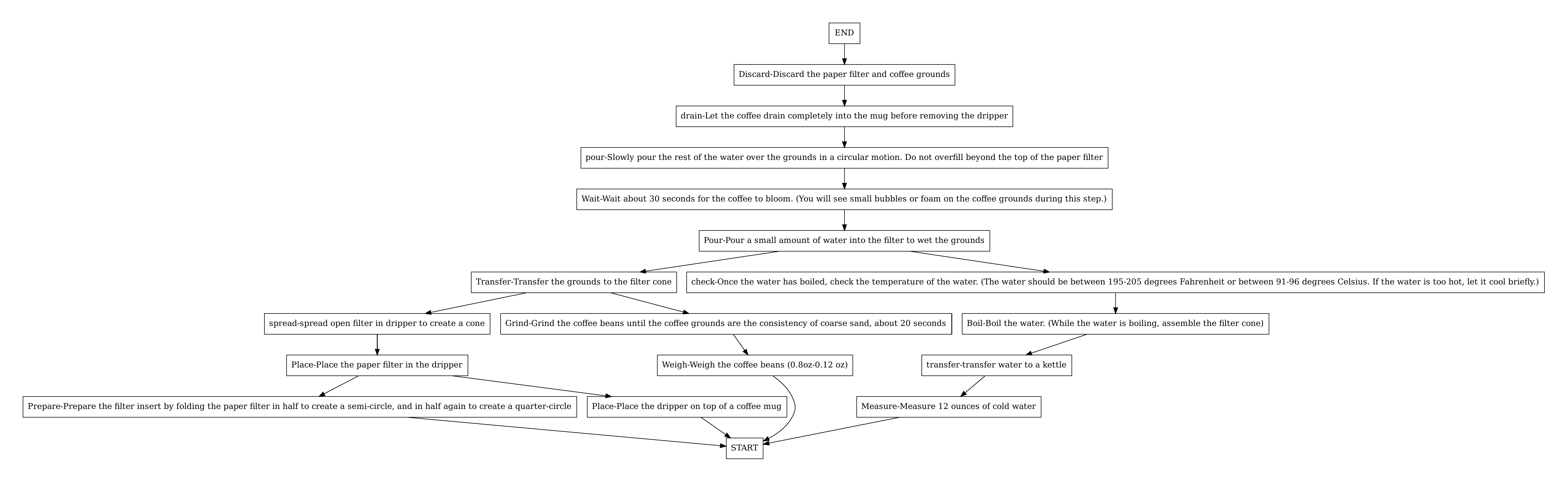}
  \caption{}
\end{subfigure}
\begin{subfigure}{.49\textwidth}
  \centering
  \includegraphics[width=\linewidth]{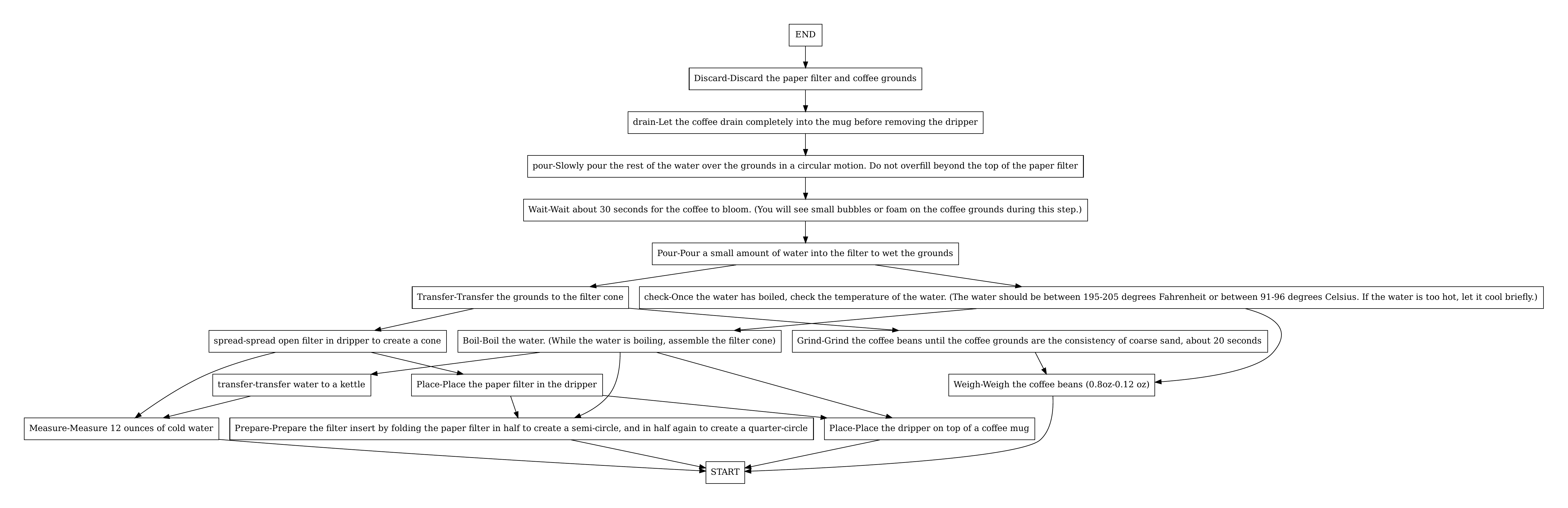}
  \caption{}
\end{subfigure}
\caption{(a) Ground truth task graph and (b) predicted task graph of the scenario Coffee.}
\end{figure}

\begin{figure}[H]
\centering
\begin{subfigure}{.49\textwidth}
  \centering
  \includegraphics[width=\linewidth]{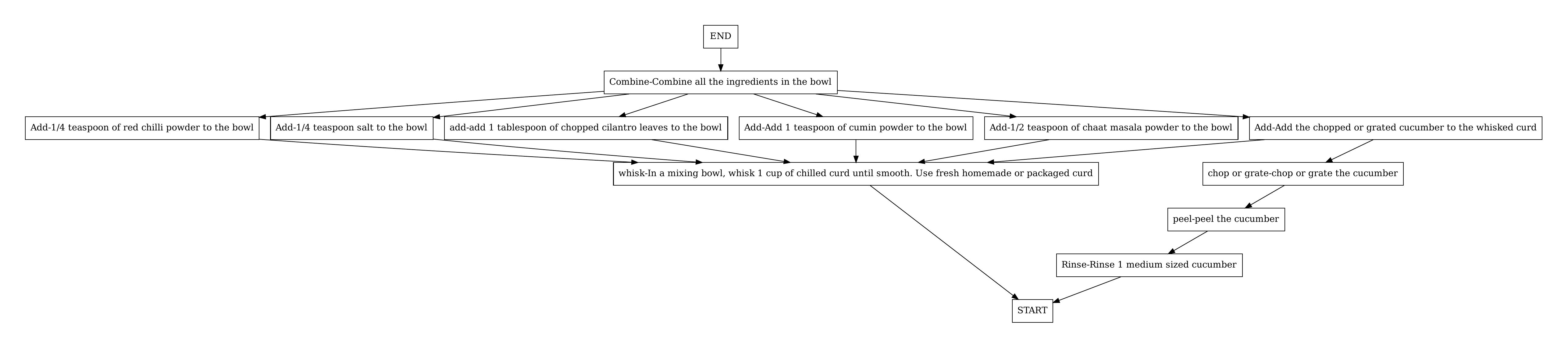}
  \caption{}
\end{subfigure}
\begin{subfigure}{.49\textwidth}
  \centering
  \includegraphics[width=\linewidth]{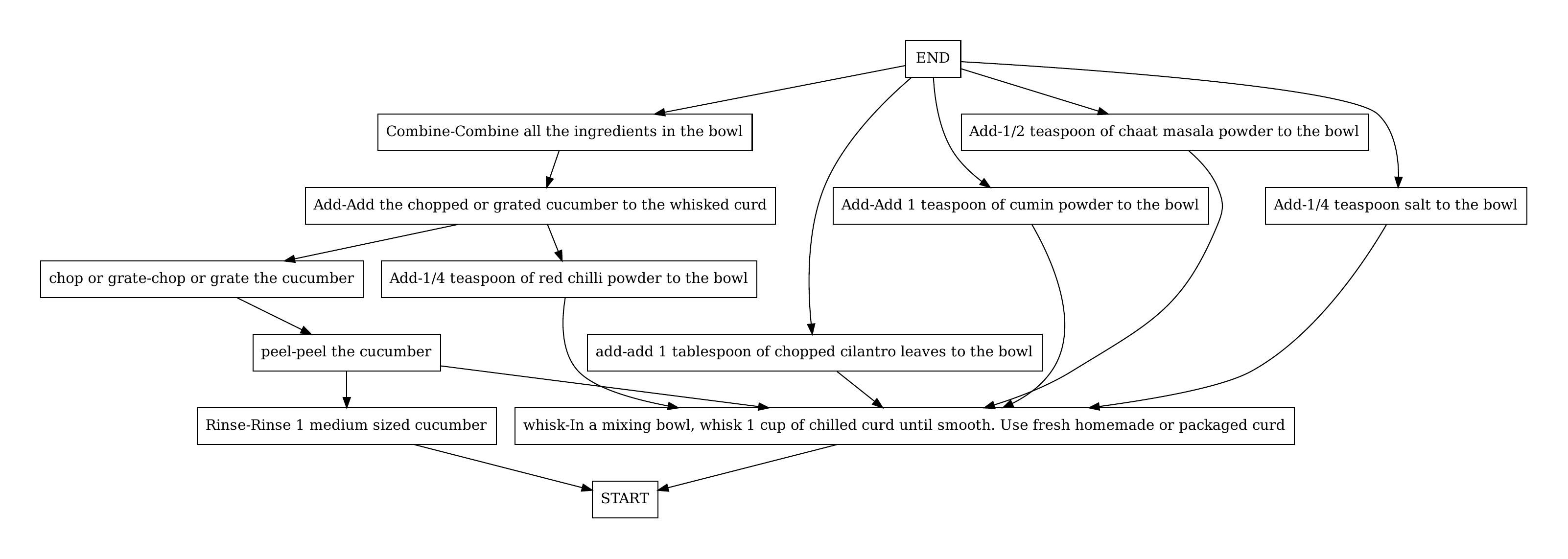}
  \caption{}
\end{subfigure}
\caption{(a) Ground truth task graph and (b) predicted task graph of the scenario Cucumber Raita.}
\end{figure}

\begin{figure}[H]
\centering
\begin{subfigure}{.49\textwidth}
  \centering
  \includegraphics[width=\linewidth]{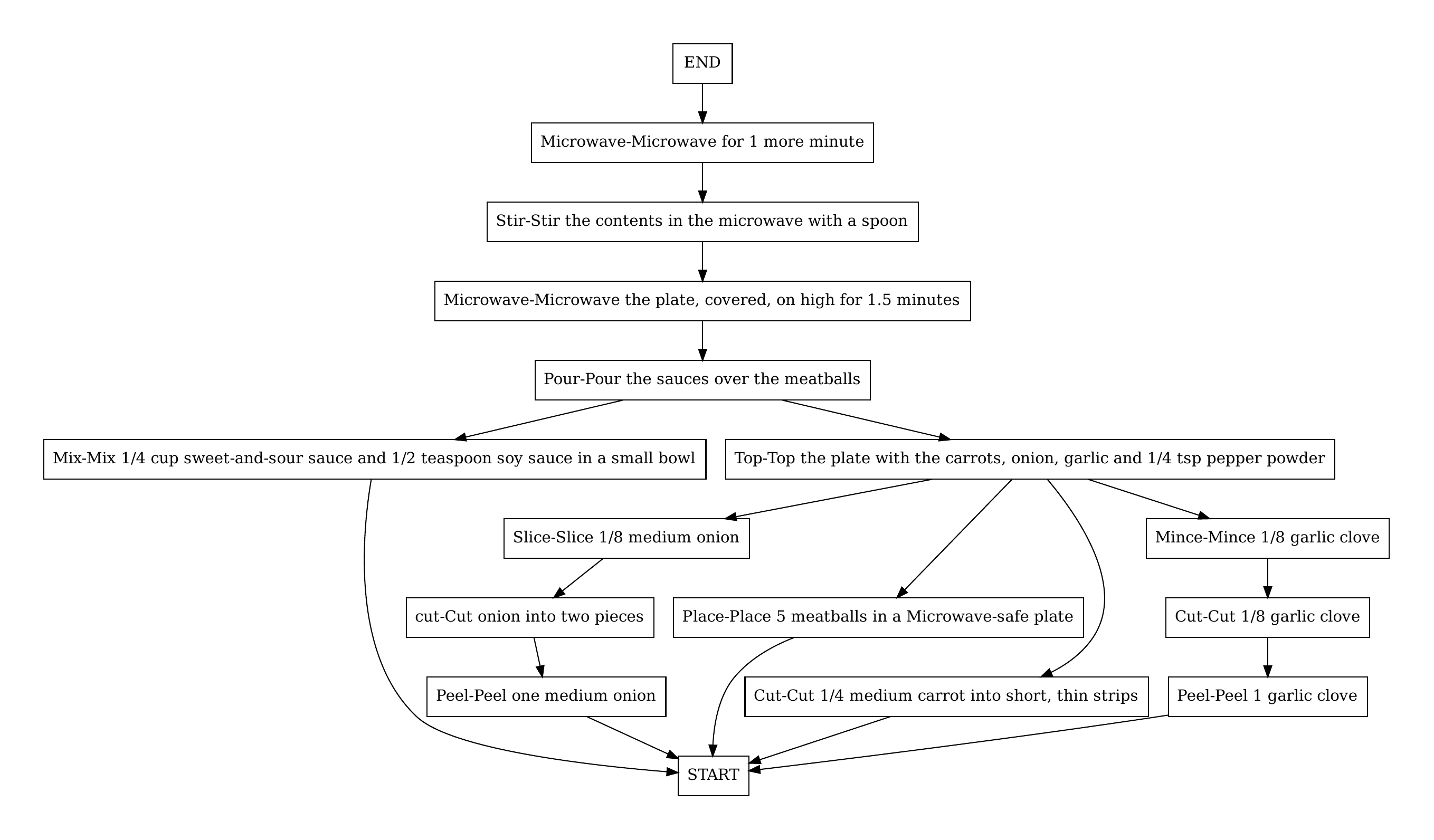}
  \caption{}
\end{subfigure}
\begin{subfigure}{.49\textwidth}
  \centering
  \includegraphics[width=\linewidth]{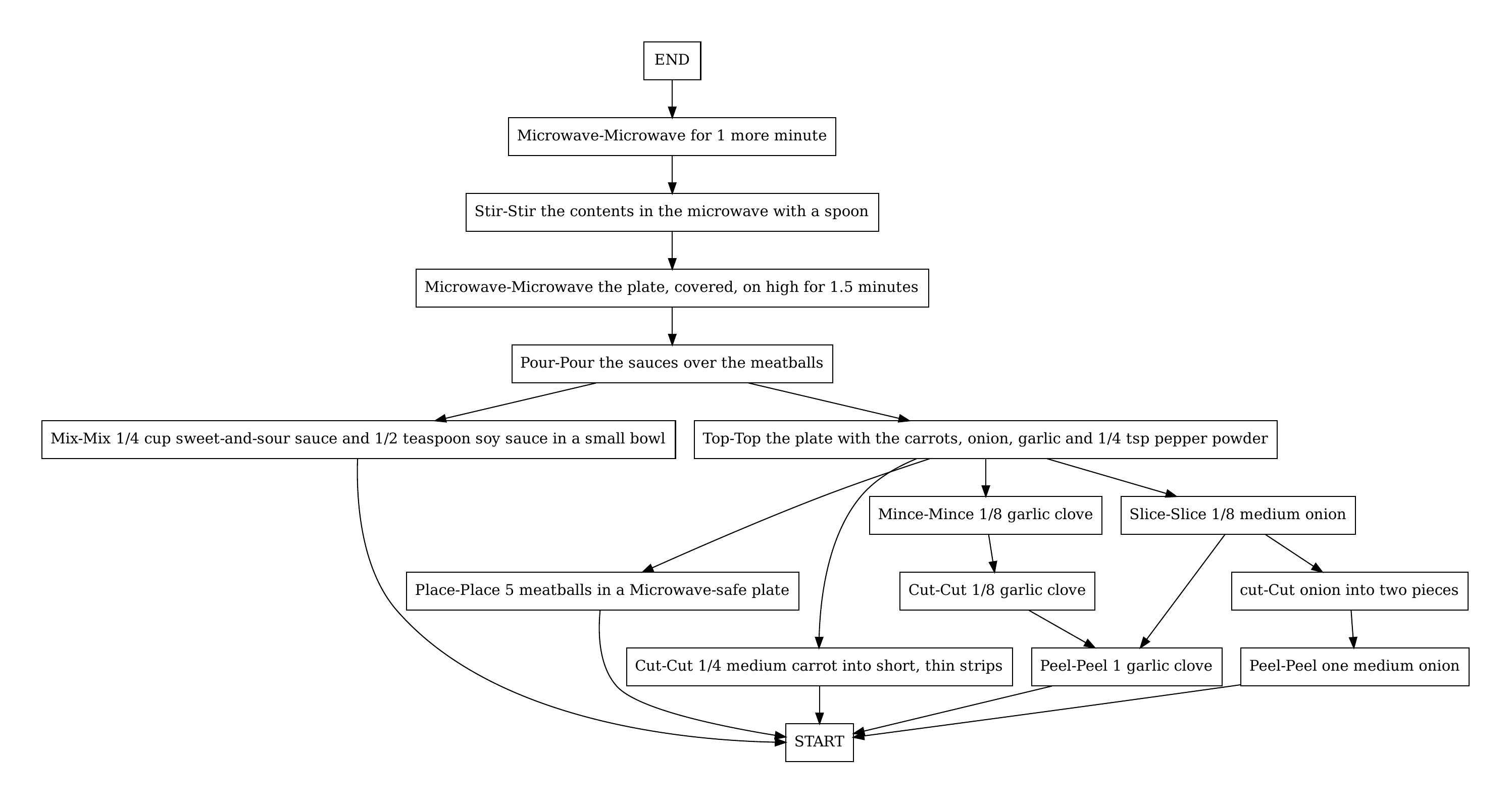}
  \caption{}
\end{subfigure}
\caption{(a) Ground truth task graph and (b) predicted task graph of the scenario Dressed Up Meatballs.}
\end{figure}

\begin{figure}[H]
\centering
\begin{subfigure}{.49\textwidth}
  \centering
  \includegraphics[width=\linewidth]{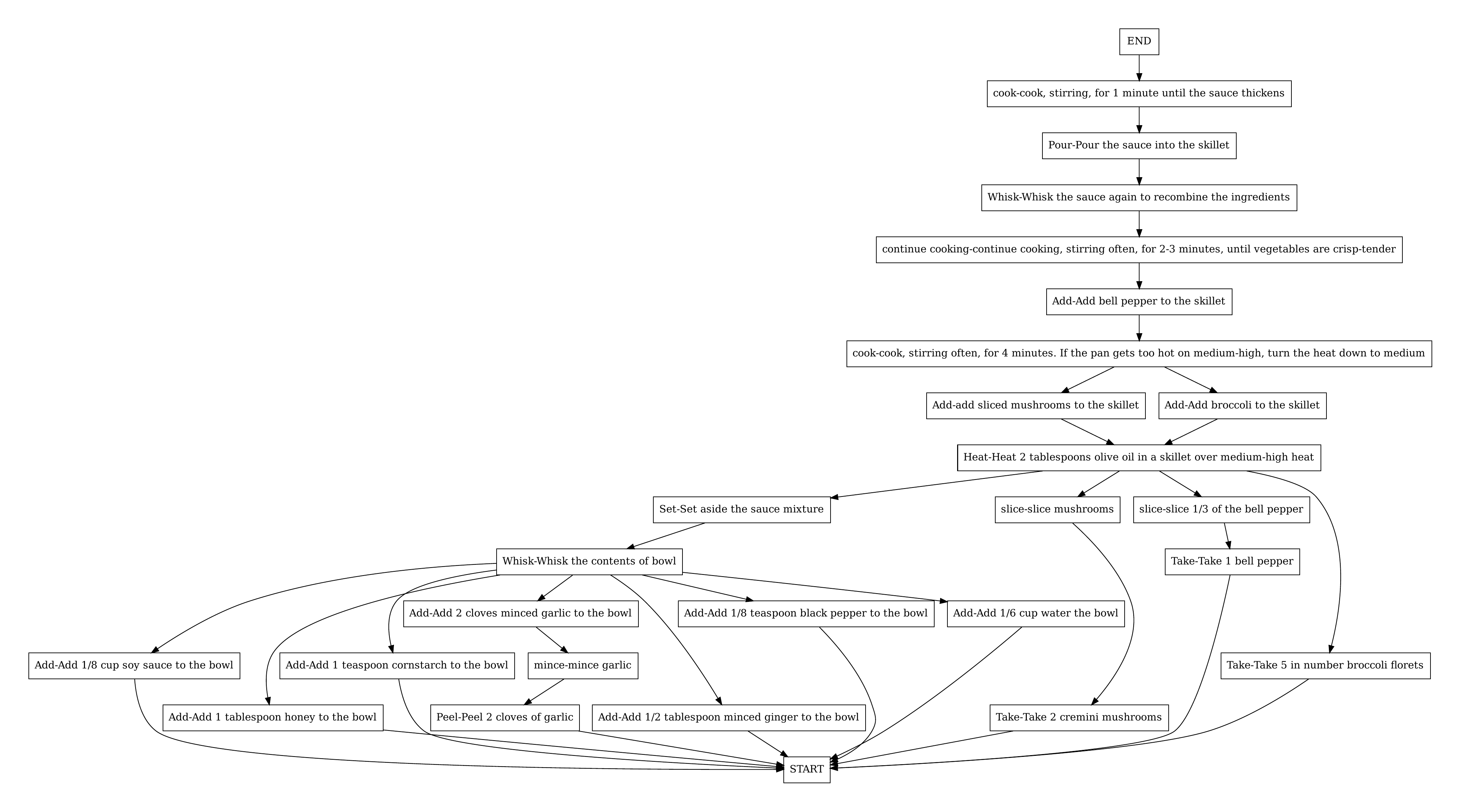}
  \caption{}
\end{subfigure}
\begin{subfigure}{.49\textwidth}
  \centering
  \includegraphics[width=\linewidth]{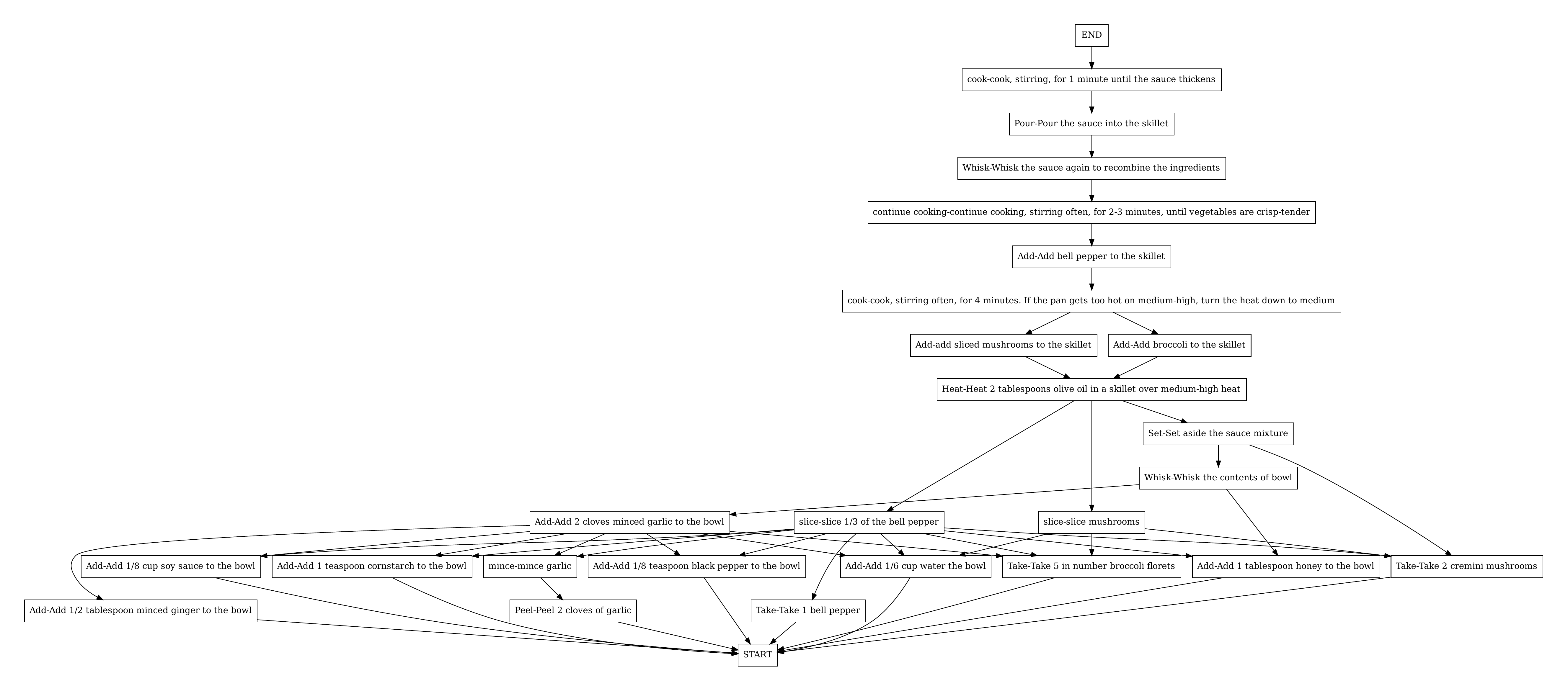}
  \caption{}
\end{subfigure}
\caption{(a) Ground truth task graph and (b) predicted task graph of the scenario Broccoli Stir Fry.}
\end{figure}

\begin{figure}[H]
\centering
\begin{subfigure}{.49\textwidth}
  \centering
  \includegraphics[width=\linewidth]{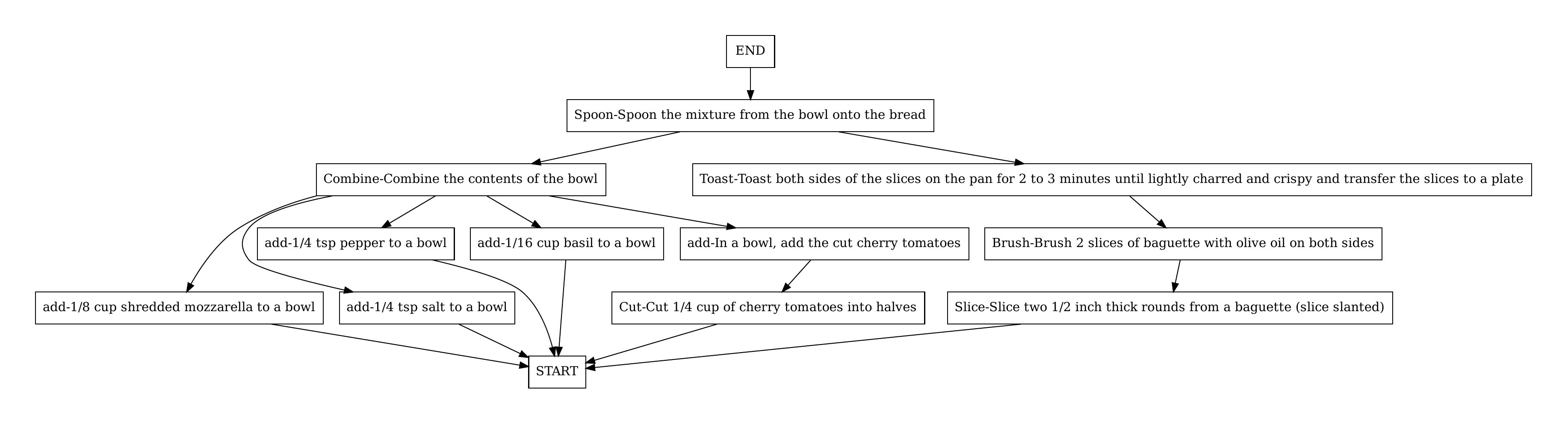}
  \caption{}
\end{subfigure}
\begin{subfigure}{.49\textwidth}
  \centering
  \includegraphics[width=\linewidth]{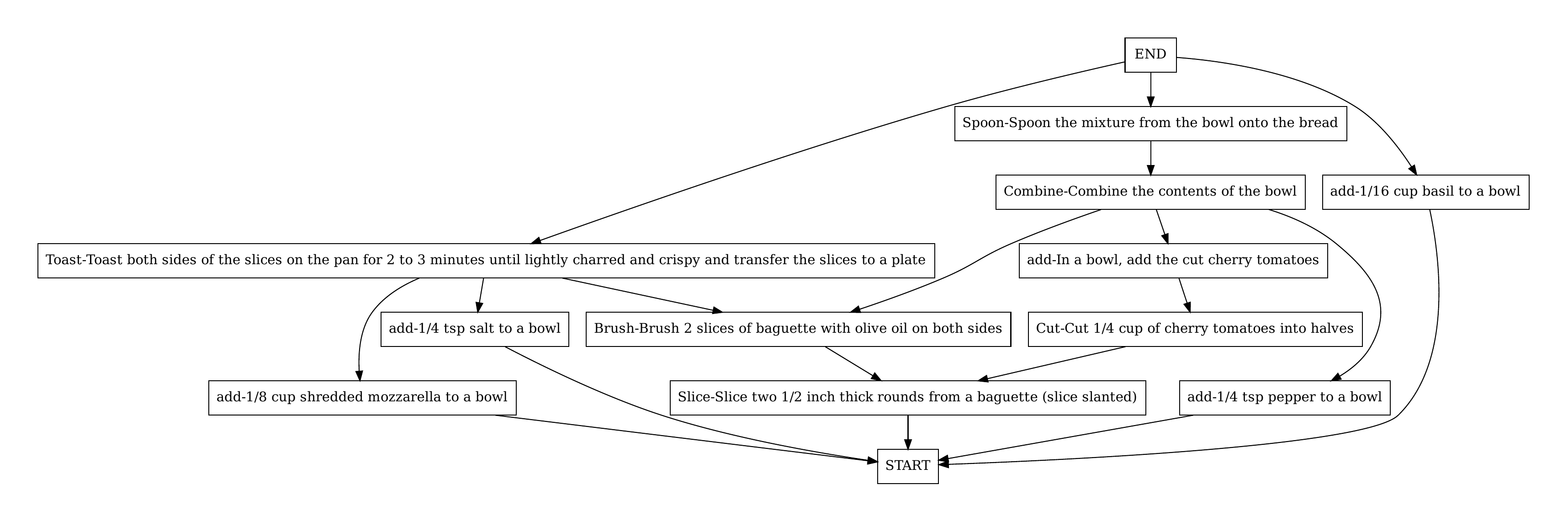}
  \caption{}
\end{subfigure}
\caption{(a) Ground truth task graph and (b) predicted task graph of the scenario Caprese Bruschetta.}
\end{figure}

\begin{figure}[H]
\centering
\begin{subfigure}{.49\textwidth}
  \centering
  \includegraphics[width=\linewidth]{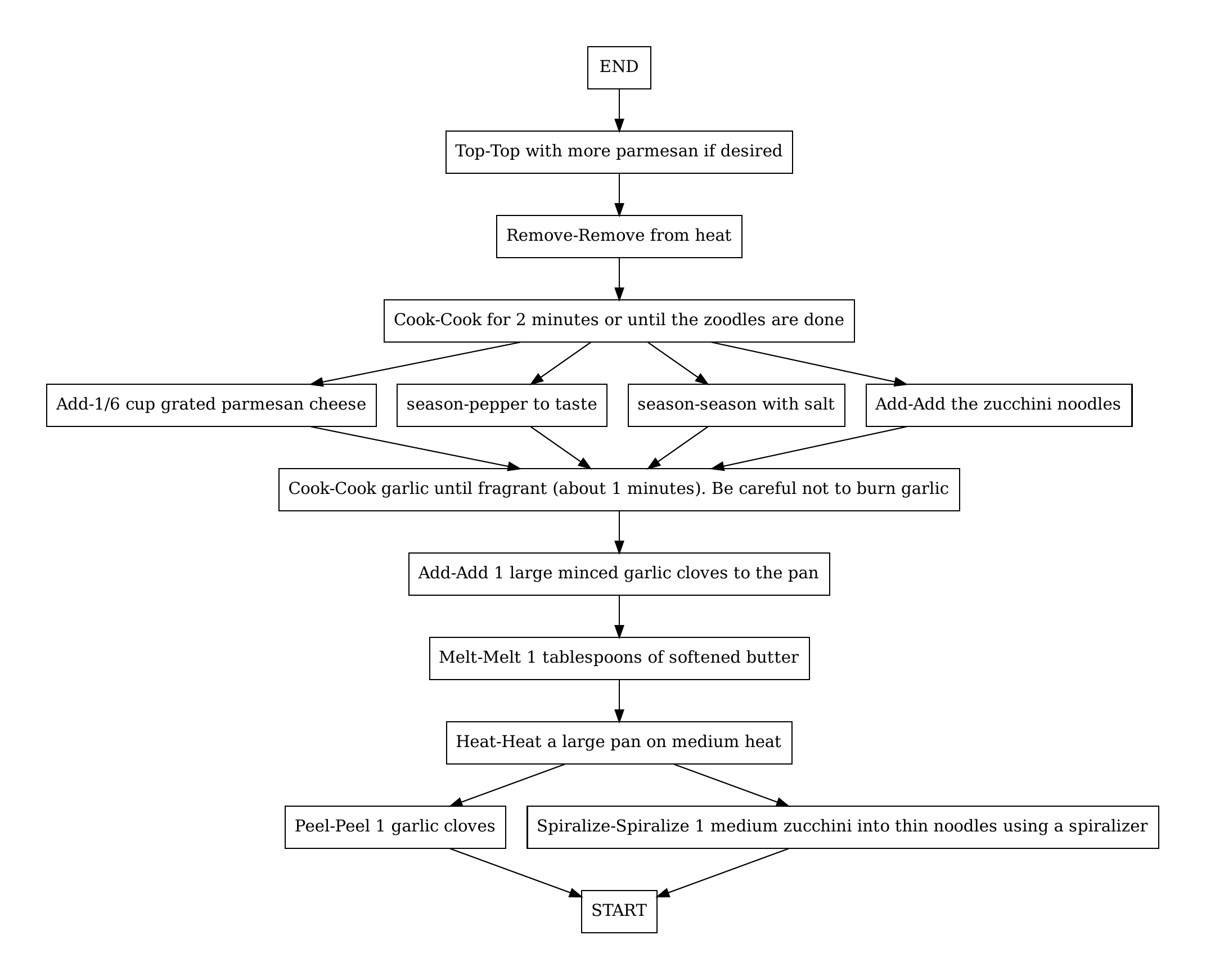}
  \caption{}
\end{subfigure}
\begin{subfigure}{.49\textwidth}
  \centering
  \includegraphics[width=\linewidth]{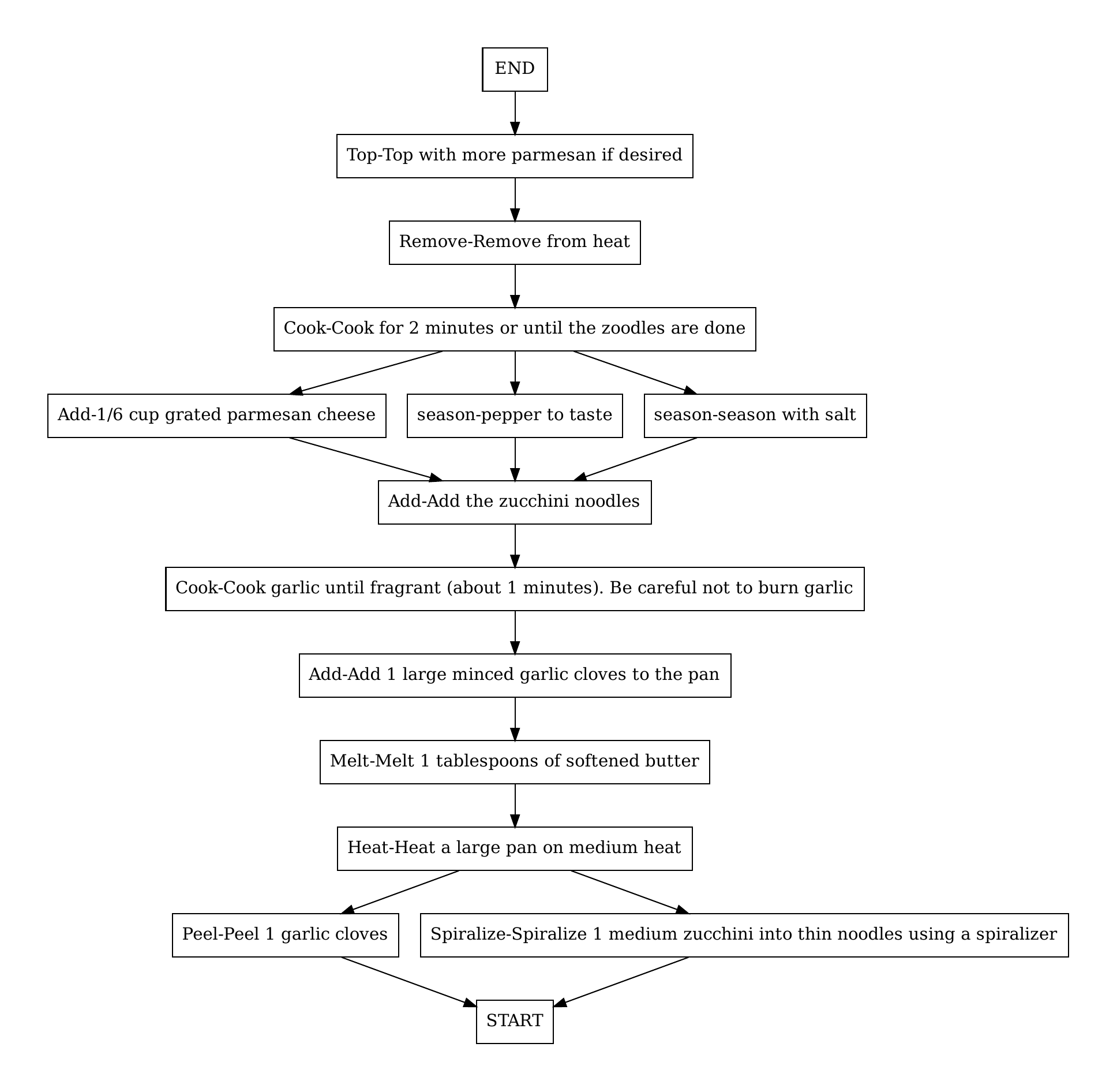}
  \caption{}
\end{subfigure}
\caption{(a) Ground truth task graph and (b) predicted task graph of the scenario Zoodles.}
\end{figure}

\begin{figure}[H]
\centering
\begin{subfigure}{.49\textwidth}
  \centering
  \includegraphics[width=\linewidth]{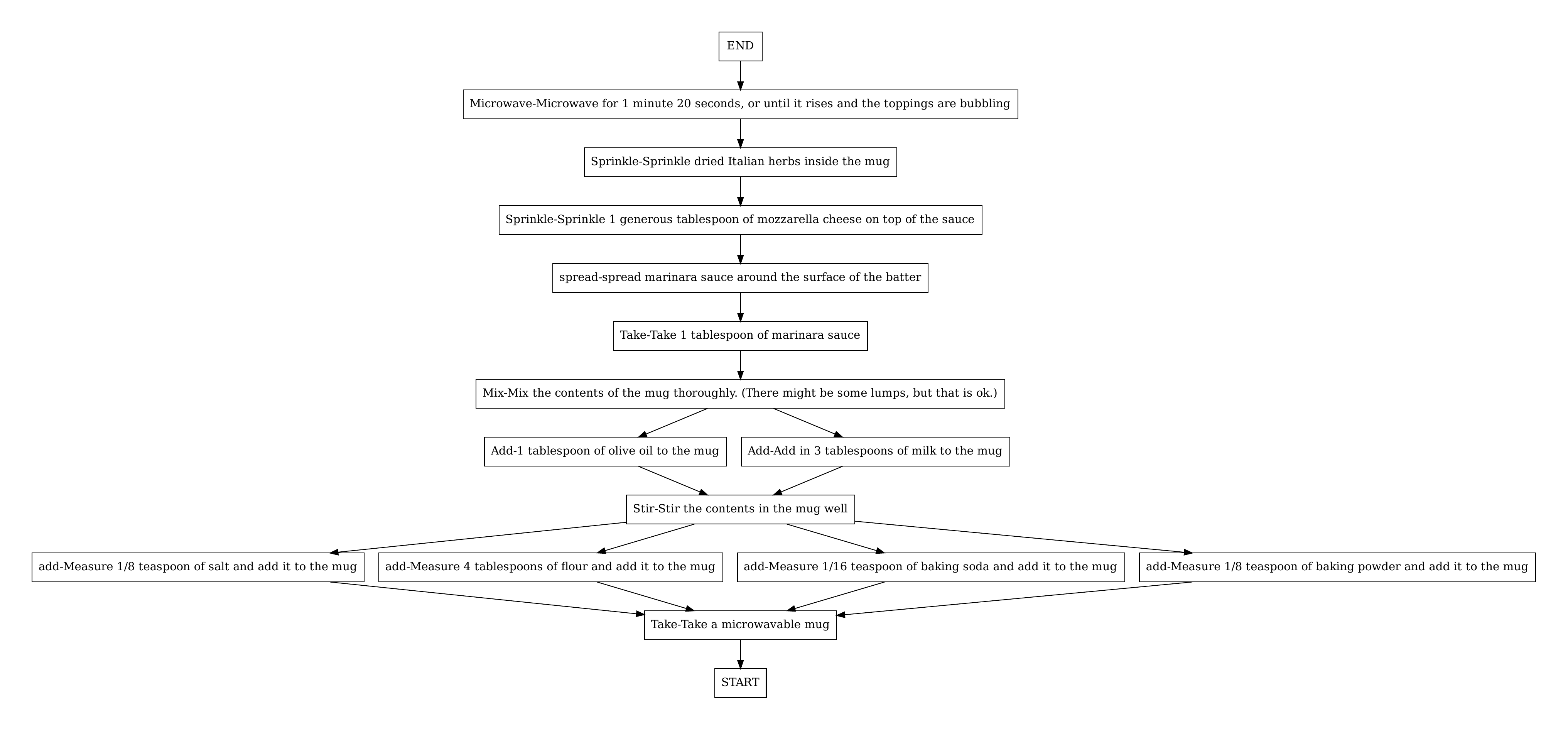}
  \caption{}
\end{subfigure}
\begin{subfigure}{.49\textwidth}
  \centering
  \includegraphics[width=\linewidth]{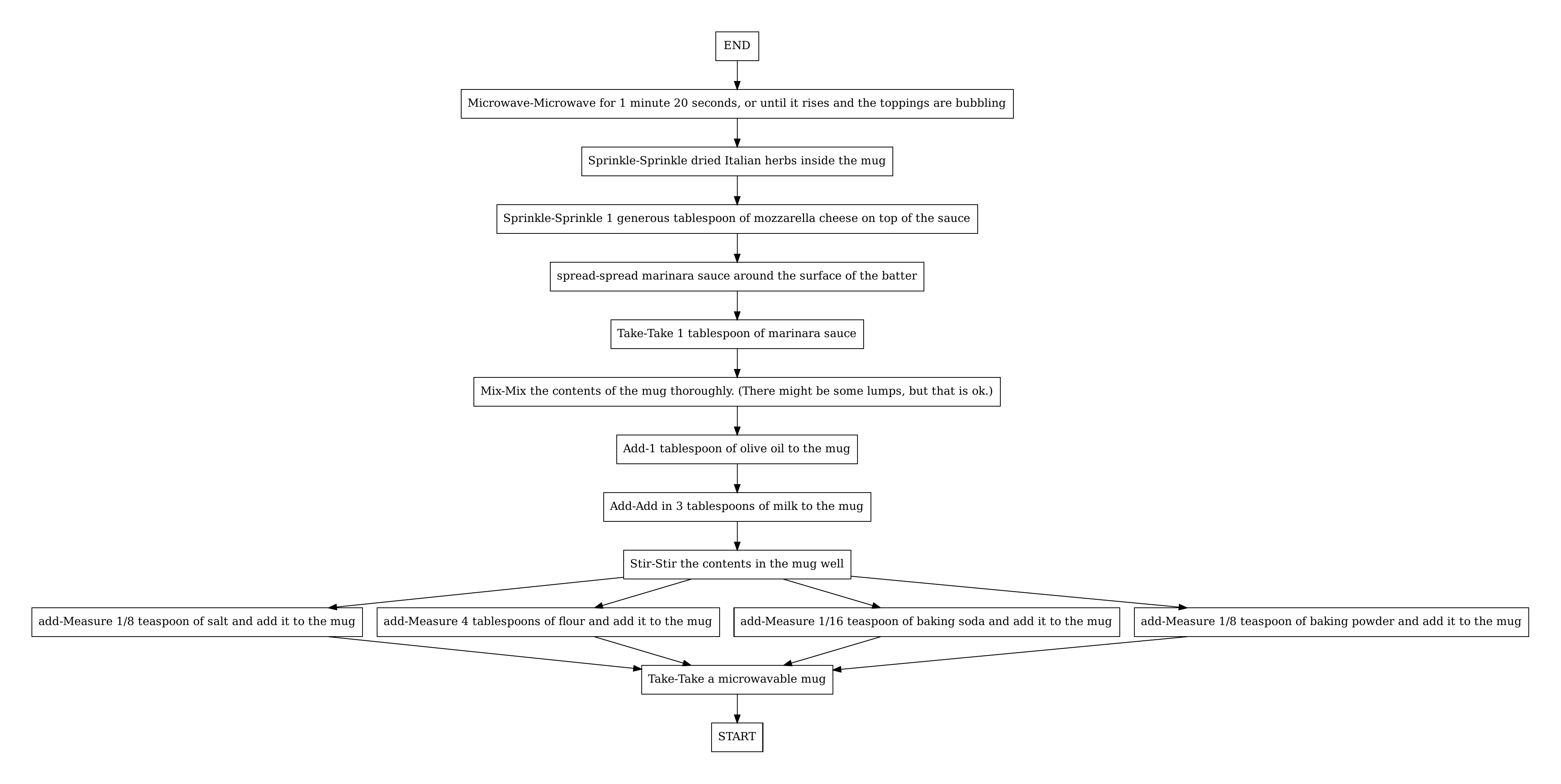}
  \caption{}
\end{subfigure}
\caption{(a) Ground truth task graph and (b) predicted task graph of the scenario Microwave Mug Pizza.}
\end{figure}

\begin{figure}[H]
\centering
\begin{subfigure}{.49\textwidth}
  \centering
  \includegraphics[width=\linewidth]{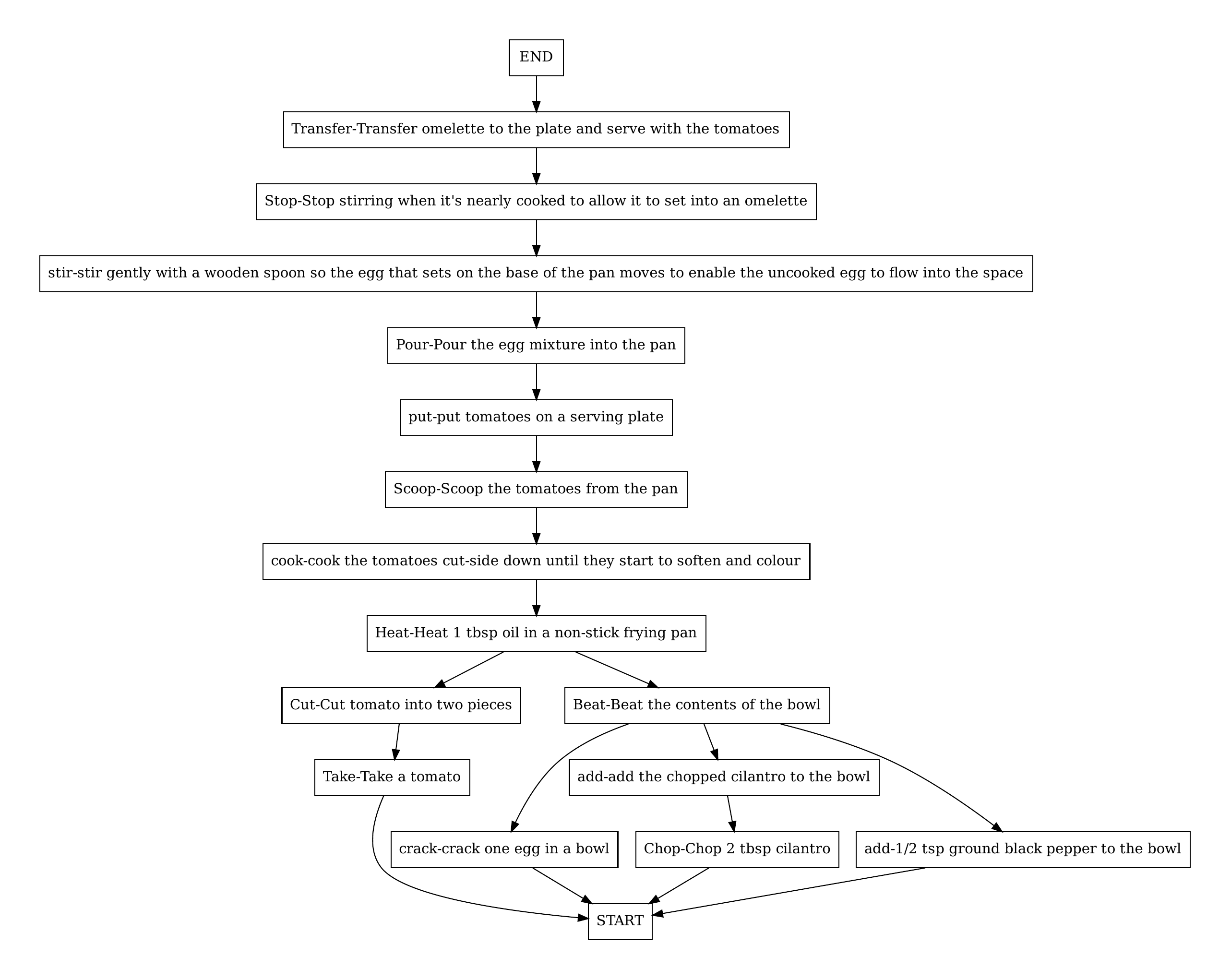}
  \caption{}
\end{subfigure}
\begin{subfigure}{.49\textwidth}
  \centering
  \includegraphics[width=\linewidth]{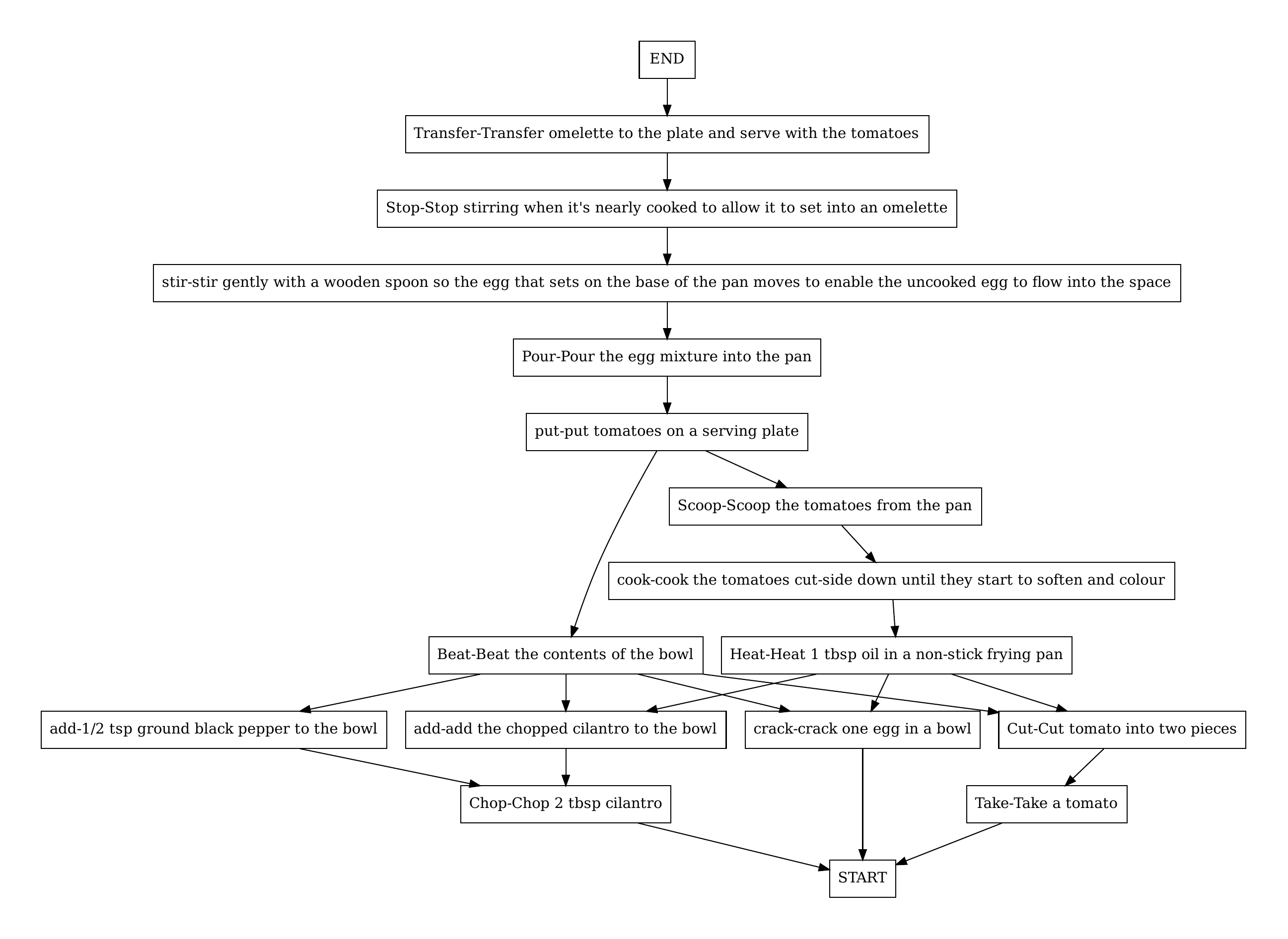}
  \caption{}
\end{subfigure}
\caption{(a) Ground truth task graph and (b) predicted task graph of the scenario Herb Omelet with Fried Tomatoes.}
\end{figure}

\begin{figure}[H]
\centering
\begin{subfigure}{.49\textwidth}
  \centering
  \includegraphics[width=\linewidth]{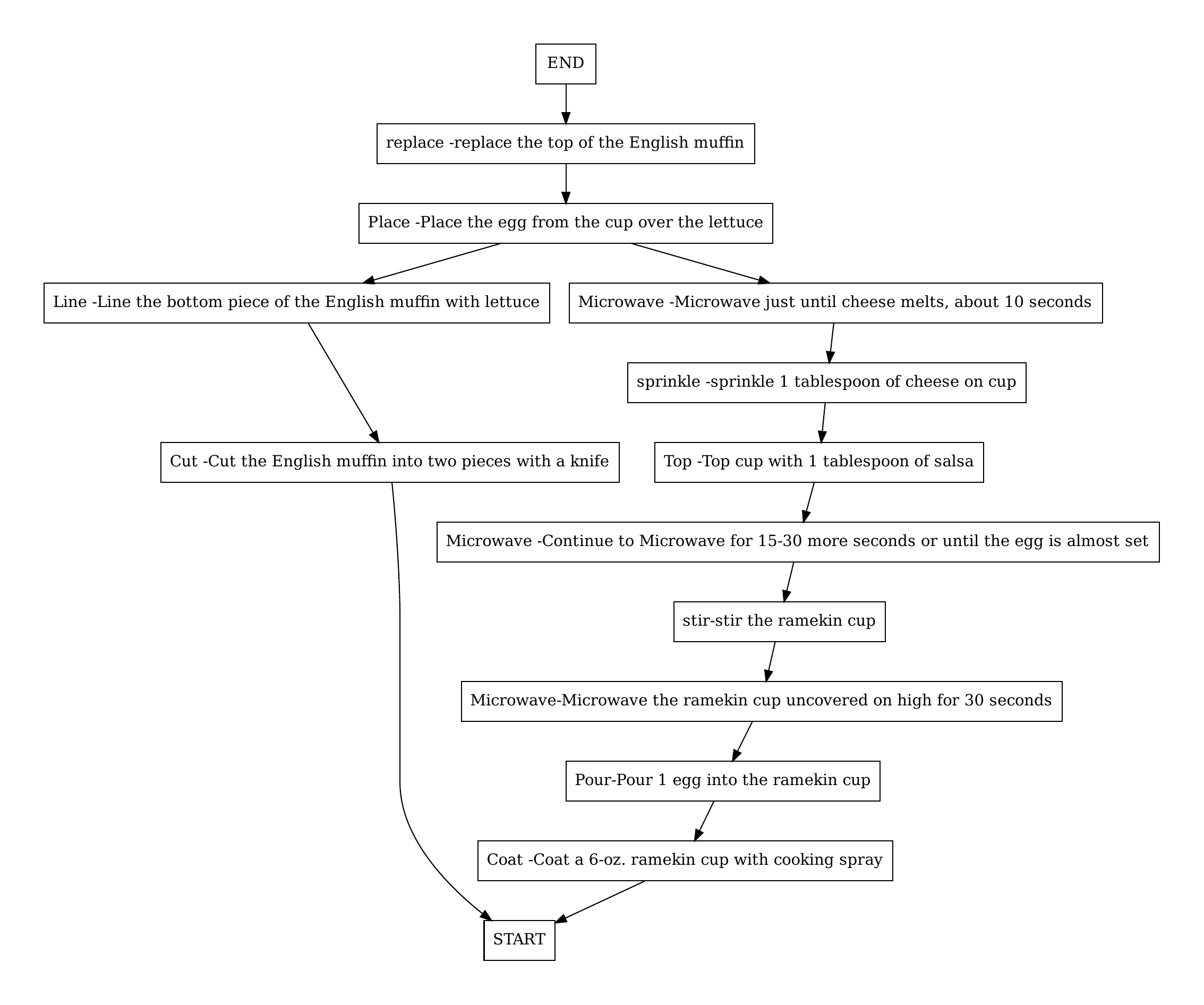}
  \caption{}
\end{subfigure}
\begin{subfigure}{.49\textwidth}
  \centering
  \includegraphics[width=\linewidth]{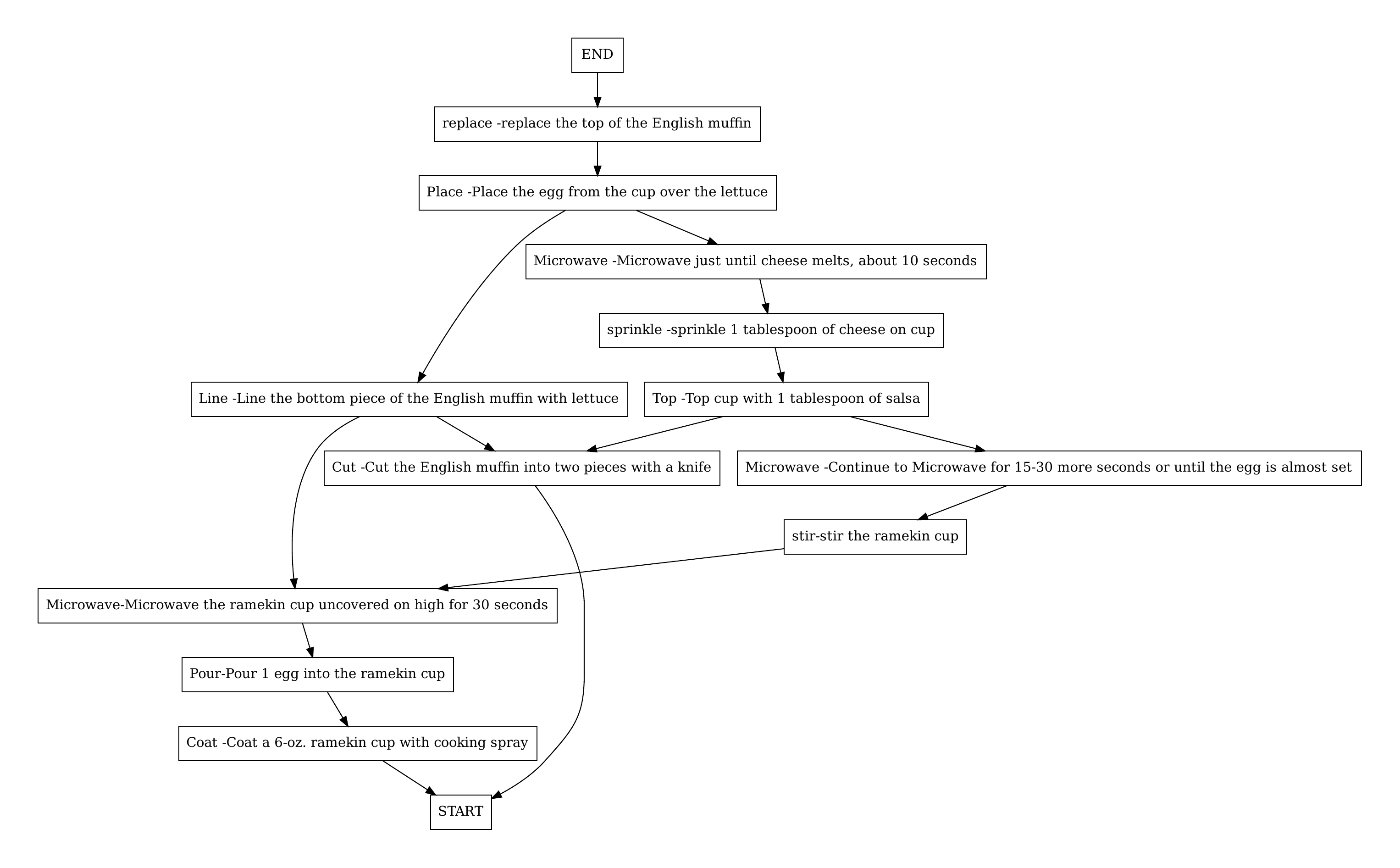}
  \caption{}
\end{subfigure}
\caption{(a) Ground truth task graph and (b) predicted task graph of the scenario Microwave Egg Sandwich.}
\end{figure}

\begin{figure}[H]
\centering
\begin{subfigure}{.49\textwidth}
  \centering
  \includegraphics[width=\linewidth]{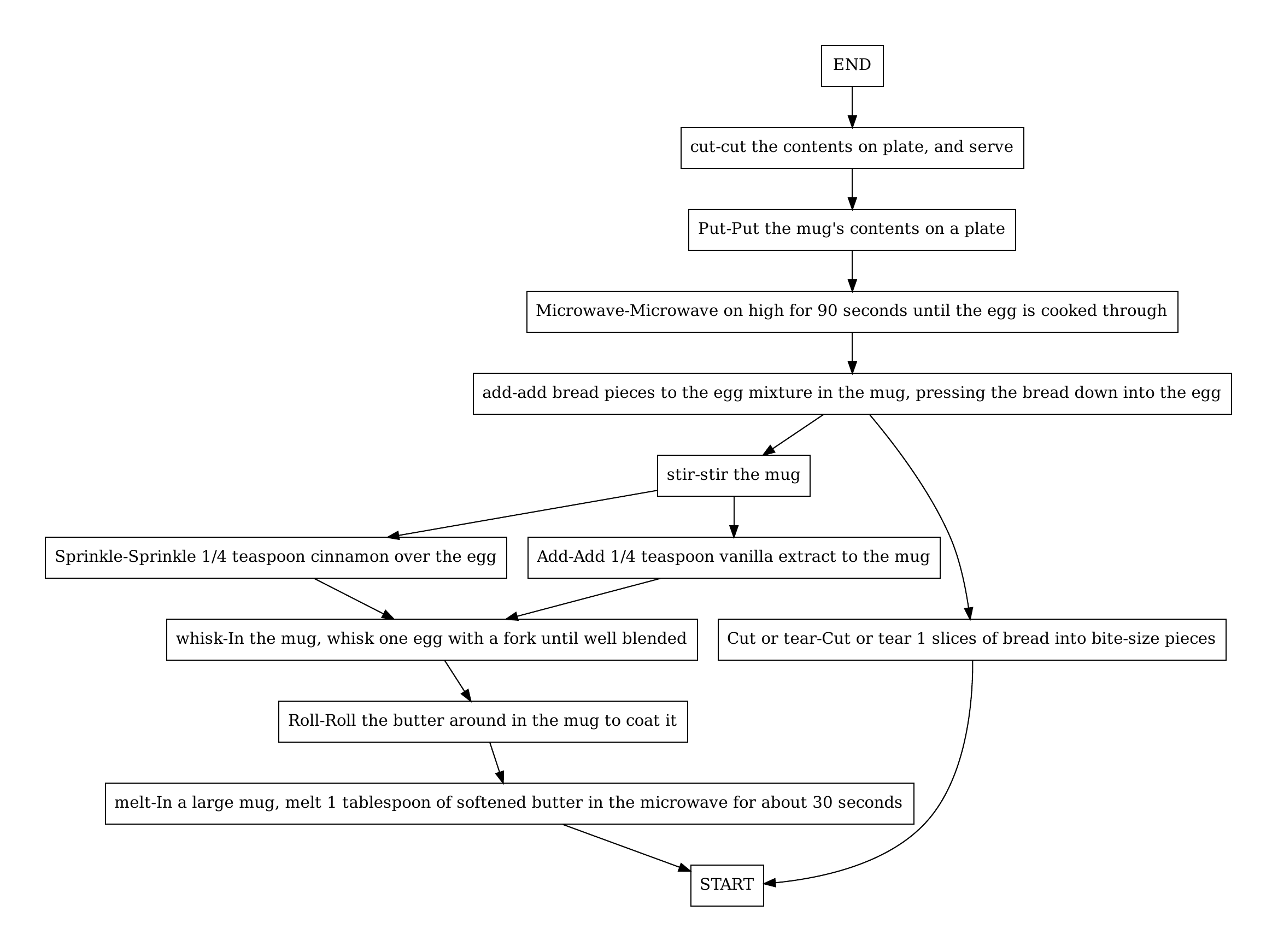}
  \caption{}
\end{subfigure}
\begin{subfigure}{.49\textwidth}
  \centering
  \includegraphics[width=\linewidth]{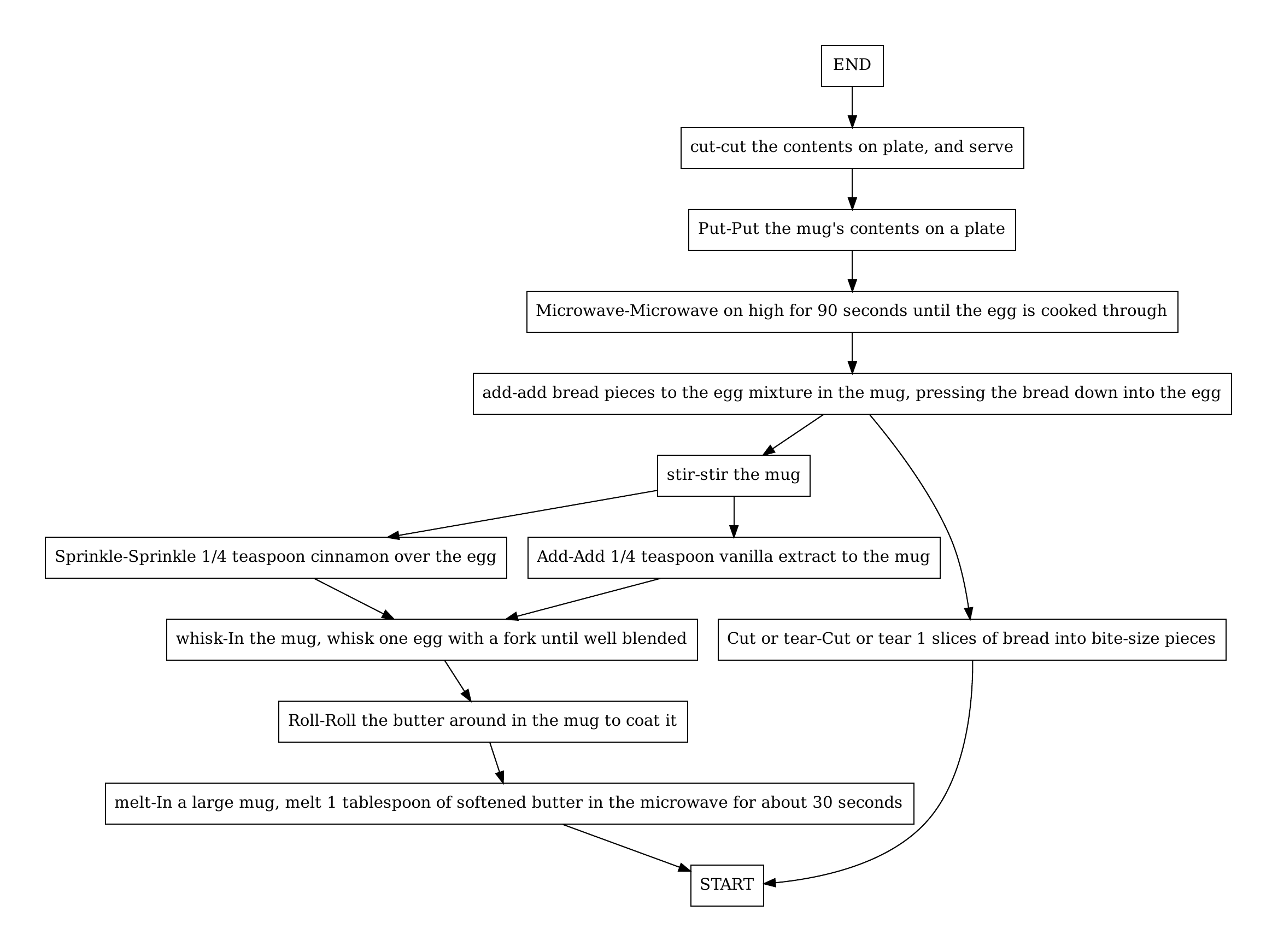}
  \caption{}
\end{subfigure}
\caption{(a) Ground truth task graph and (b) predicted task graph of the scenario Microwave French Toast.}
\end{figure}

\begin{figure}[H]
\centering
\begin{subfigure}{.49\textwidth}
  \centering
  \includegraphics[width=\linewidth]{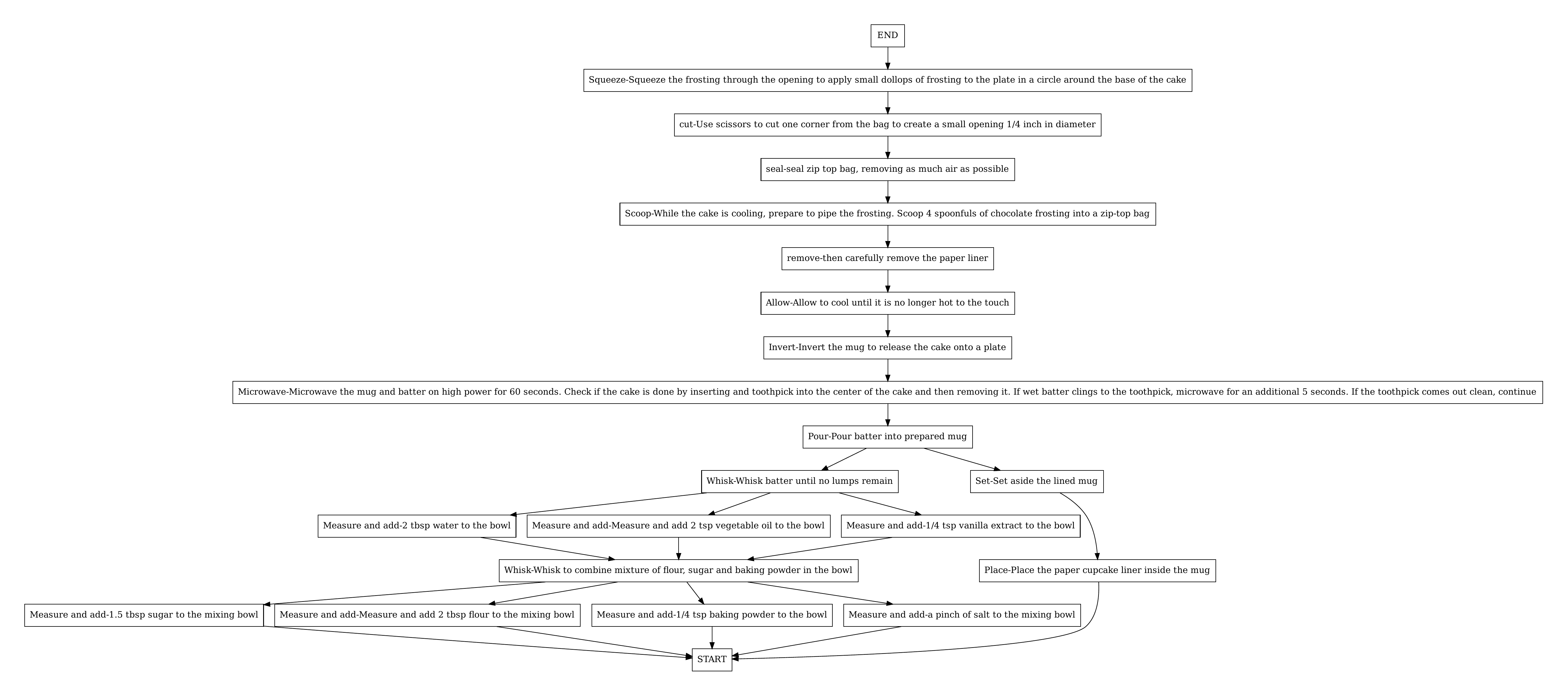}
  \caption{}
\end{subfigure}
\begin{subfigure}{.49\textwidth}
  \centering
  \includegraphics[width=\linewidth]{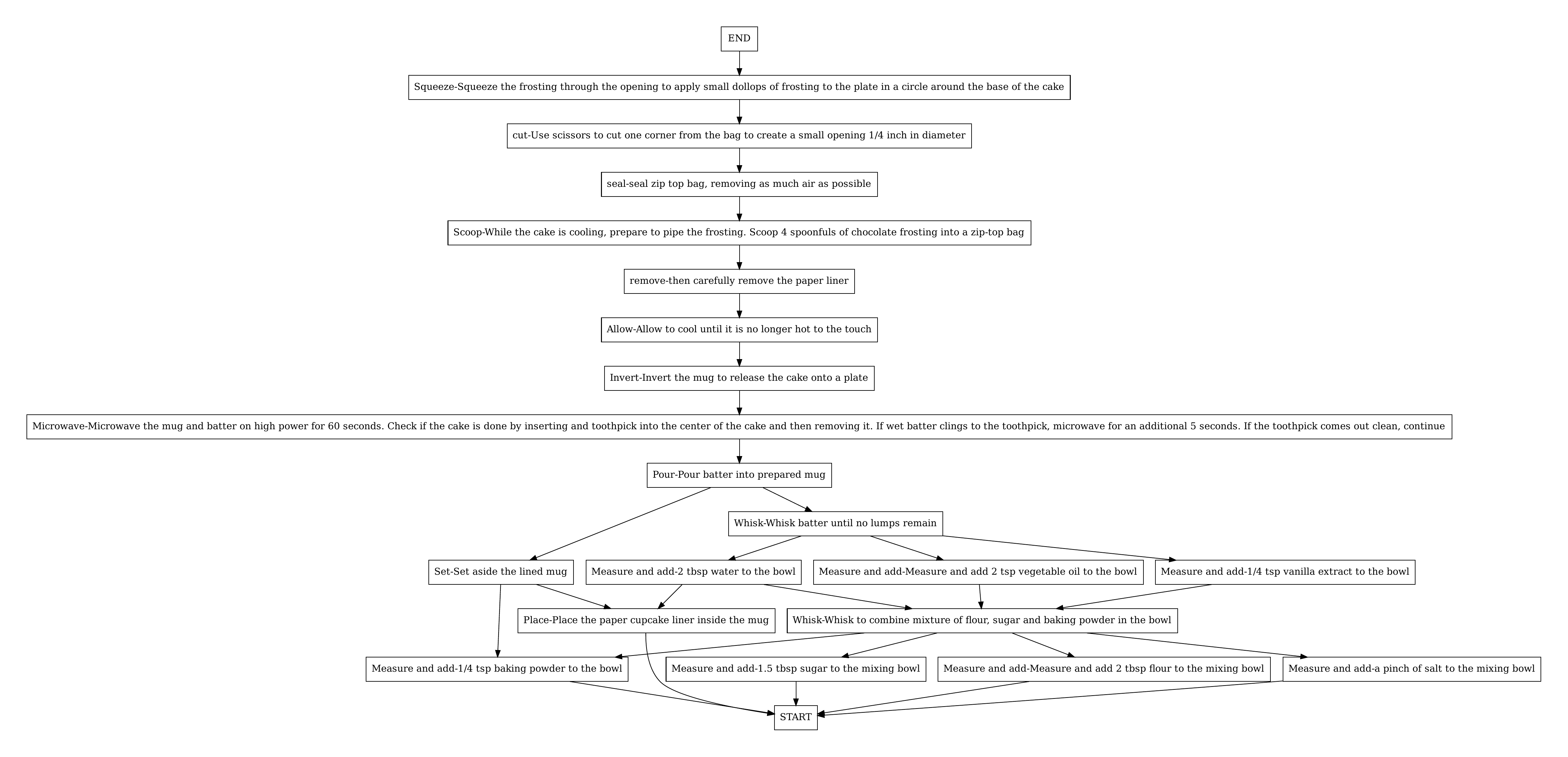}
  \caption{}
\end{subfigure}
\caption{(a) Ground truth task graph and (b) predicted task graph of the scenario Mug Cake.}
\end{figure}

\begin{figure}[H]
\centering
\begin{subfigure}{.49\textwidth}
  \centering
  \includegraphics[width=\linewidth]{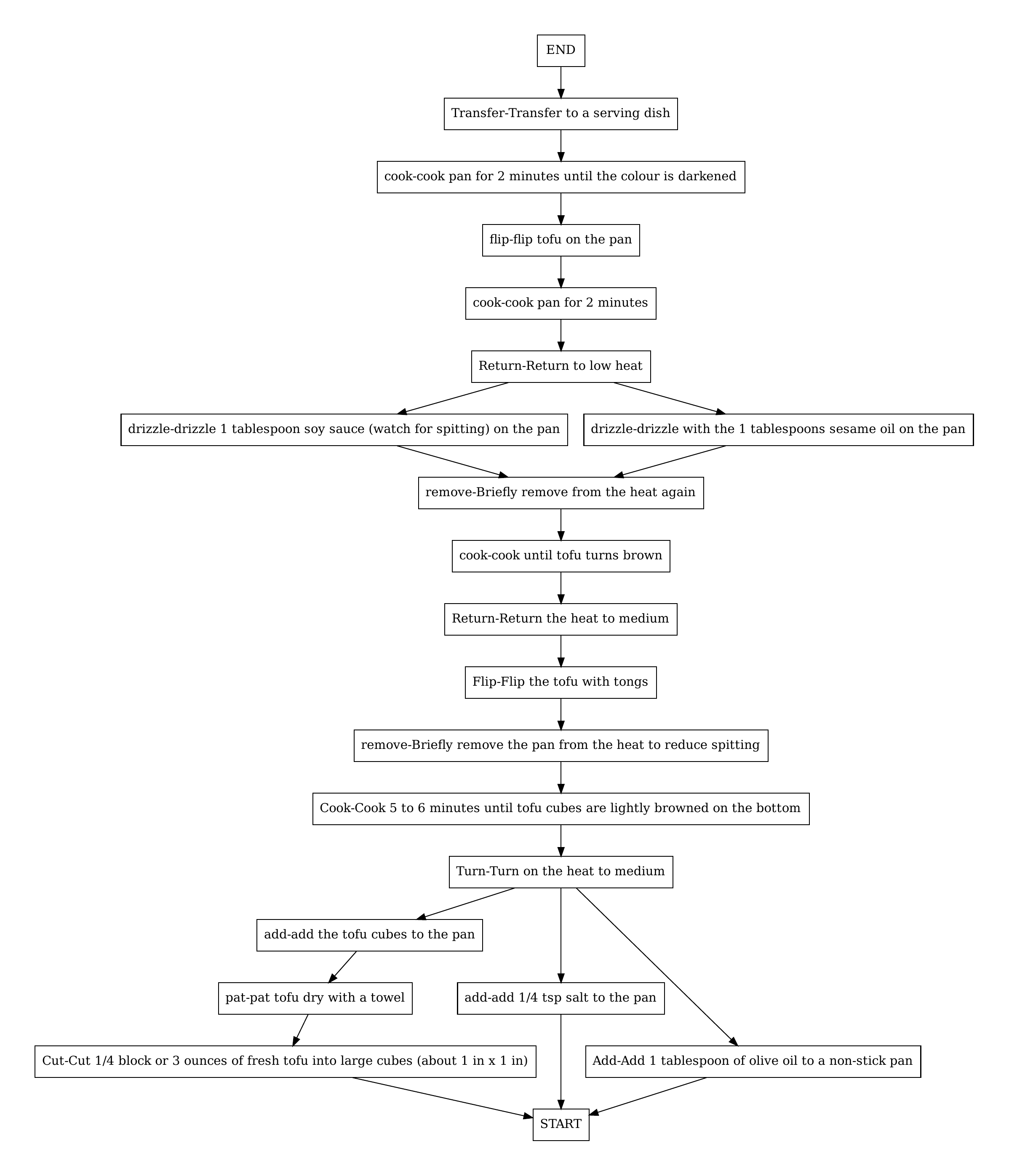}
  \caption{}
\end{subfigure}
\begin{subfigure}{.49\textwidth}
  \centering
  \includegraphics[width=\linewidth]{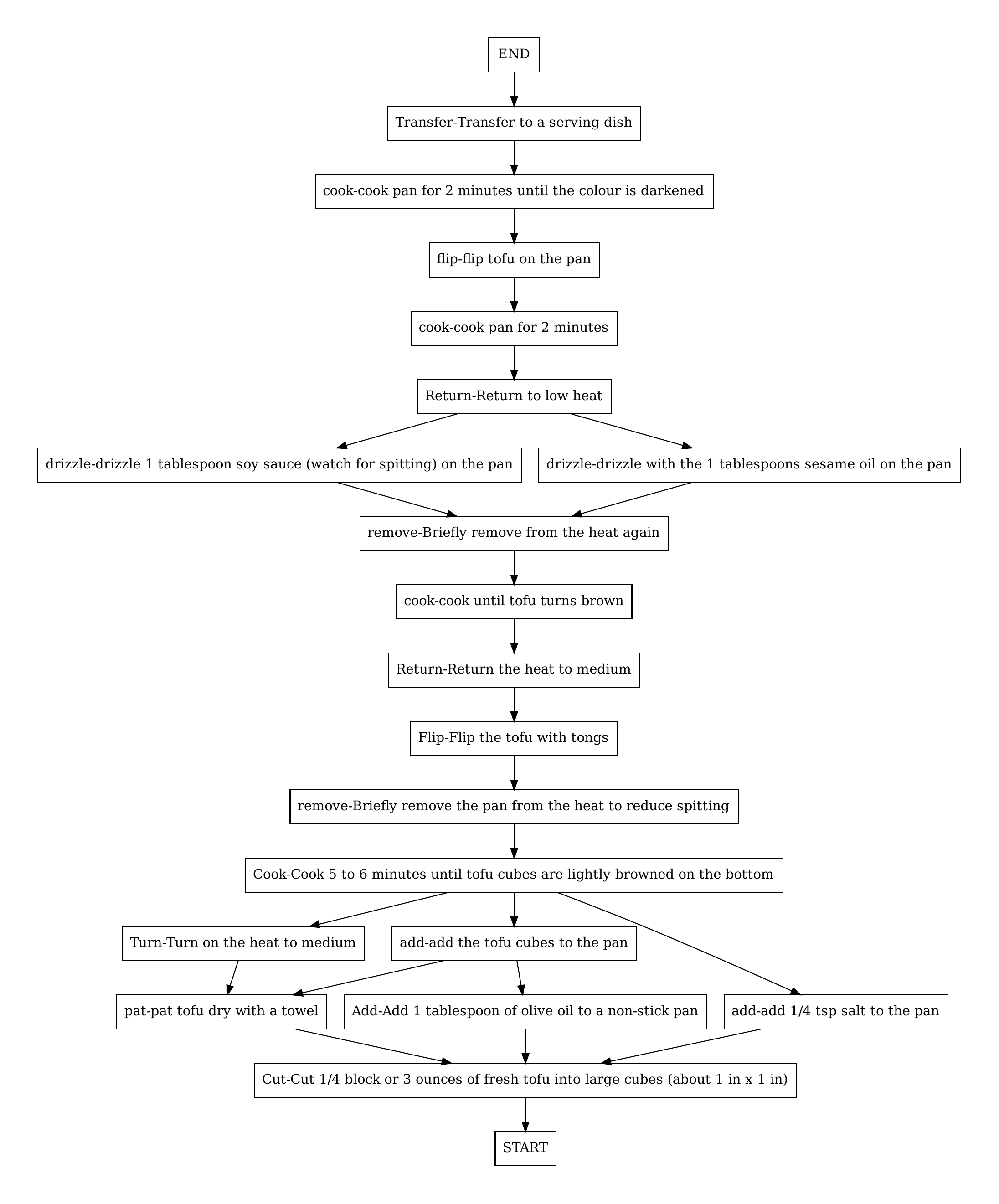}
  \caption{}
\end{subfigure}
\caption{(a) Ground truth task graph and (b) predicted task graph of the scenario Pan Fried Tofu.}
\end{figure}

\begin{figure}[H]
\centering
\begin{subfigure}{.49\textwidth}
  \centering
  \includegraphics[width=\linewidth]{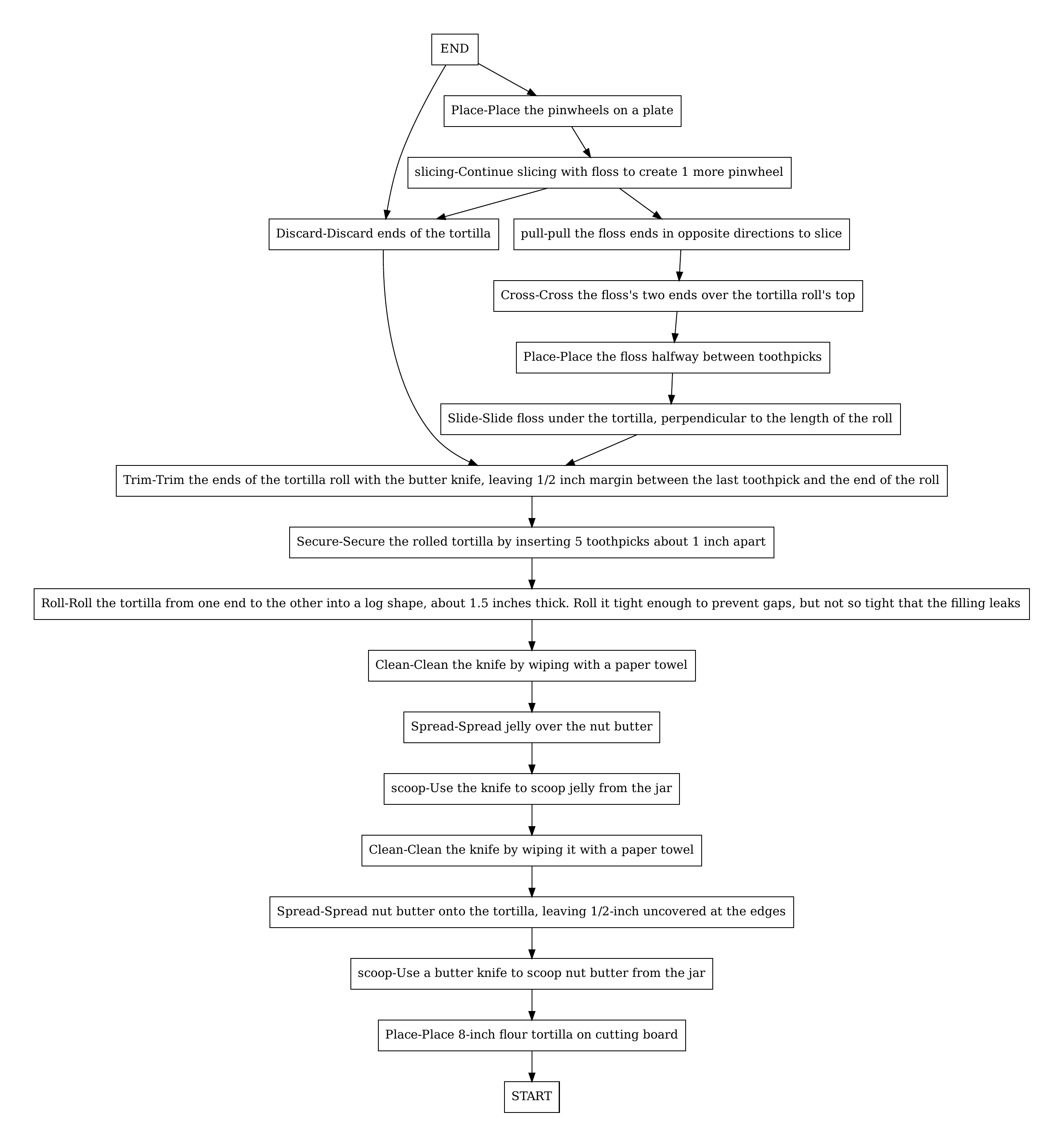}
  \caption{}
\end{subfigure}
\begin{subfigure}{.49\textwidth}
  \centering
  \includegraphics[width=\linewidth]{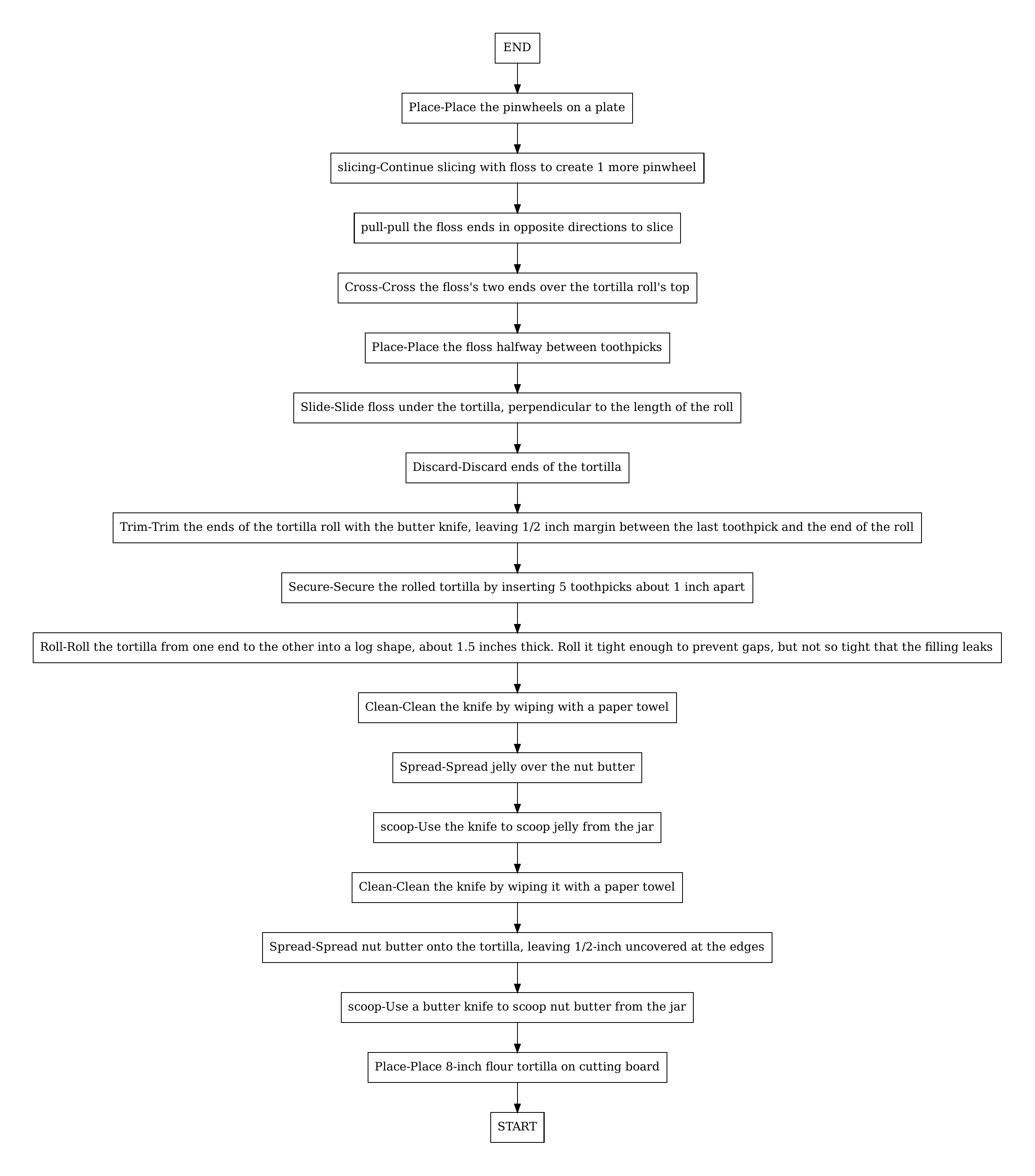}
  \caption{}
\end{subfigure}
\caption{(a) Ground truth task graph and (b) predicted task graph of the scenario Pinwheels.}
\end{figure}

\begin{figure}[H]
\centering
\begin{subfigure}{.49\textwidth}
  \centering
  \includegraphics[width=\linewidth]{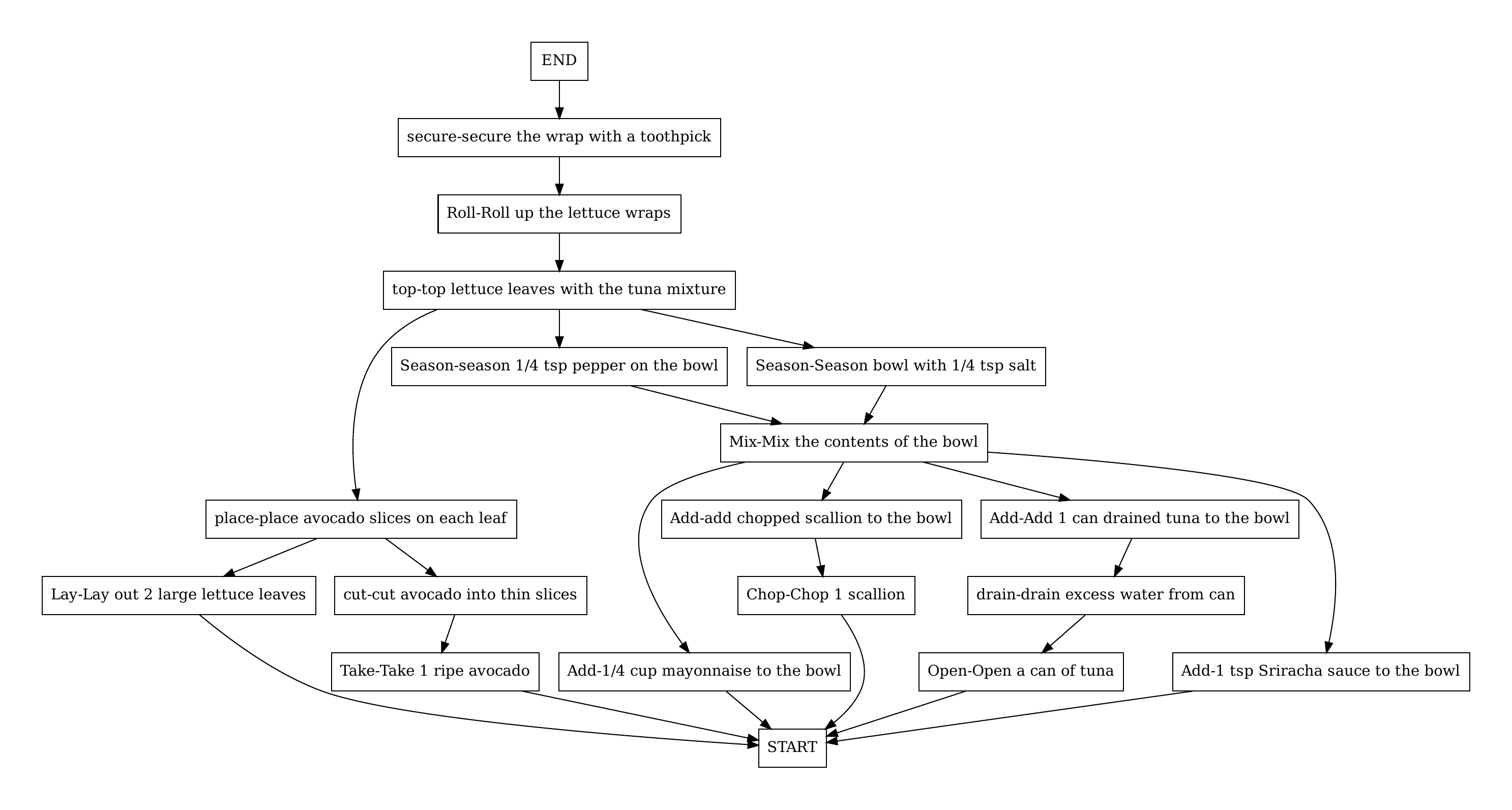}
  \caption{}
\end{subfigure}
\begin{subfigure}{.49\textwidth}
  \centering
  \includegraphics[width=\linewidth]{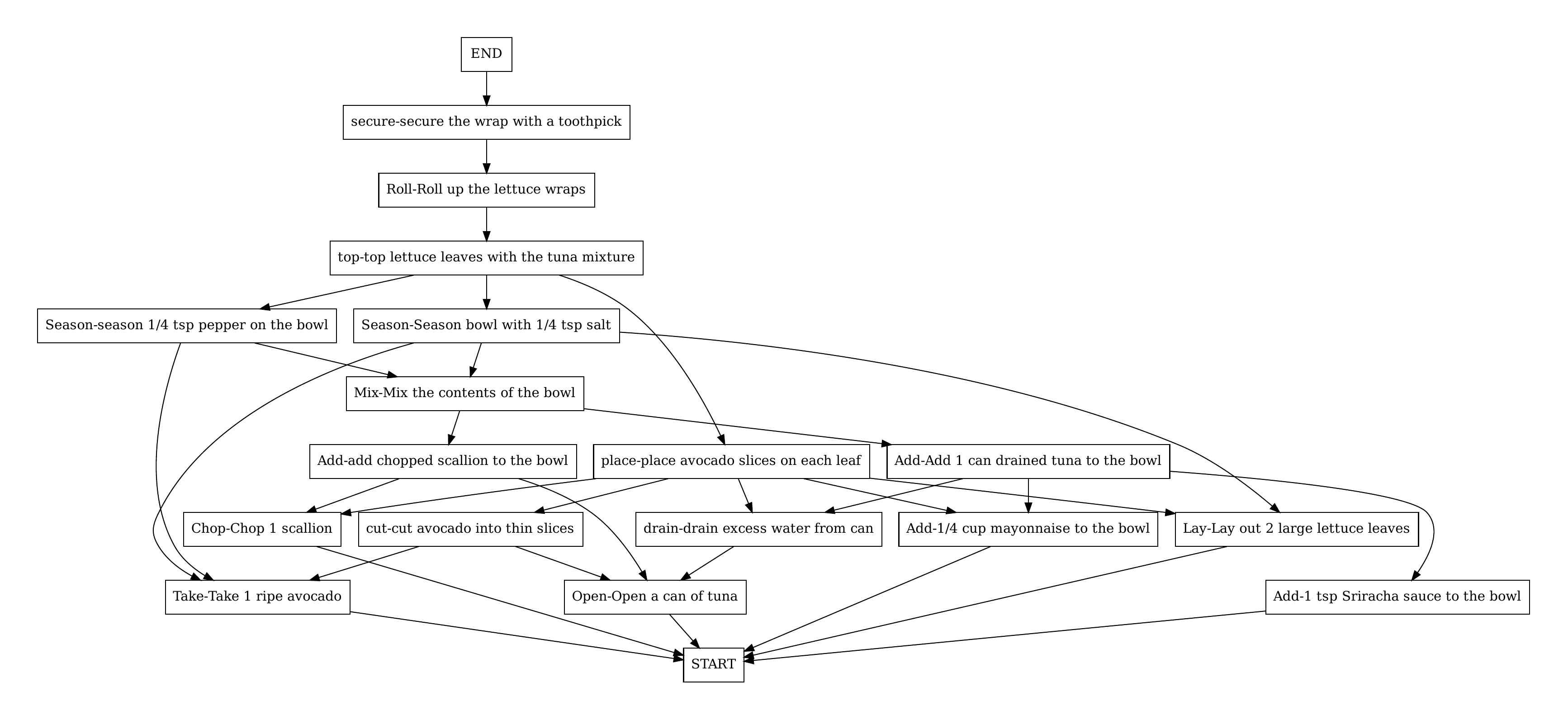}
\end{subfigure}
\caption{(a) Ground truth task graph and (b) predicted task graph of the scenario Spicy Tuna Avocado Wraps.}
\end{figure}

\begin{figure}[H]
\centering
\begin{subfigure}{.49\textwidth}
  \centering
  \includegraphics[width=\linewidth]{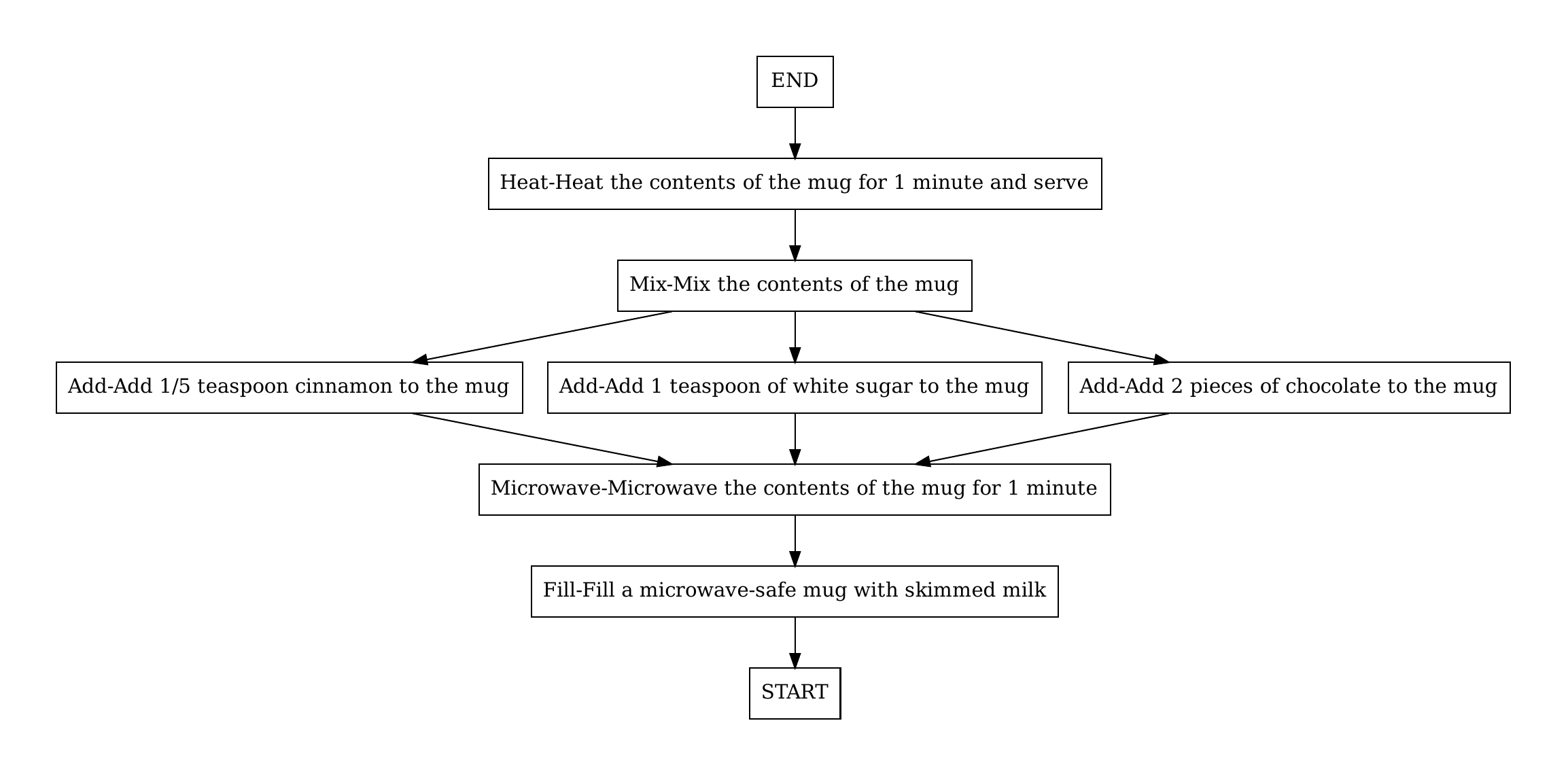}
  \caption{}
\end{subfigure}
\begin{subfigure}{.49\textwidth}
  \centering
  \includegraphics[width=\linewidth]{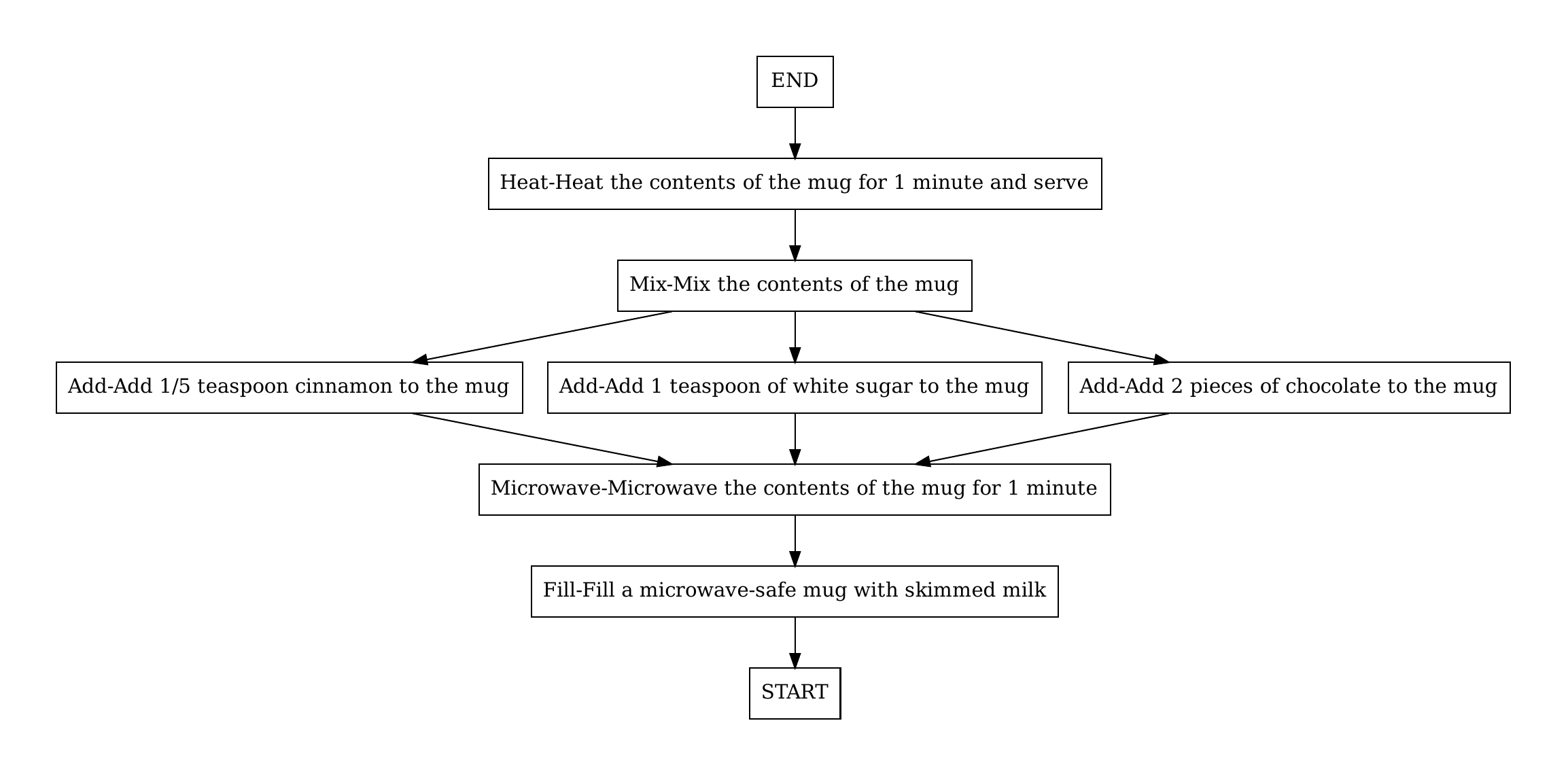}
  \caption{}
\end{subfigure}
\caption{(a) Ground truth task graph and (b) predicted task graph of the scenario Spiced Hot Chocolate.}
\end{figure}

\begin{figure}[H]
\centering
\begin{subfigure}{.49\textwidth}
  \centering
  \includegraphics[width=\linewidth]{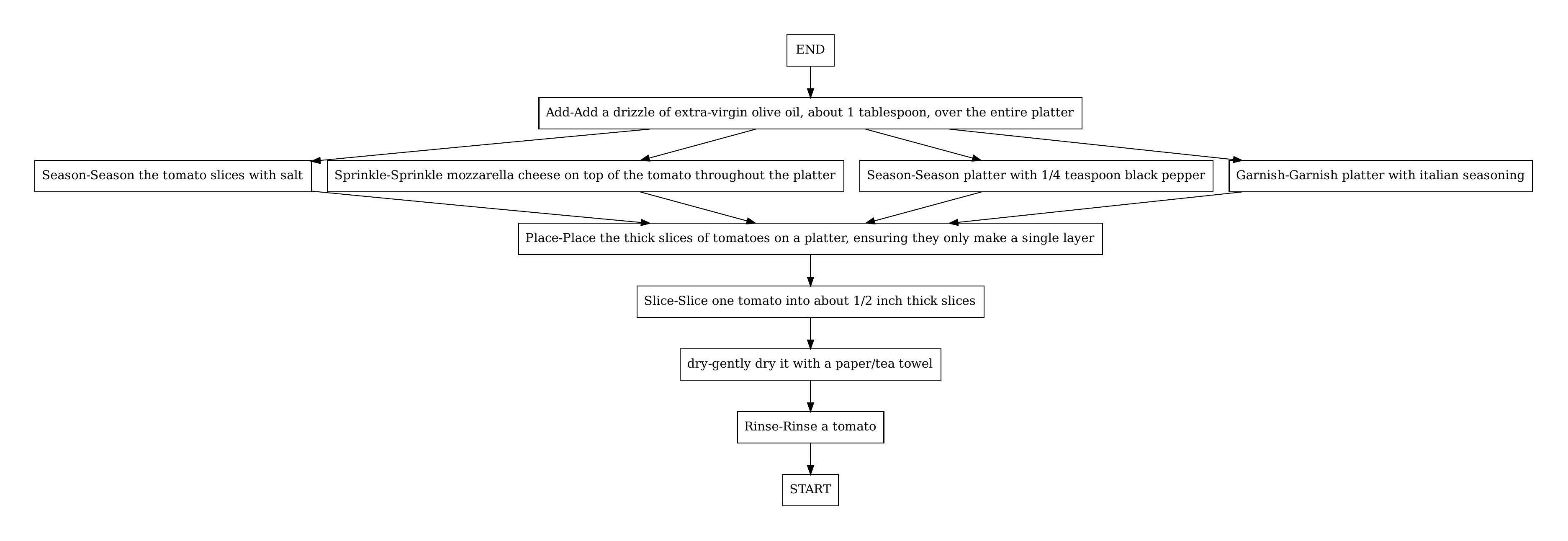}
  \caption{}
\end{subfigure}
\begin{subfigure}{.49\textwidth}
  \centering
  \includegraphics[width=\linewidth]{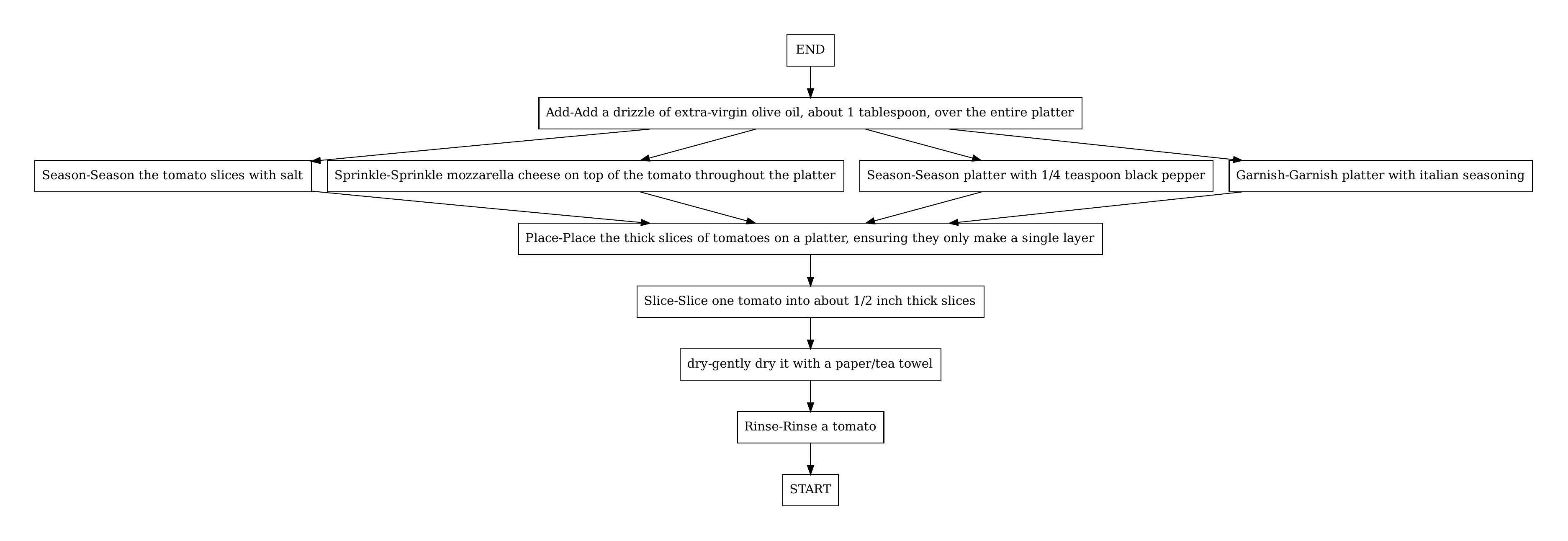}
  \caption{}
\end{subfigure}
\caption{(a) Ground truth task graph and (b) predicted task graph of the scenario Tomato Mozzarella Salad.}
\end{figure}

\begin{figure}[H]
\centering
\begin{subfigure}{.49\textwidth}
  \centering
  \includegraphics[width=\linewidth]{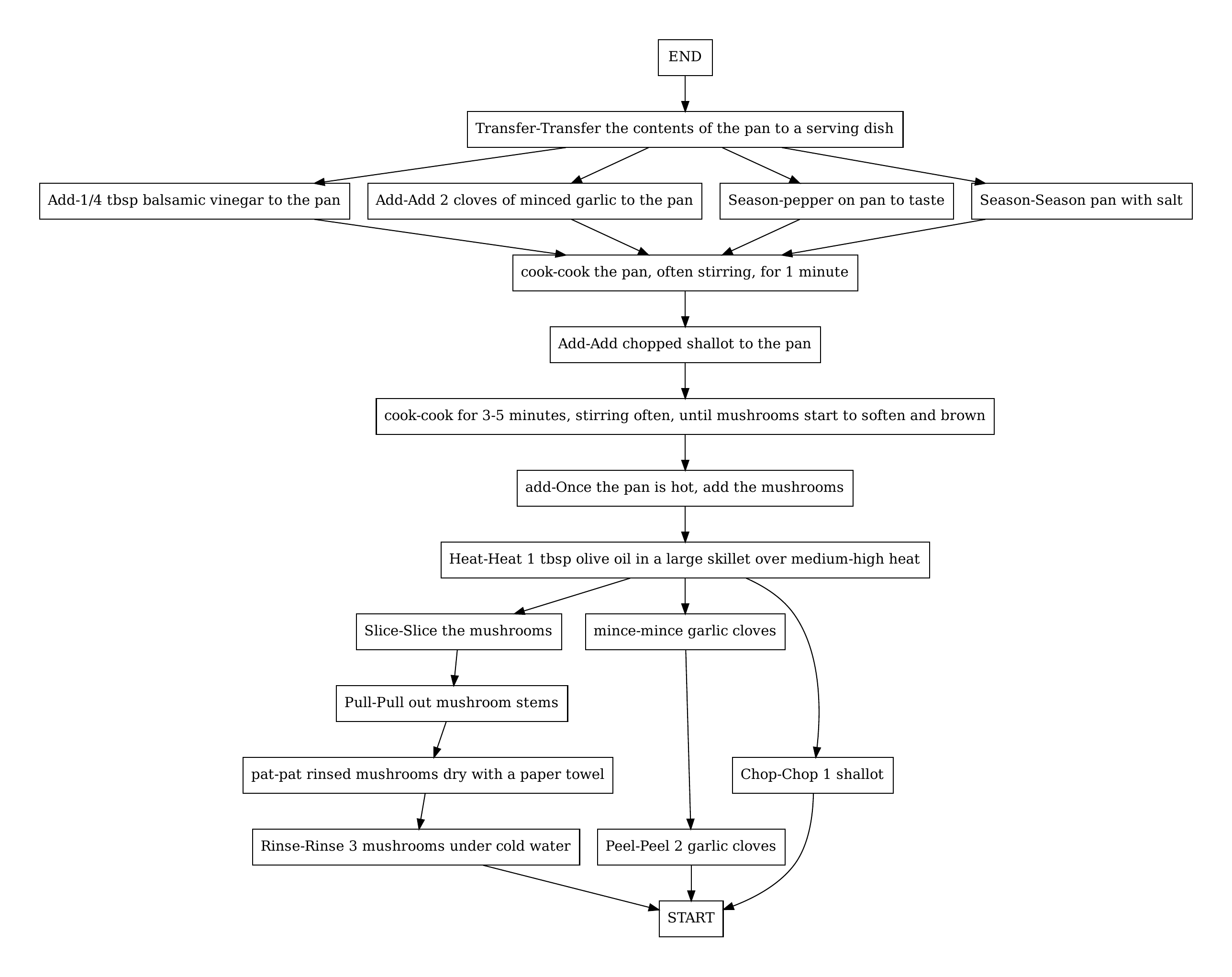}
  \caption{}
\end{subfigure}
\begin{subfigure}{.49\textwidth}
  \centering
  \includegraphics[width=\linewidth]{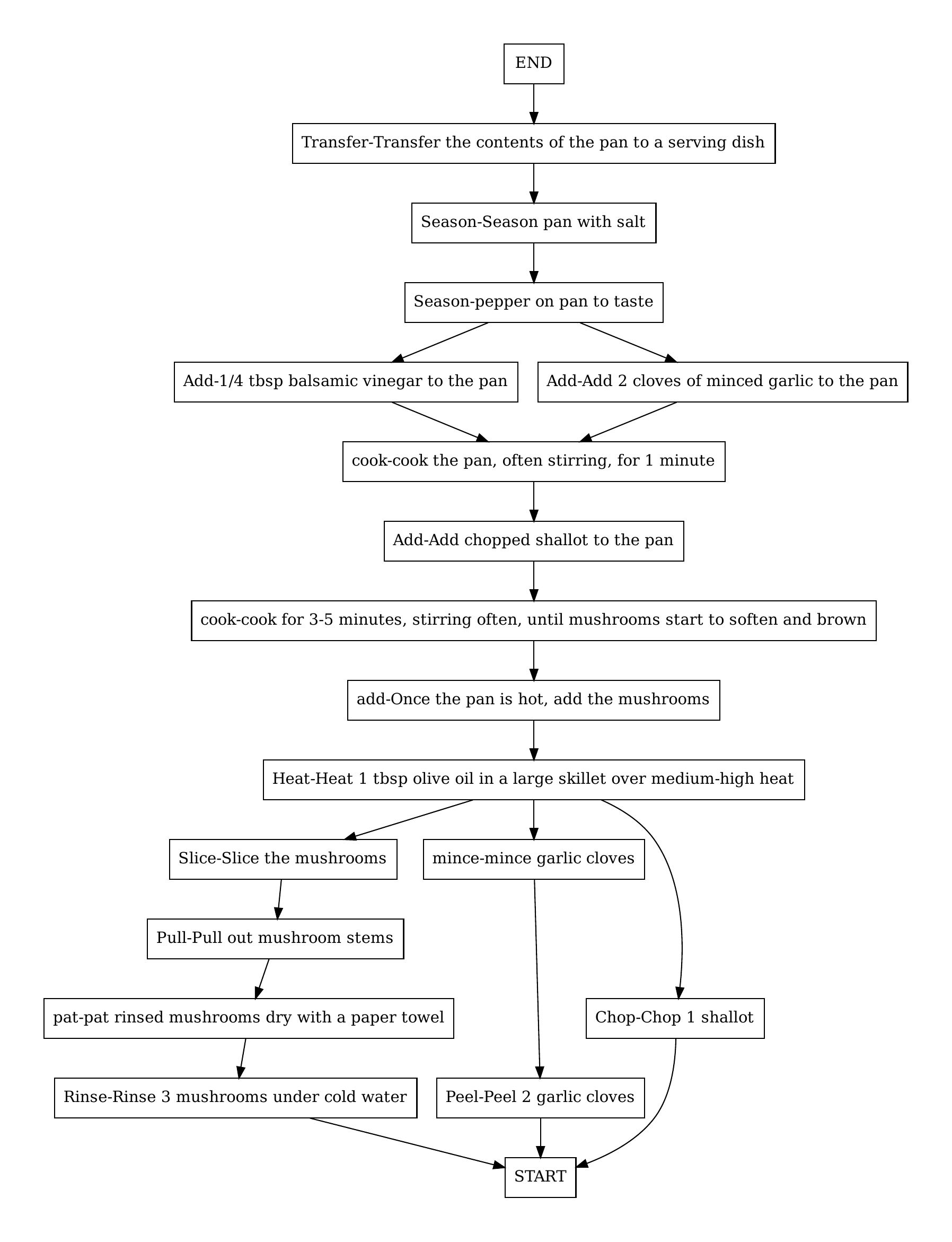}
  \caption{}
\end{subfigure}
\caption{(a) Ground truth task graph and (b) predicted task graph of the scenario Salted Mushrooms.}
\end{figure}

\begin{figure}[H]
\centering
\centering
\begin{subfigure}{.49\textwidth}
  \centering
  \includegraphics[width=\linewidth]{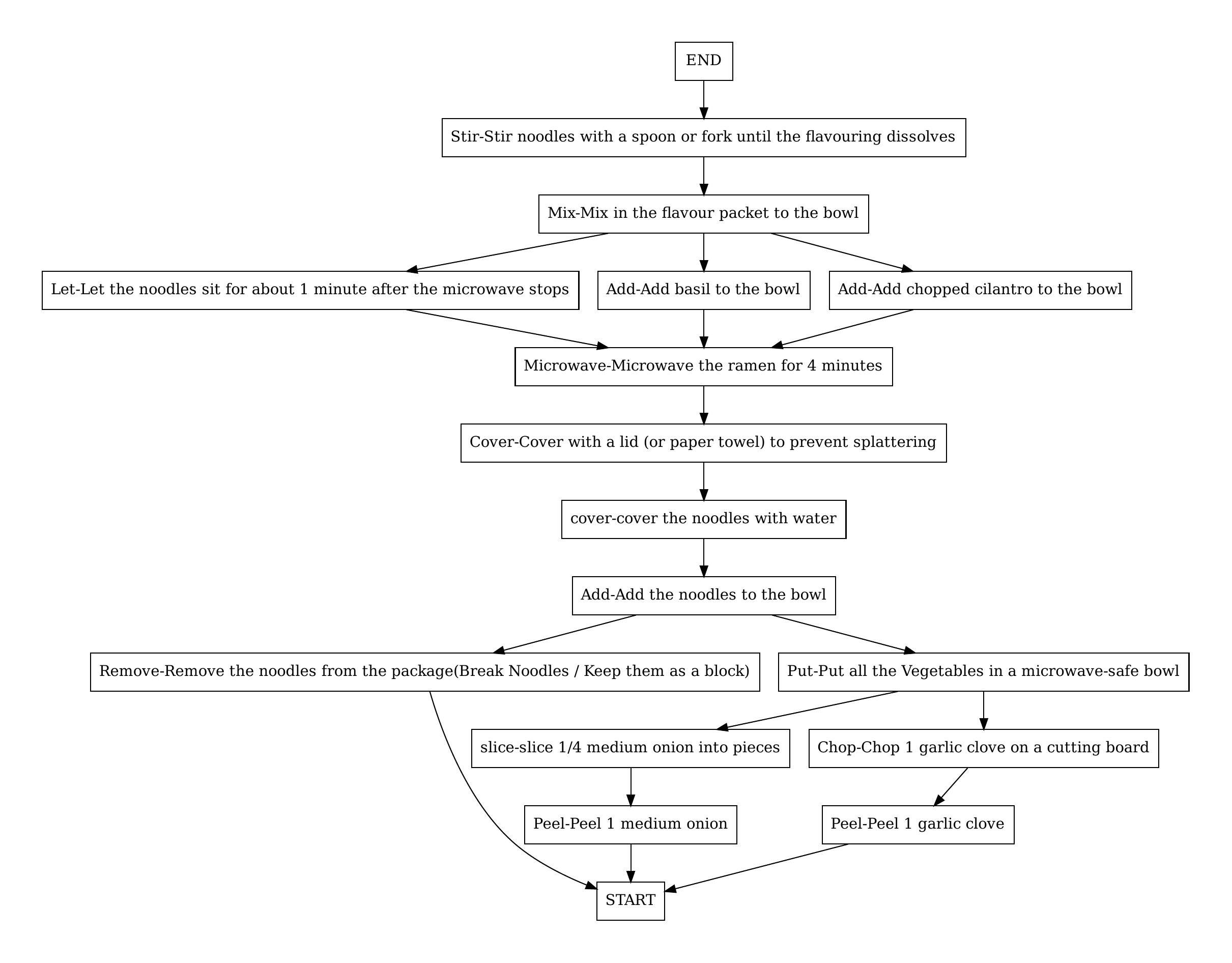}
  \caption{}
\end{subfigure}
\begin{subfigure}{.49\textwidth}
  \centering
  \includegraphics[width=\linewidth]{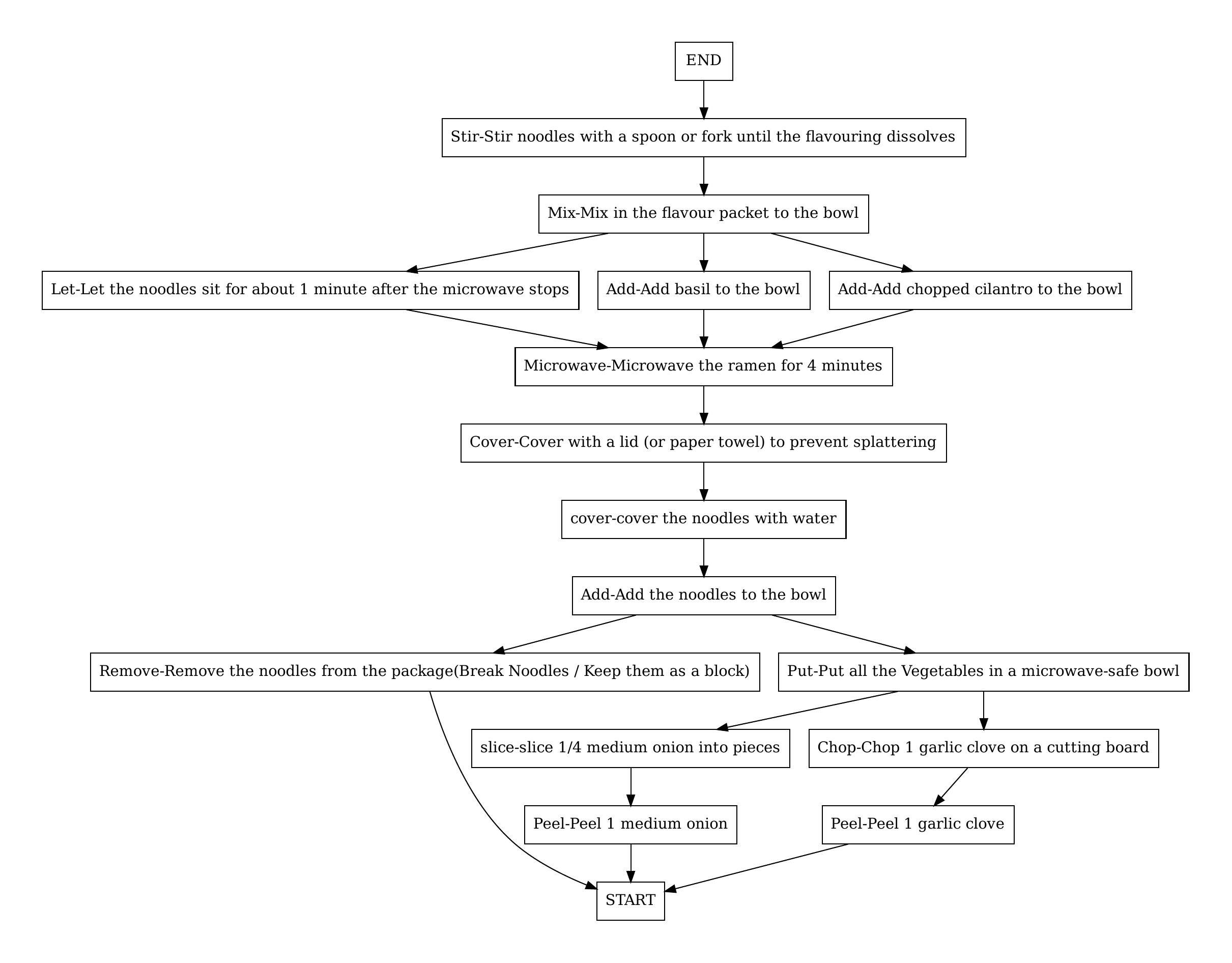}
  \caption{}
\end{subfigure}
\caption{(a) Ground truth task graph and (b) predicted task graph of the scenario Ramen.}
\end{figure}

\begin{figure}[H]
\centering
\begin{subfigure}{.49\textwidth}
  \centering
  \includegraphics[width=\linewidth]{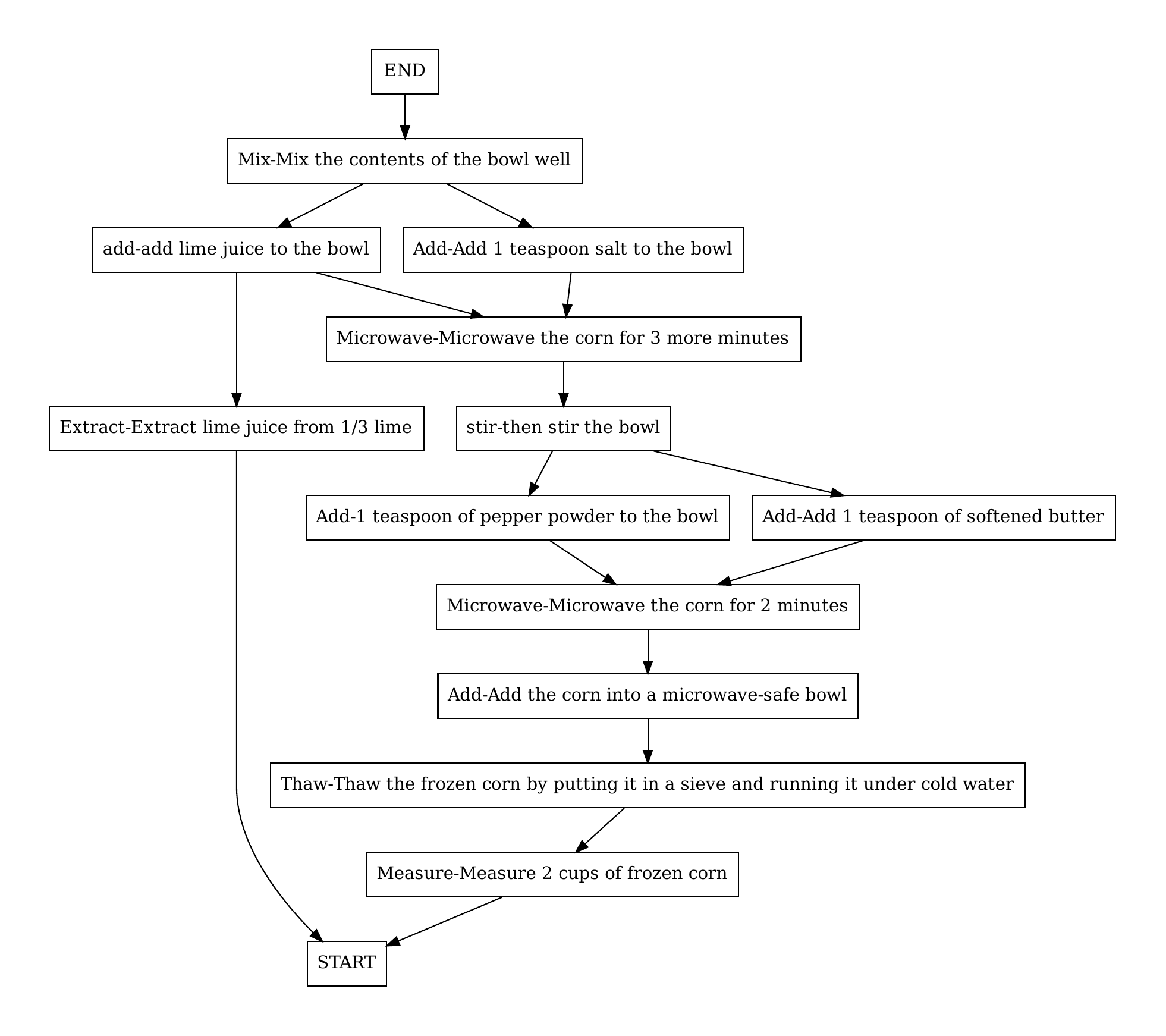}
  \caption{}
\end{subfigure}
\begin{subfigure}{.49\textwidth}
  \centering
  \includegraphics[width=\linewidth]{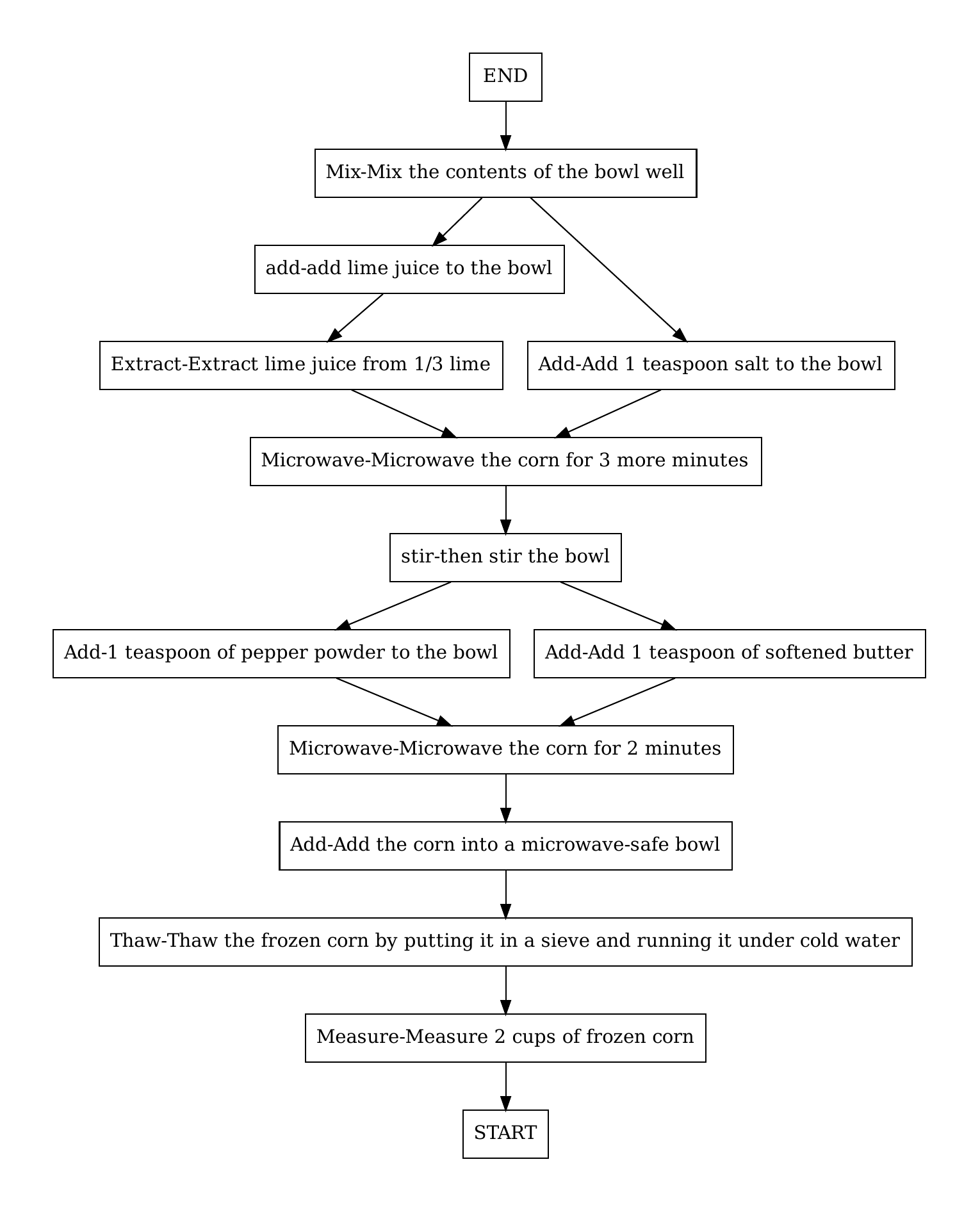}
  \caption{}
\end{subfigure}
\caption{(a) Ground truth task graph and (b) predicted task graph of the scenario Butter Corn Cup.}
\end{figure}

\begin{figure}[H]
\centering
\begin{subfigure}{.49\textwidth}
  \centering
  \includegraphics[width=\linewidth]{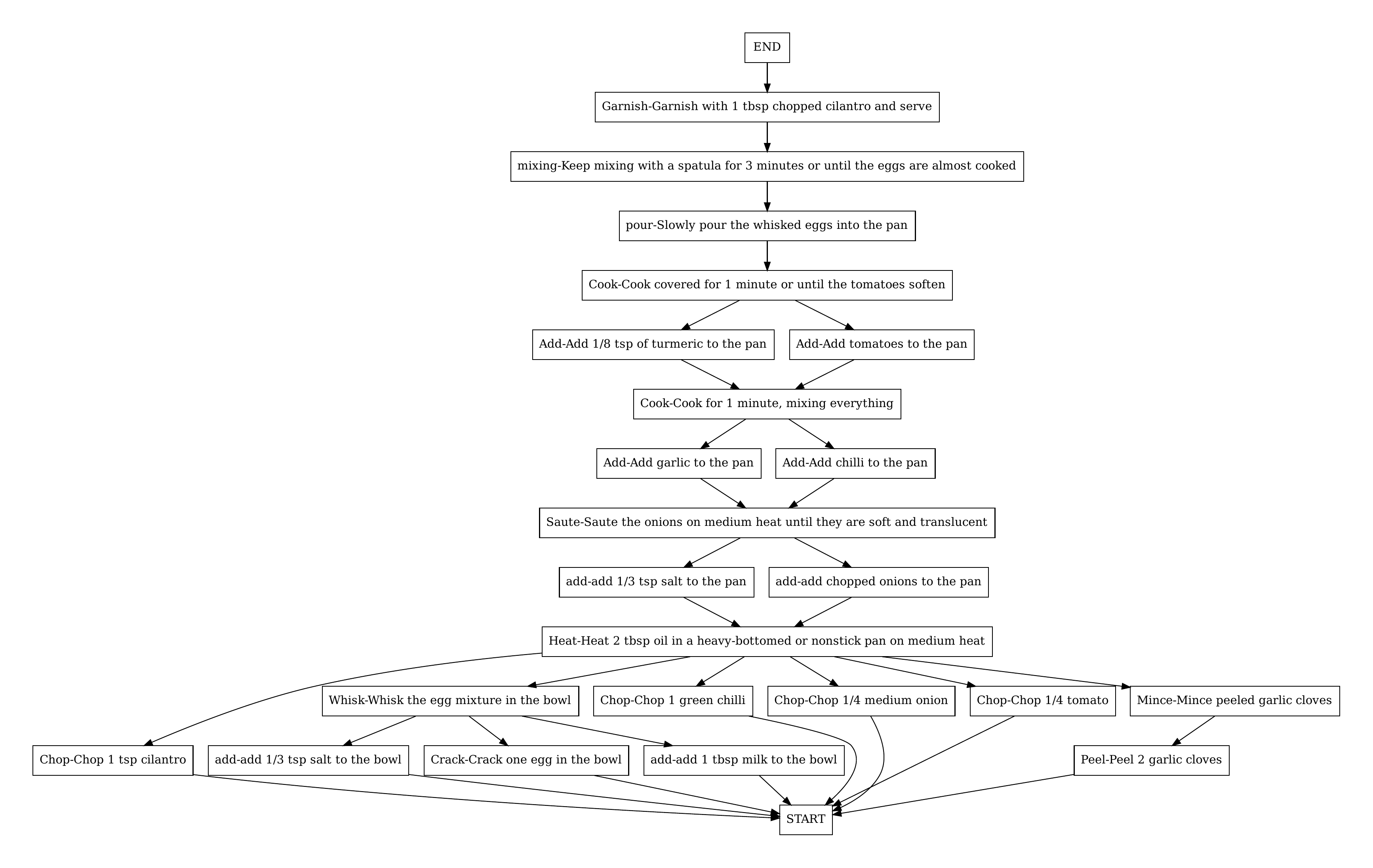}
  \caption{}
\end{subfigure}
\begin{subfigure}{.49\textwidth}
  \centering
  \includegraphics[width=\linewidth]{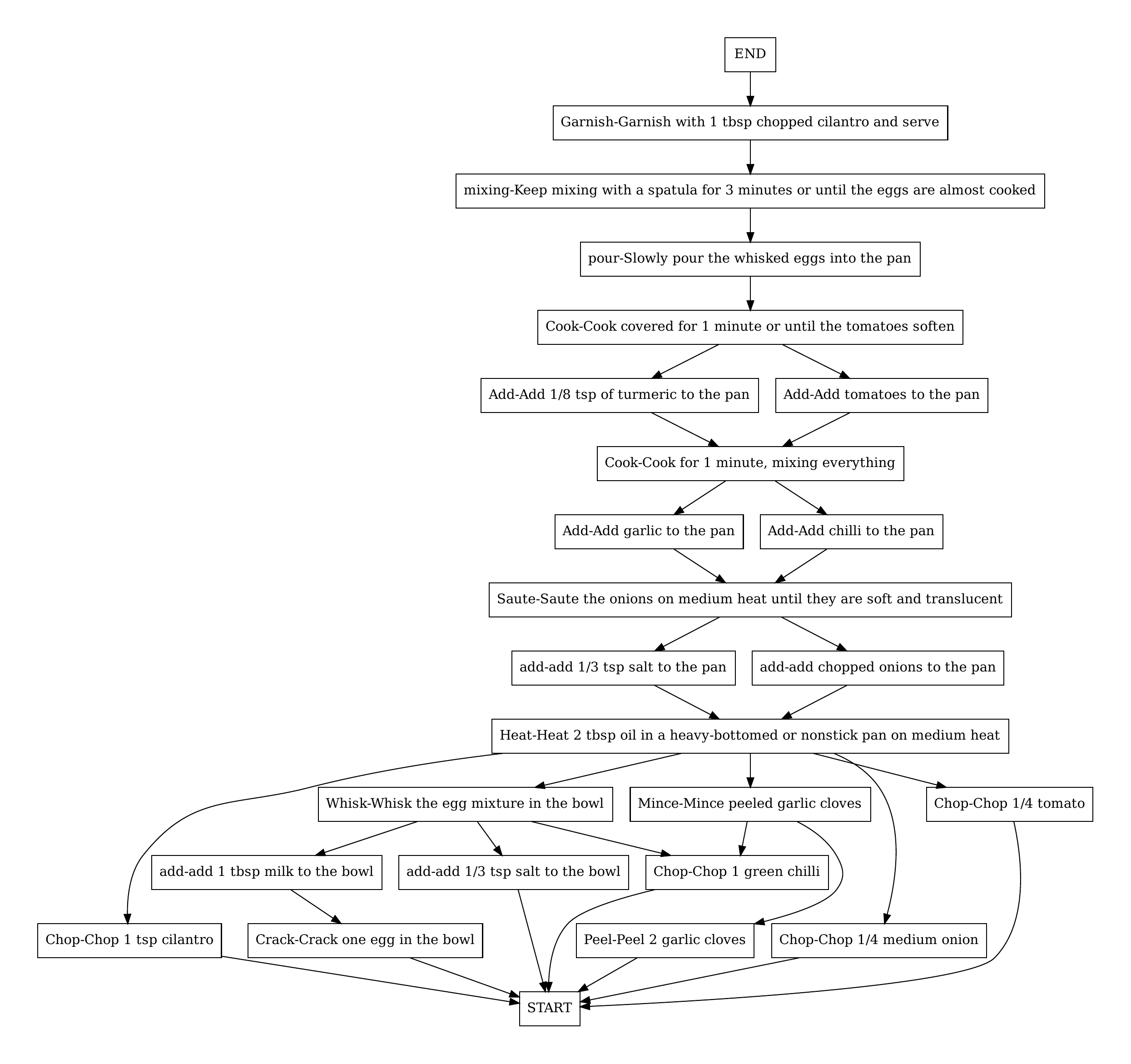}
  \caption{}
\end{subfigure}
\caption{(a) Ground truth task graph and (b) predicted task graph of the scenario Scrambled Eggs.}
\end{figure}

\end{document}